\documentclass[10pt,twocolumn,letterpaper]{article}

\usepackage{cvpr}      % To produce the REVIEW version
\usepackage{graphicx}
\usepackage{amsmath}
\usepackage{amssymb}
\usepackage{booktabs}
\usepackage{animate}
\usepackage{enumitem}
\usepackage{multirow}
\usepackage{diagbox}
\usepackage{amsmath}
\usepackage{accents}
\usepackage{graphicx}
\usepackage{amsmath}
\usepackage{amssymb}
\usepackage{booktabs}
\usepackage{float}
\usepackage{diagbox}
\usepackage{bbding}
\usepackage{utfsym}
\usepackage{pifont}
\usepackage{multirow}
\usepackage{enumitem}
\usepackage{tabularx}
\usepackage{microtype}
\usepackage{threeparttable}
\usepackage{MnSymbol}
\makeatletter
\@namedef{ver@everyshi.sty}{}
\makeatother
\usepackage{tikz}
\usepackage{makecell}

\usepackage{algorithm}
\usepackage{algpseudocode}
\usepackage{xcolor}

\usepackage{graphicx}
\usepackage{subcaption}   % 仅用于左右两栏（可选）
\usepackage{booktabs}     % \cmidrule 横线
\usepackage{array}        % 自定义列宽
\usepackage{makecell}     % 竖排文字更稳
\usepackage{ragged2e}

\usepackage{tikz}
\usetikzlibrary{calc}
\usepackage{graphicx}
\usepackage{subcaption}
\usepackage{tikz}
\usetikzlibrary{calc}
\usepackage{balance}

\definecolor{cvprblue}{rgb}{0.21,0.49,0.74}
\usepackage[pagebackref,breaklinks,colorlinks,allcolors=cvprblue]{hyperref}

\usepackage[capitalize]{cleveref}
\crefname{section}{Sec.}{Secs.}
\Crefname{section}{Section}{Sections}
\Crefname{table}{Table}{Tables}
\crefname{table}{Tab.}{Tabs.}

\def\paperID{1576} % *** Enter the Paper ID here
\def\confName{CVPR}
\def\confYear{2026}

\definecolor{Gray}{gray}{0.5}
\definecolor{LightCyan}{rgb}{0.88,1,1}

\newcolumntype{a}{>{\columncolor{Gray}}c}
\newcolumntype{b}{>{\columncolor{white}}c}

\newcommand{\topparaheading}[1]{\vspace{-0.2em}\paragraph{#1.}}

\newlength{\dhatheight}

\def\ie{\textit{i.e.}}

\title{MotionAdapter: Video Motion Transfer via Content-Aware \\ Attention Customization \vspace{-0.2cm}}

\author{Zhexin Zhang$^{1,2}$,
       Yangyang Xu$^{2}$,
       Yifeng Zhu$^{2}$,
       Long Chen$^3$,
       Yong Du$^4$,
       Shengfeng He$^{5}$,
       Jun Yu$^{2}$
       \\
       \\
       \normalsize{$^1$Hangzhou Dianzi University \hspace{5mm} $^2$Harbin Institute of Technology (Shenzhen) \hspace{5mm} } \\
       \normalsize{$^3$The Hong Kong University of Science and Technology} \\ 
       \normalsize{$^4$Ocean University of China \hspace{5mm} $^5$Singapore Management University \hspace{5mm} \vspace{-1mm}}}

\begin{document}

\teaser{
    \parbox{\linewidth}{
        \centering
        {\scriptsize\texttt{Carousel spinning in a playground $\rightarrow$ Race car drafting in circles in a parking lot}}\\
        % 各方法的第1行动画
        \begin{subfigure}{.12\linewidth}
            \centering
            \animategraphics[autoplay,loop,width=\textwidth]{12}{./teaser/Original/carousel/}{0000}{0008}
        \end{subfigure}\hspace{-0.5mm}
        \begin{subfigure}{.12\linewidth}
            \centering
            \animategraphics[autoplay,loop,width=\textwidth]{12}{./teaser/SMM/carousel/}{0000}{0008}
        \end{subfigure}\hspace{-0.5mm}
        \begin{subfigure}{.12\linewidth}
            \centering
            \animategraphics[autoplay,loop,width=\textwidth]{12}{./teaser/MOFT/carousel/}{0000}{0008}
        \end{subfigure}\hspace{-0.5mm}
        \begin{subfigure}{.12\linewidth}
            \centering
            \animategraphics[autoplay,loop,width=\textwidth]{12}{./teaser/DeT/carousel/}{0000}{0008}
        \end{subfigure}\hspace{-0.5mm}
        \begin{subfigure}{.12\linewidth}
            \centering
            \animategraphics[autoplay,loop,width=\textwidth]{12}{./teaser/DiTFlow/carousel/}{0000}{0008}
        \end{subfigure}\hspace{-0.5mm}
        \begin{subfigure}{.12\linewidth}
            \centering
            \animategraphics[autoplay,loop,width=\textwidth]{12}{./teaser/FastVMT/carousel/}{0000}{0008}
        \end{subfigure}\hspace{-0.5mm}
        \begin{subfigure}{.12\linewidth}
            \centering
            \animategraphics[autoplay,loop,width=\textwidth]{12}{./teaser/ropec/carousel/}{0000}{0008}
        \end{subfigure}\hspace{-0.5mm}
        \begin{subfigure}{.12\linewidth}
            \centering
            \animategraphics[autoplay,loop,width=\textwidth]{12}{./teaser/Ours/carousel/}{0000}{0008}
        \end{subfigure}
    }
    % ========== 第二行 ==========
    \parbox{\linewidth}{
        \centering
        {\scriptsize\texttt{BMX biker jumps over a hill on a mountain bike $\rightarrow$ Woman on water slide goes up and down}}\\
        % 各方法的第2行动画
        \begin{subfigure}{.12\linewidth}
            \centering
            \animategraphics[autoplay,loop,width=\textwidth]{12}{./teaser/Original/motocross-bumps/}{0000}{0008}
        \end{subfigure}\hspace{-0.5mm}
        \begin{subfigure}{.12\linewidth}
            \centering
            \animategraphics[autoplay,loop,width=\textwidth]{12}{./teaser/SMM/motocross-bumps/}{0000}{0008}
        \end{subfigure}\hspace{-0.5mm}
        \begin{subfigure}{.12\linewidth}
            \centering
            \animategraphics[autoplay,loop,width=\textwidth]{12}{./teaser/MOFT/motocross-bumps/}{0000}{0008}
        \end{subfigure}\hspace{-0.5mm}
        \begin{subfigure}{.12\linewidth}
            \centering
            \animategraphics[autoplay,loop,width=\textwidth]{12}{./teaser/DeT/motocross-bumps/}{0000}{0008}
        \end{subfigure}\hspace{-0.5mm}
        \begin{subfigure}{.12\linewidth}
            \centering
            \animategraphics[autoplay,loop,width=\textwidth]{12}{./teaser/DiTFlow/motocross-bumps/}{0000}{0008}
        \end{subfigure}\hspace{-0.5mm}
        \begin{subfigure}{.12\linewidth}
            \centering
            \animategraphics[autoplay,loop,width=\textwidth]{12}{./teaser/FastVMT/motocross-bumps/}{0000}{0008}
        \end{subfigure}\hspace{-0.5mm}
        \begin{subfigure}{.12\linewidth}
            \centering
            \animategraphics[autoplay,loop,width=\textwidth]{12}{./teaser/ropec/motocross-bumps/}{0000}{0008}
        \end{subfigure}\hspace{-0.5mm}
        \begin{subfigure}{.12\linewidth}
            \centering
            \animategraphics[autoplay,loop,width=\textwidth]{12}{./teaser/Ours/motocross-bumps/}{0000}{0008}
        \end{subfigure}
    }
    % \vspace{3mm}
    % ========== 第三行 ==========
    \parbox{\linewidth}{
        \centering
        {\scriptsize\texttt{A black swan swimming in a river $\rightarrow$ A paper boat floating in a bathtub}}\\
        % 各方法的第3行动画
        \begin{subfigure}{.12\linewidth}
            \centering
            \animategraphics[autoplay,loop,width=\textwidth]{12}{./teaser/Original/black_swan/}{0000}{0008}
        \end{subfigure}\hspace{-0.5mm}
        \begin{subfigure}{.12\linewidth}
            \centering
            \animategraphics[autoplay,loop,width=\textwidth]{12}{./teaser/SMM/black_swan/}{0000}{0008}
        \end{subfigure}\hspace{-0.5mm}
        \begin{subfigure}{.12\linewidth}
            \centering
            \animategraphics[autoplay,loop,width=\textwidth]{12}{./teaser/MOFT/black_swan/}{0000}{0008}
        \end{subfigure}\hspace{-0.5mm}
        \begin{subfigure}{.12\linewidth}
            \centering
            \animategraphics[autoplay,loop,width=\textwidth]{12}{./teaser/DeT/black_swan/}{0000}{0008}
        \end{subfigure}\hspace{-0.5mm}
        \begin{subfigure}{.12\linewidth}
            \centering
            \animategraphics[autoplay,loop,width=\textwidth]{12}{./teaser/DiTFlow/black_swan/}{0000}{0008}
        \end{subfigure}\hspace{-0.5mm}
        \begin{subfigure}{.12\linewidth}
            \centering
            \animategraphics[autoplay,loop,width=\textwidth]{12}{./teaser/FastVMT/black_swan/}{0000}{0008}
        \end{subfigure}\hspace{-0.5mm}
        \begin{subfigure}{.12\linewidth}
            \centering
            \animategraphics[autoplay,loop,width=\textwidth]{12}{./teaser/ropec/black_swan/}{0000}{0008}
        \end{subfigure}\hspace{-0.5mm}
        \begin{subfigure}{.12\linewidth}
            \centering
            \animategraphics[autoplay,loop,width=\textwidth]{12}{./teaser/Ours/black_swan/}{0000}{0008}
        \end{subfigure}
    }
    \parbox{\linewidth}{
        \centering
        {\scriptsize\texttt{Bus driving on a street $\rightarrow$ Ant crawling in a desert}}\\
        \begin{subfigure}{.12\linewidth}
            \centering
            \animategraphics[autoplay,loop,width=\textwidth]{12}{./teaser/Original/bus/}{0012}{0020}
            \caption*{\scriptsize{Reference}}
        \end{subfigure}\hspace{-0.5mm}
        \begin{subfigure}{.12\linewidth}
            \centering
            \animategraphics[autoplay,loop,width=\textwidth]{12}{./teaser/SMM/bus/}{0012}{0020}
            \caption*{\scriptsize{SMM}}
        \end{subfigure}\hspace{-0.5mm}
        \begin{subfigure}{.12\linewidth}
            \centering
            \animategraphics[autoplay,loop,width=\textwidth]{12}{./teaser/MOFT/bus/}{0012}{0020}
            \caption*{\scriptsize{MOFT}}
        \end{subfigure}\hspace{-0.5mm}
        \begin{subfigure}{.12\linewidth}
            \centering
            \animategraphics[autoplay,loop,width=\textwidth]{12}{./teaser/DeT/bus/}{0012}{0020}
            \caption*{\scriptsize{DeT}}
        \end{subfigure}\hspace{-0.5mm}
        \begin{subfigure}{.12\linewidth}
            \centering
            \animategraphics[autoplay,loop,width=\textwidth]{12}{./teaser/DiTFlow/bus/}{0012}{0020}
            \caption*{\scriptsize{DiTFlow}}
        \end{subfigure}\hspace{-0.5mm}
        \begin{subfigure}{.12\linewidth}
            \centering
            \animategraphics[autoplay,loop,width=\textwidth]{12}{./teaser/FastVMT/bus/}{0012}{0020}
            \caption*{\scriptsize{FastVMT}}
        \end{subfigure}\hspace{-0.5mm}
        \begin{subfigure}{.12\linewidth}
            \centering
            \animategraphics[autoplay,loop,width=\textwidth]{12}{./teaser/ropec/bus/}{0012}{0020}
            \caption*{\scriptsize{RoPECraft}}
        \end{subfigure}\hspace{-0.5mm}
        \begin{subfigure}{.12\linewidth}
            \centering
            \animategraphics[autoplay,loop,width=\textwidth]{12}{./teaser/Ours/bus/}{0012}{0020}
            \caption*{\scriptsize{Ours}}
        \end{subfigure}
    }
    \caption{Qualitative comparison of motion transfer methods. Our \emph{MotionAdapter} enables robust, content-aware motion transfer within DiT-based video diffusion models, achieving temporally coherent and semantically aligned videos that preserve both reference motion and target appearance. This figure contains \emph{animated videos}, which are best viewed in Adobe Acrobat.}
    \label{fig:teaser}
}

\maketitle

\begin{abstract}
    Recent advances in diffusion-based text-to-video models, particularly those built on the diffusion transformer architecture, have achieved remarkable progress in generating high-quality and temporally coherent videos. However, transferring complex motions between videos remains challenging. In this work, we present MotionAdapter, a content-aware motion transfer framework that enables robust and semantically aligned motion transfer within DiT-based video diffusion models. Our key insight is that effective motion transfer requires \romannumeral1) explicit disentanglement of motion from appearance and \romannumeral2) adaptive customization of motion to target content. MotionAdapter first isolates motion by analyzing cross-frame attention within 3D full-attention modules to extract attention-derived motion fields. To bridge the semantic gap between reference and target videos, we further introduce a DINO-guided motion customization module that rearranges and refines motion fields based on content correspondences. The customized motion field is then used to guide the DiT denoising process, ensuring that the synthesized video inherits the reference motion while preserving target appearance and semantics. Extensive experiments demonstrate that MotionAdapter outperforms state-of-the-art methods in both qualitative and quantitative evaluations. Morever, MotionAdapter naturely support complex motion transfer and motion editing tasks such as zooming in/out and composition.
\textbf{Project Page:} \url{https://zhexin-zhang.github.io/MotionAdapter/}
\end{abstract}

% \begin{abstract}
% Recent diffusion-based Text-to-Video (T2V) models, especially those adopting the Diffusion Transformer (DiT) architecture, have demonstrated remarkable capability in generating high-quality and temporally coherent videos. However, accurately transferring complex motion dynamics across videos remains a significant challenge, as existing motion transfer methods either lack explicit motion representations or fail to adapt transferred motions to target semantics. In this work, we present MotionAdapter, a content-aware motion transfer framework for DiT-based T2V models. Our key insight is that robust motion transfer requires (i) explicit disentanglement of motion and appearance, and (ii) adaptive customization of motion to target content. We first analyze the 3D full-attention mechanism of DiT to extract attention-derived motion fields that accurately capture temporal dynamics while remaining independent of visual appearance. To bridge semantic gaps between reference and target videos, we further propose a content-aware motion customization module that leverages DINO-based semantic correspondences to rearrange and refine motion fields with forward–backward temporal consistency. This enables semantically aligned and structurally coherent motion transfer, even under large appearance or shape variations. Extensive experiments show that MotionAdapter achieves state-of-the-art performance on challenging motion transfer benchmarks, supporting multi-object transfer and motion editing tasks such as zoom and trajectory modification.
% \end{abstract}

\section{Introduction}
\label{sec:intro}

Recent advances in diffusion models have made remarkable progress in generating high-quality visual content~\cite{song2021denoising,rombach2022high}. Video Diffusion Models (VDMs), particularly those based on the Diffusion Transformer (DiT) architecture~\cite{peebles2023scalable}, have shown exceptional performance in producing temporally consistent and visually appealing video sequences~\cite{yang2025cogvideox,wan2025wan,chen2025goku}. While current VDM can generate simple motions based on text prompts, they struggle with accurately capturing the intricate dynamics in complex scenarios. Motion transfer~\cite{ling2025motionclone,shi2025decouple,Motion_Inversion,pondaven2025video,COMD,yatim2024space,huang2025videomage,ma2025follow} is proposed to overcoming this challenge, which involves transferring motion from a source video to a target video.

Early works in motion transfer primarily focused on designing space-time feature losses to guide the generation process~\cite{yatim2024space}. These methods, however, lacked an explicit motion representation, which limits their effectiveness capturing motion patterns. Several subsequent methods attempted to learn motion representations explicitly by modulating temporal-attention layers within video diffusion models~\cite{ling2025motionclone,COMD,zhao2024motiondirector}. While these approaches have successfully transferred simpler motions, they fall short in capturing more complex motion dynamics. Additionally, these methods are often tailored to 3D U-Net-based video diffusion models and do not generalize well to more advanced models like DiT. In response, recent attempts to apply motion transfer in DiT-based video diffusion models have proposed methods. DiTFlow~\cite{pondaven2025video} calculates displacement maps within DiT blocks as motion representations, DeT~\cite{shi2025decouple} smooths DiT features using temporal kernels to represent motion, and Follow-Your-Motion~\cite{ma2025follow} discriminates the temporal tokens in the attention heads, and embeds the motion representations by optimizing the temporal tokens. However, these methods shares the same limitation, once motion representations are obtained, they are applied directly to generating target videos without customization for the target content. This results in failures when reference and target videos with large semantic gap, as shown in the third sample in Fig.~\ref{fig:teaser}, most of existing works can not transfer the motion of ``\texttt{Black Swan}'' to ``\texttt{Paper Boat}'' due to the large shape gap between two objects. 

To enable robust motion transfer in video diffusion models, we argue that two core capabilities are indispensable: \romannumeral1) \emph{Explicit disentanglement of motion and appearance:} source video’s motion must be separated from its appearance to avoid appearance leakage. \romannumeral2)  \emph{Adaptive customization of motion to target content:} the transferred motion must be adjusted to match the target’s semantics and structure.

In this paper, we propose \emph{MotionAdapter}, a content-aware video motion transfer framework that refines and customizes attention flow. We first analyze the 3D full attention mechanism of DiT to explicitly disentangle motion and appearance. By comparing the similarity between attention-derived motion fields and Ground Truth (GT) optical flows, we extract DiT motion representations that accurately capture temporal dynamics while remaining independent of visual appearance. %These extracted DiT motions serve as coarse motion priors for transferring source motions to target videos. 

However, the extracted DiT motions inherently encode source-specific structural and shape information, which can lead to semantic inconsistencies when directly applied to target videos. To mitigate this issue, we introduce a content-aware motion customization strategy. Specifically, we compute semantic correspondences between source and target contents using DINO features~\cite{oquab2024dinov} for the foreground objects, then adapt and merge these customized motions with background motion fields, followed by refinement for temporal coherence. As illustrated in Fig.~\ref{fig:teaser}, our approach enables fine-grained and semantically aligned motion transfer, even in challenging scenarios involving substantial shape or structural variations. Benefiting from this content-aware customization, \emph{MotionAdapter} achieves robust and generalizable motion transfer across diverse scenes, naturally supporting complex motion transfer and motion editing tasks such as zooming in and out. Extensive experiments on motion transfer benchmarks demonstrate that our framework consistently outperforms State-of-The-Art (SoTA) methods in both quantitative and qualitative evaluations.

In summary, our contributions are threefold: 
\begin{itemize}[leftmargin=*,nosep] 
\item We propose \emph{MotionAdapter}, a training-free and content-aware framework that enables robust video motion transfer and editing through explicit motion extraction and semantic customization.
\item We disentangle motion from appearance via in-depth analysis of the 3D full-attention mechanism in DiT, and customize motion based on semantic correspondences between source and target contents. 
\item Extensive experiments demonstrate that our framework significantly outperforms SoTA methods both qualitatively and quantitatively, especially under complex motion scenarios.
\end{itemize}
  
\begin{figure*}[t]
    \centering
    \includegraphics[width=0.96\linewidth]{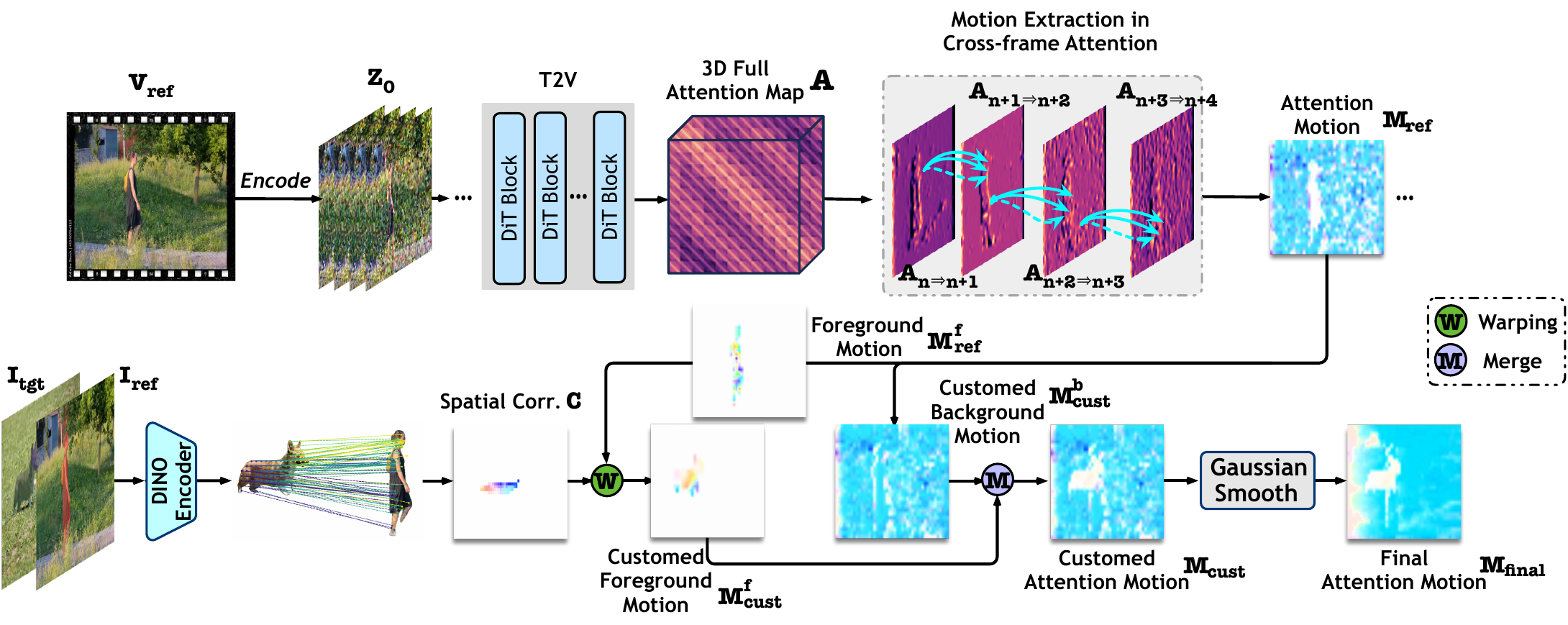} %
    \vspace{-3mm}
    \caption{Overview of our MotionAdapter. Given a reference video $\mathcal{V}_{ref}$, we first encode it to latent representations $z_0$, and pass to the T2V to get the full attention map $\mathcal{A}$. We then extract the motion in the cross-frame attention by seeking nearest Top-$K$ pixels. By analyzing temporal correspondences in cross-frame attention maps, we derive the \textit{attention motion} $\mathcal{M}_{ref}$ that disentangles motion from appearance. For custom the motion that is compatible with the target context, we introduce the \textit{content-aware motion customization}. We compute a semantic correspondence between the reference and target content using DINO features, and customize the motion field accordingly to obtain $\mathcal{M}_{cust}$, composed of foreground and background motion. Finally, Gaussian smoothing refines the customized motion into $\mathcal{M}_{final}$, which is used to guide motion transfer.}
    \label{fig:overview}
    \vspace{-3mm}
\end{figure*}

\section{Related Works}\label{sec:related}

\vspace{-1mm}
\topparaheading{Video Diffusion Models}
Following the immense success of diffusion models in Text-to-Image (T2I) generation~\cite{ho2020denoising,song2021denoising,nichol2021glide,saharia2022photorealistic}, early video generation approaches~\cite{blattmann2023stable,blattmann2023align} were extended from pre-trained T2I model~\cite{rombach2022high} by inserting temporal modules. To achieve a more unified modeling of space and time, subsequent works~\cite{wang2023modelscope,bar2024lumiere} employ 3D convolutions in a U-Net structure. More recently, the DiT architecture with 3D full attention~\cite{peebles2023scalable} has shown superior performance in video generation. By operating on spatio-temporal tokens simultaneously, current VDM works~\cite{wan2025wan,yang2025cogvideox} gains great progress on vivid and coherent videos. While successful in capturing simple motions, they struggle to model complex dynamic motions.

\vspace{-1mm}
\topparaheading{Video Motion Transfer}\label{sec:related_VM}
Video motion transfer aims to synthesize a novel video that adheres to the motion of a given reference video, which overcomes the limitations of current VDMs in complex motion capturing. Existing works have approached this by conditioning the generation process on explicit motion representations but require training on large-scale datasets~\cite{namekata2025sg,geng2025motion,xing2025motioncanvas,yang2024direct,ma2024follow,ma2024follow1,qiu2024freetraj,gokmen2025ropecraft,ma2024trailblazer}. Consequently, reference video based methods~\cite{jeong2024vmc,zhao2024motiondirector,wang2025motion} have been proposed that decouple motion and content from a single video by leveraging pre-trained VDMs. The core idea is to extract motion representations from the reference video. DMT~\cite{yatim2024space} introduces a space-time feature loss to preserve overall motion. MotionDirector~\cite{zhao2024motiondirector}, MotionInversion~\cite{wang2025motion}, and VideoMage~\cite{huang2025videomage} share a similar idea by introducing LoRA~\cite{hu2022lora} to fine-tune the motion embedding such that it decouples the motion features of the reference video. MOFT~\cite{xiao2024video} introduces Principal Component Analysis (PCA) to extract motion aware features, while MotionClone~\cite{ling2025motionclone} subtracts cross frame attention features. These works are specialized for 3D U-Net based video diffusion models, and cannot be applied to DiT based video diffusion models since DiT based models utilize 3D full attention without explicit spatio-temporal decoupling. Recently, DiTFlow~\cite{pondaven2025video} computes displacements from full DiT attention features to extract motion features. Follow-Your-Motion~\cite{ma2025follow} discriminates motion tokens from full attention tokens and introduces LoRA to learn motion representations. DeT~\cite{shi2025decouple} encodes motion representations in DiT video based diffusion into temporal kernels. 
In the very recent, RoPECraft~\cite{gokmen2025ropecraft} facilitates motion transfer by optimizing the RoPE within transformers, while FastVMT~\cite{ma2026fastvmt} accelerates the process by streamlining motion extraction and optimization.
After obtaining the motion representations, these works transfer them to new videos directly, facing challenges when there is a large semantic gap between two videos.

\vspace{-1mm}

\topparaheading{Diffusion Feature Decoupling} 
Current large-scale diffusion models demonstrate strong visual quality and generative capability. Many studies explore feature decoupling to obtain more interpretable representations.
Recent works decouple diffusion features at the frequency level for temporal consistency~\cite{wu2024freeinit,lu2024freelong} or separate spatial information~\cite{yuan2025freqprior,park2025spectral}. DIFT~\cite{tang2023emergent} and SD-DINO~\cite{zhang2023tale} use PCA to extract semantic components.
For video diffusion models, MOFT~\cite{xiao2024video} analyzes inter-channel motion relations, while Follow-Your-Motion~\cite{ma2025follow} selects motion relevant attention heads.
In this paper, we introduce DINO~\cite{oquab2024dinov} to learn the correspondence between objects in two videos, guiding the customization of attentions of video diffusions that enables precise and coherent motion transfer. It is worth noting that MotionShot~\cite{liu2025motionshot} also introduce DINO to learn object correspondence for motion tranfer. However, it neglects motion in background regions, leading to failures in transferring camera motion, which is primarily embedded in the background.    
\vspace{-1mm}
\section{Method}\label{sec:method}

\vspace{-2mm}

\subsection{Problem Formulation}

\vspace{-2mm}

Given a reference video $\mathcal{V}_{\text{ref}} = \{I_1, I_2, \dots \}$ that provides motion information and a target text prompt $\mathcal{P}_{\text{tgt}}$ describing the desired scene and subjects, 
our goal is to synthesize a video $\hat{\mathcal{V}}_{\text{tgt}}$ that preserves the motion pattern of $\mathcal{V}_{\text{ref}}$ 
while matching the appearance and semantics specified by $\mathcal{P}_{\text{tgt}}$. As discussed in Sec.\ref{sec:intro}, the key challenge lies in disentangling motion from appearance and customizing the transferred motion to ensure semantic alignment with the target scene. In this paper, we propose \emph{MotionAdapter} that enables robust video motion transfer by extract and customize reference motions, the overview of our framework is shown in Fig.~\ref{fig:overview}.

\subsection{Preliminary}\label{sec:method_dit}

\noindent{\textbf{Video Diffusion Models}}.
Video diffusion models~\cite{yang2025cogvideox,chen2025goku,wan2025wan} consist of a pretrained 3D Variational AutoEncoder (VAE) and a $T$-step denoising network with transformer structure. 
Given a video $\mathcal{V}$, the encoder $\mathcal{E}$ maps it to a latent representation ${z}_0=\mathcal{E}(\mathcal{V})$, and the decoder $\mathcal{D}$ reconstructs it as $\hat{\mathcal{V}}=\mathcal{D}({z}_0)$. 
At each step $t\in\{0,\dots,T\}$, the latent ${z}_t\in\mathbb{R}^{f\times c\times h\times w}$ is estimated from ${z}_{t+1}$, where $f$, $c$, $h$, and $w$ denote the frame, channel, height, and width dimensions, respectively. 

For text guidance, a T5 encoder~\cite{raffel2020exploring} converts the prompt $\mathcal{P}$ into embeddings $\tau_{\theta}(\mathcal{P})$, which are concatenated with ${z}_t$ in the full-attention module. 
The denoising process is formulated as:
\begin{equation}
{z}_t = \epsilon_{\theta}({z}_{t-1}, t, \tau_{\theta}(\mathcal{P})),
\end{equation}
where $\epsilon_{\theta}$ denotes a time-conditional Transformer composed of $N$ DiT blocks. 
During inference, ${z}_T \sim \mathcal{N}(0,1)$ is iteratively denoised to obtain ${z}_0$, which is then decoded by $\mathcal{D}$ to generate the final video. Current VDMs commonly adopt the DiT architecture, and our \emph{MotionAdapter} is built upon the Video DiT backbone.

\noindent{\textbf{3D Full Attention in Video DiT}}.\label{para:full_Transformer}
Current DiT-based diffusion models utilize a 3D full-attention module to capture complex spatio-temporal dependencies within a video. 
During each denoising step, the 3D full-attention module takes as input a concatenated sequence composed of the text embeddings $\tau_{\theta}(\mathcal{P})$ and the flattened latent video features ${z}_t$. 
The input sequence $S_{\text{in}}$ and the self-attention map $\mathcal{A}$, computed using learnable matrices $\mathcal{W}_\text{Q}$ and $\mathcal{W}_\text{K}$, are formulated as:
\begin{equation}
S_{\text{in}} = \texttt{Concat}(\tau_{\theta}(\mathcal{P}), z_t),
\end{equation}
\begin{equation}
\mathcal{A} =
\texttt{Softmax}\!\left(
\frac{(\mathcal{W}_\text{Q} S_{\text{in}})
(\mathcal{W}_\text{K} S_{\text{in}})^{\top}}
{\sqrt{d}}
\right),
\label{eq2}
\end{equation}
where $\texttt{Concat}(\cdot,\cdot)$ denotes concatenation along the sequence dimension, and $d$ represents the feature dimension of each token. 
This full-attention mechanism enables global interactions across spatial, temporal, and textual tokens, allowing the model to jointly capture intra-frame appearance relations and inter-frame motion dependencies. 
In this work, we analyze the temporal correlations encoded in the 3D full-attention maps to explicitly disentangle appearance and motion representations.

\begin{figure}[!t] % {r} aligns it to the right; {l} for left
    \centering
    \vspace{-1em}
    \includegraphics[width=0.48\textwidth]{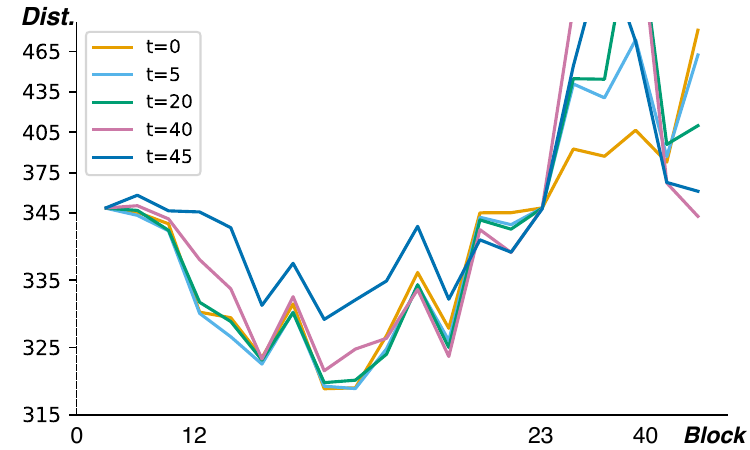}
    \vspace{-1mm}
    \caption{We plot the MSE distance between GT flow and cross-frame attention motions. Mid-level DiT blocks and lower noise timesteps yield lower distances.}
    \label{fig:analyze}
    \vspace{-2em} % Adjust spacing to fit your text layout
\end{figure}

\subsection{Disentanglement of Motion and Appearance}\label{para:disentanglement}
As discussed in Sec.~\ref{sec:intro}, the first key of motion transfer is disentangles the motion and appearance from reference video. In this section, we conduct a detailed analysis of 3D full attention features in VDMs to identify and extract robust motion cues.

\noindent{\textbf{Cross-Frame Attention Motion Extraction.}} 
As shown in the upper part of Fig.~\ref{fig:overview}, given a reference video $\mathcal{V}_{\text{ref}}$, 
we first encode it using the VAE encoder $\mathcal{E}$ to obtain latent representation ${z}_0 = \mathcal{E}(\mathcal{V}_{\text{ref}})$. 
The latent representations is perturbed with noise and then fed into the denoising network $\epsilon_{\theta}(\cdot)$ with an empty text prompt to extract the 3D full-attention maps $\mathcal{A}$ defined in Eq.~\ref{eq2}.

Let $\mathcal{A}_{n\rightarrow m} \in \mathbb{R}^{(h \times w) \times (h \times w)}$ denotes the cross-frame attention map between two frames $f_n$ and $f_{m}$, it can be extractd along the temporal dimension from $\mathcal{A}$. For a pixel $p_i$ at coordinate $(u_i, v_i)$ in frame $f_n$, we seek its nearest Top-$K$ pixels in frame $f_{m}$ using $\mathcal{A}_{n\rightarrow m}(i,j)$. Let $\mathcal{N}_K(p_i)$ denote the indices of the top-$K$ pixels in $f_\text{m}$ with the highest similarity scores. The destination coordinate $(\hat{u}_i, \hat{v}_i)$ is then obtained by averaging the coordinates of the Top-$K$ matched pixels:
\begin{equation}
(\hat{u}_i, \hat{v}_i) = \frac{1}{K} \sum_{j \in \mathcal{N}_K(p_i)} (u_j, v_j),
\label{topk}
\end{equation}
then the motion vector $(\Delta u_i, \Delta v_i)$ of pixel $p_i$ between the two frames is defined as:
\begin{equation}
(\Delta u_i, \Delta v_i) = (\hat{u}_i - u_i,~ \hat{v}_i - v_i).
\end{equation}
Thus, the attention motion $\mathcal{M}_{n\rightarrow m} = \{(\Delta u_i, \Delta v_i)\}$ represents the temporal correspondence derived from the cross-frame attention, effectively capturing motion while being disentangled from appearance.

\begin{figure}[t]
    \centering
    \vspace{-7.5mm}
    \captionsetup[subfloat]{justification=centering,labelformat=empty}
    \subfloat[Video Frame]{\includegraphics[width=0.32\linewidth]{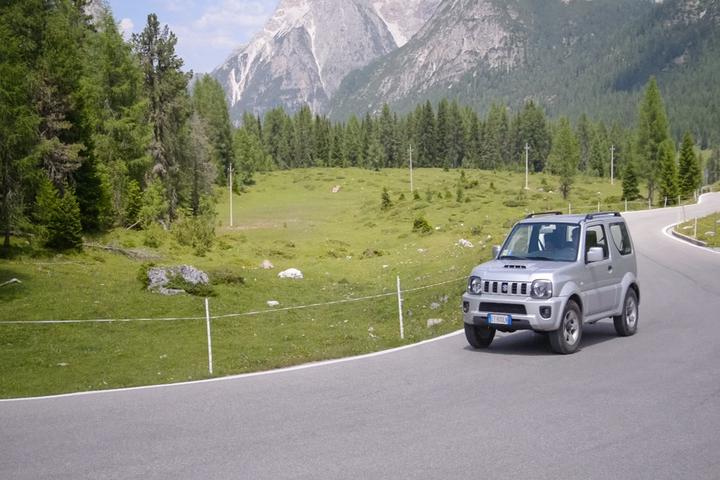}}
    \hfill
    \subfloat[GT Flow]{\includegraphics[width=0.32\linewidth]{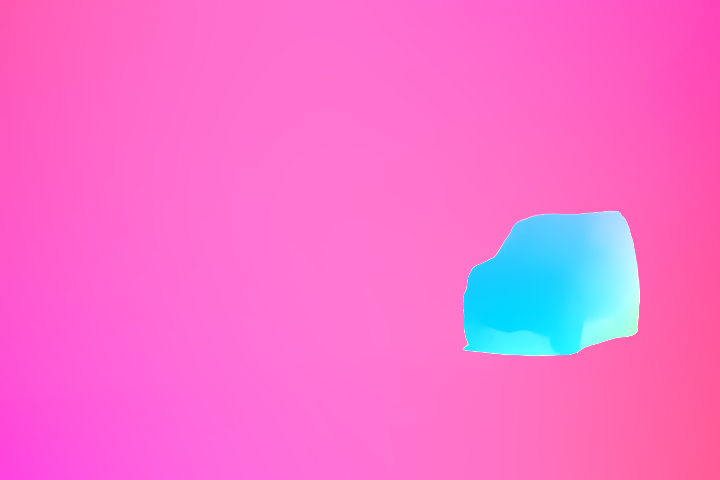}}
    \hfill
    \subfloat[$t=0,b=40$]{\includegraphics[width=0.32\linewidth]{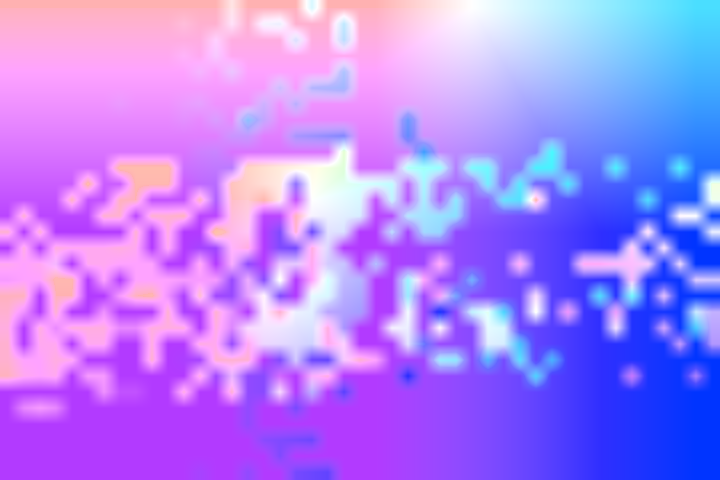}}\\
    \hfill
    \subfloat[$t=40,b=4$]{\includegraphics[width=0.32\linewidth]{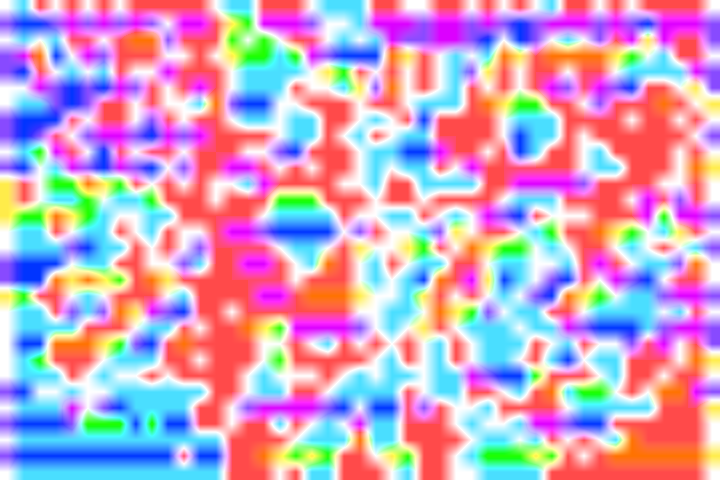}}
    \hfill
    \subfloat[$t=35,b=12$]{\includegraphics[width=0.32\linewidth]{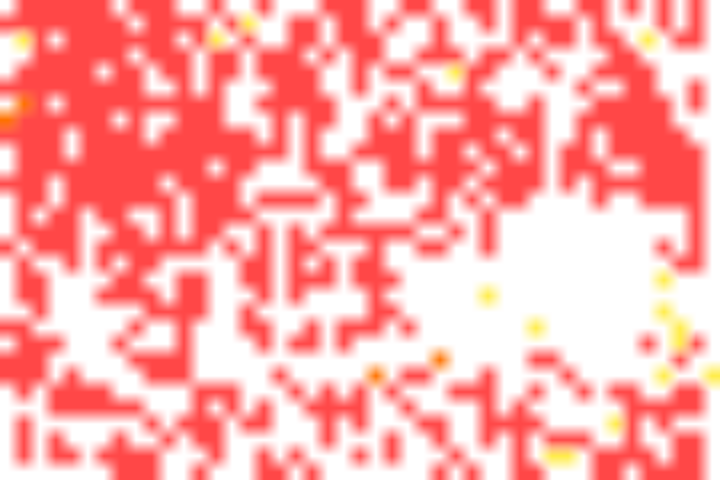}}
    \hfill
    \subfloat[$t=5,b=18$]{\includegraphics[width=0.32\linewidth]{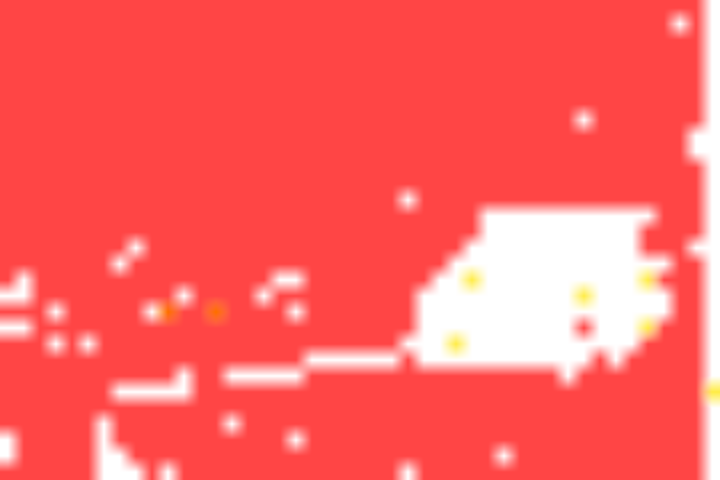}}
    \vspace{-1mm}
    \caption{We visualize the attention motion field obtained from various noise steps and attention blocks, and the motion derived from $t=5,b=18$ is more simarity with the GT optical flow.}
    \vspace{-7.5mm}
    \label{fig:figure_attn}
\end{figure}

\noindent{\textbf{Selection of Cross-Frame Attention Motions.}}
The VDMs typically consists of multiple DiT blocks, each producing its own cross-frame attention motion at every denoising step. To identify which DiT block and timestep of the DiT features best align with the GT motion, we add noise to the latent representation ${z}_0$ at different timesteps $t$ and extract the cross-frame attention motion $\mathcal{M}^\text{b}_\text{t}$ from various DiT blocks $b$. We then compute the Mean Squared Error (MSE) between each $\mathcal{M}^\text{b}_\text{t}$ and the GT optical flow extracted by~\cite{dong2024memflow}.

As shown in Fig.~\ref{fig:analyze}, by averaging the results over 100 videos, we observe that attention flows extracted at lower noise levels exhibit smaller MSE distances, indicating a stronger correspondence with real motion. In addition, the flows obtained from mid-level DiT blocks ($13 \leq b \leq 21$) are more consistent with the GT optical flow. As illustrated in Fig.~\ref{fig:figure_attn}, the flow derived from the $5_{th}$ timestep and the $18_{th}$ DiT block shows the closest alignment to the GT motion. Therefore, we adopt this configuration for disentangling appearance and motion representations in our subsequent analysis.

\noindent{\textbf{Motion Transfer.}}
After selecting the cross-frame attention motion $\mathcal{M}_{\text{ref}}$ from the reference video, we transfer the motion to the target video by aligning it with the target motion field $\mathcal{M}_{\text{tgt}}$ extracted during inference. 
To achieve this, we follow prior works~\cite{xiao2024video,pondaven2025video,yatim2024space,ling2025motionclone} that optimize the latent representation ${z}_t$ of the target video to minimize the discrepancy between the two motion fields:
\begin{equation}
{z}_t^{*} = 
\arg\min_{{z}_t} 
\| \mathcal{M}_{\text{tgt}} - \mathcal{M}_{\text{ref}} \|_2^2.
\label{eq:motion_opt}
\end{equation}
By minimizing Eq.~\ref{eq:motion_opt}, the motion field of the target video are adapted from the reference motion while maintaining the appearance guided by the text prompt.

\subsection{Content-Aware Motion Customization}\label{sec:motion_customization}

\begin{figure}[t]
    \centering
    \vspace{-9.5mm}
    \captionsetup[subfloat]{justification=centering}
    \subfloat[\tiny{Reference Video}]{
     \begin{minipage}{0.300\linewidth}
     \includegraphics[width=\linewidth]{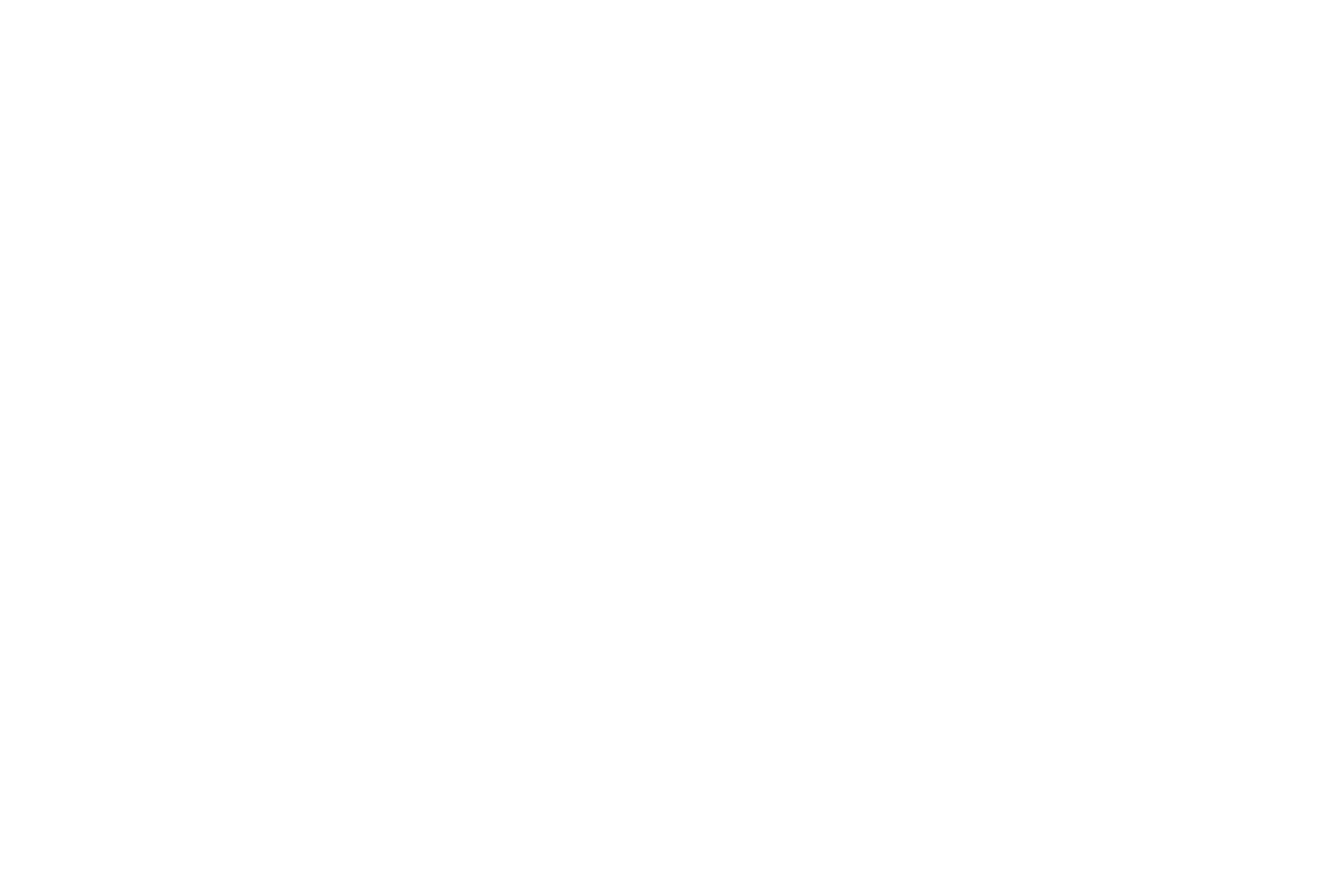}
     \includegraphics[width=\linewidth]{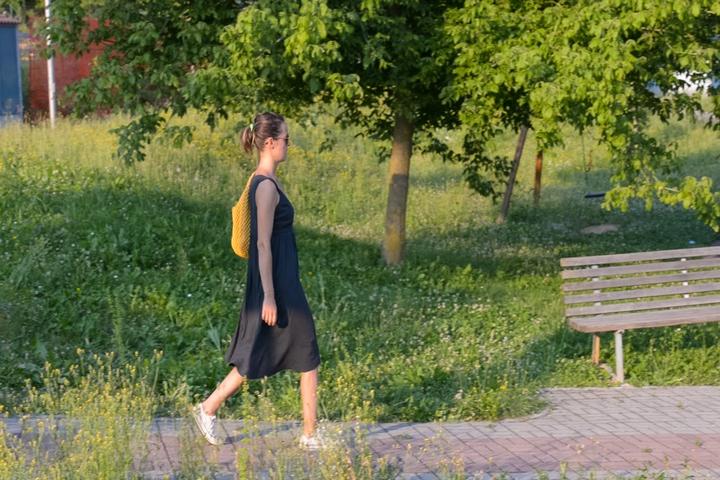}
     \includegraphics[width=\linewidth]{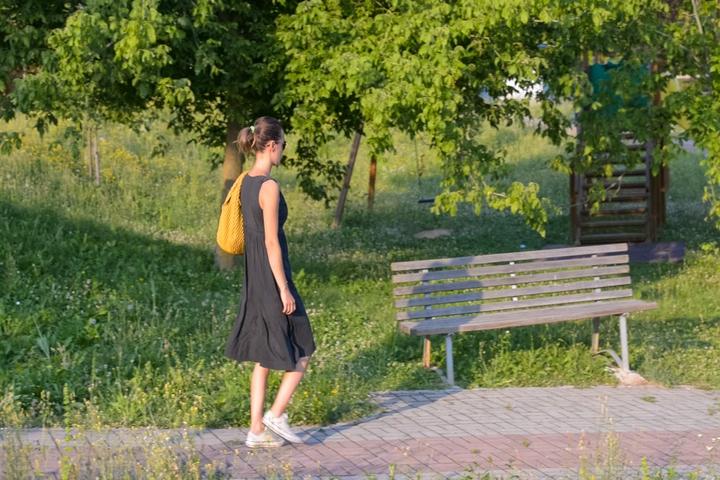}
     \end{minipage}\label{fig:ref}
     }
    \hspace{-3mm}
    \subfloat[{w/o Motion Customization}]{
     \begin{minipage}{0.300\linewidth}
     \includegraphics[width=\linewidth]{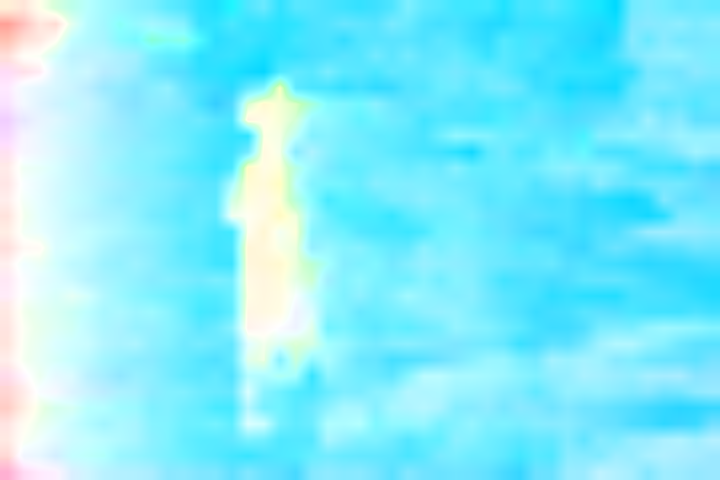}
     \includegraphics[width=\linewidth]{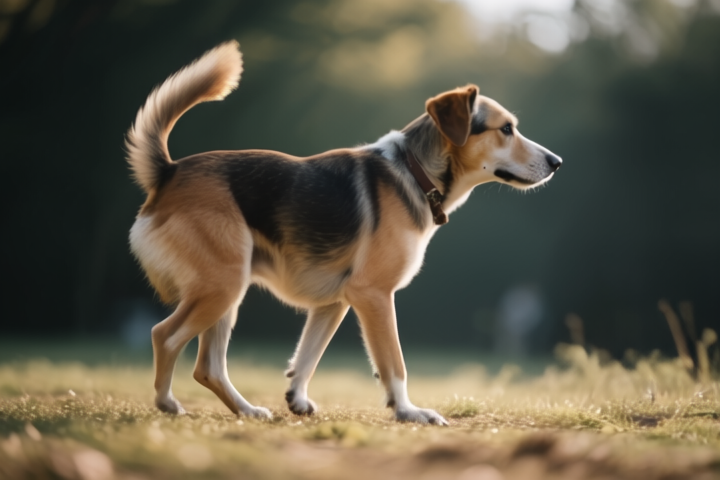}
     \includegraphics[width=\linewidth]{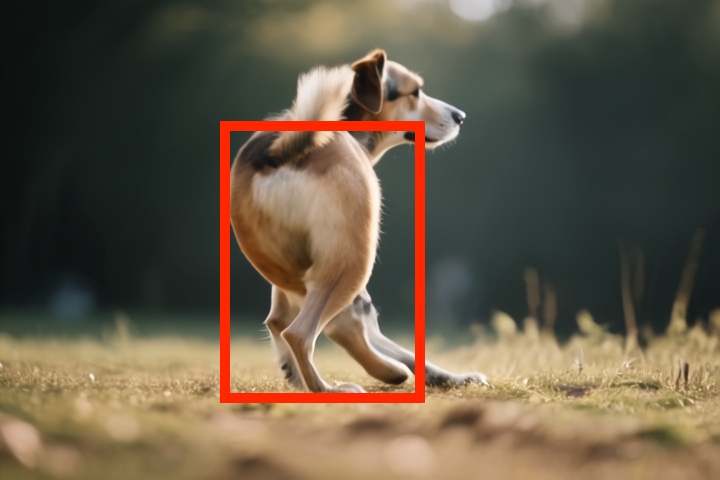}
     \end{minipage}\label{fig:womc}
     }
    \hspace{-3mm}
    \subfloat[{w/ Motion Customization}]{
     \begin{minipage}{0.300\linewidth}
     \includegraphics[width=\linewidth]{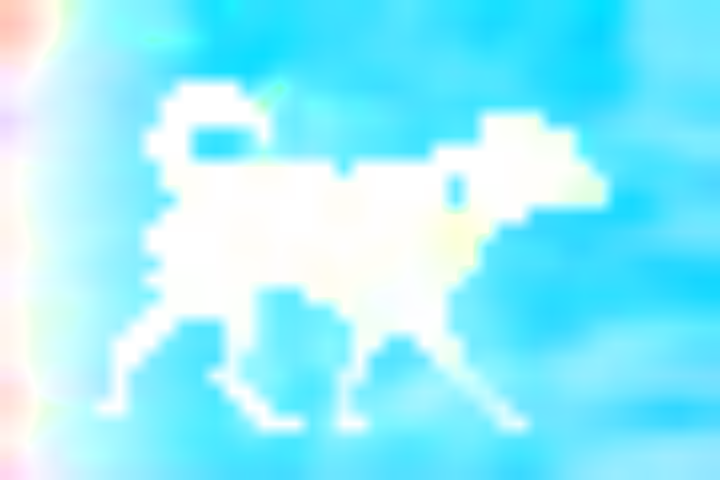}
     \includegraphics[width=\linewidth]{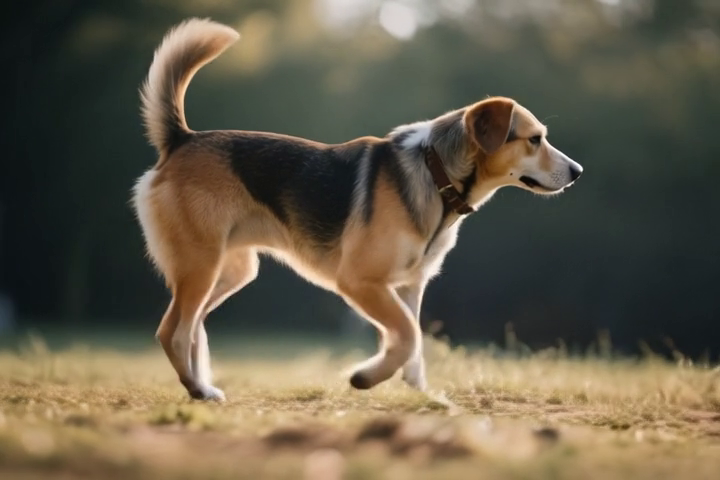}
     \includegraphics[width=\linewidth]{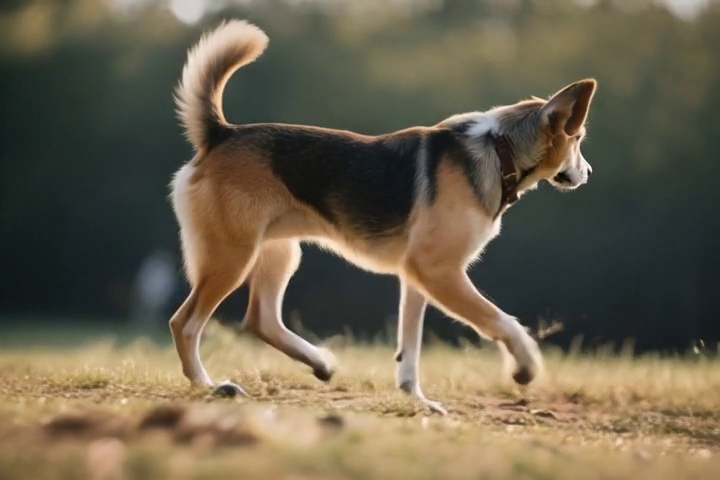}
     \end{minipage}\label{fig:wmc}
     }
\rotatebox[origin=c]{270}{\hspace{-20mm} \small{$f_{n}$} \hspace{4mm}  \small{$f_{n+1}$}}\\
\vspace{-2mm}
\caption{The attention extracted from reference video contains the reference-specific shape information, resulting in the geometric distortions in the target videos (see in red box).}
\label{fig:figure_cust}
\vspace{-7.5mm}
\end{figure}

Although the attention motion field $\mathcal{M}_{\text{ref}}$ effectively represents the temporal dynamics of the reference video, 
it inevitably encodes reference-specific structural and shape information, and directly aligning the target motion to $\mathcal{M}_{\text{ref}}$ often leads to geometric distortions when there are significant shape or scale discrepancies between the reference and target videos (see in Fig.~\ref{fig:womc}), especially for the object motion trasnfer. To address this issue, we propose a \textit{content-aware motion customization} module that adapts the reference motion field to the semantics and geometry of the target content.

\noindent{\textbf{Attention Motion Customization.}}  
As shown in the bottom part of Fig.~\ref{fig:overview}, we first segment the foreground objects from source and target frames with Lang-SAM~\cite{medeiros2023langsegmentanything}, we also segment the foreground and background motion from $\mathcal{M}_{{ref}}$ accordingly.
Then we utilize DINO~\cite{oquab2024dinov} feature extractor $\mathcal{E}_\texttt{DINO}$ to compute a spatial correspondence map $\mathcal{C}$ between the reference and target objects. Speficically, for each pixel $(j,k)$ in the target object, we find its most similar feature in the reference object $(\hat{j}, \hat{k})$ by nearest-neighbor search in the feature space. 
To ensure global consistency, we apply the Hungarian algorithm to resolve conflicts and obtain an optimal correspondence set:
\begin{equation}
\mathcal{C} = \{(j,k) \leftrightarrow (\hat{j},\hat{k}) ~|~\texttt{Hungarian}(O_{\text{tgt}}{(j,k)}, O_\text{ref}{(\hat{j},\hat{k})}) \}.
\label{eq:feature_match}
\end{equation}
This correspondence map $\mathcal{C}$ captures semantic and structural alignment between reference and target objects, allowing motion to be transferred even when appearance differs significantly. Now we can obtain the customized motion field of foreground object $\mathcal{M}^\text{f}_\text{cust}$ by warping the foreground attention motion field $\mathcal{M}^\text{f}_\text{ref}$ based on the spatial correspondence map $\mathcal{C}$, that is,
\begin{equation}
\mathcal{M}^\text{f}_\text{cust} = \texttt{W}(\mathcal{M}^\text{f}_\text{ref}, \mathcal{C})),
\end{equation}
where $\texttt{W}(\cdot,\cdot)$ is the warping operation.

For the background motion, we first inpaint it using nearest-neighbor interpolation over the foreground-missing regions to obtain a complete motion field.
Then, we merge the foreground motion with the inpainted background motion to produce the final customized motion field, that is:
\begin{equation}
\mathcal{M}^\text{b}_{\text{cust}}
= \texttt{NN}(\mathcal{M}^\text{b}_{\text{ref}}, \texttt{Mask}),
\end{equation}
\begin{equation}
\mathcal{M}_{\text{cust}}
= (1-\texttt{Mask}) \odot \mathcal{M}^\text{f}_{\text{cust}}
+ \texttt{Mask} \odot \mathcal{M}^\text{b}_{\text{cust}},
\end{equation}
where $\texttt{NN}(\cdot,\cdot)$ is the nearest-neighbor interpolation, and $\texttt{Mask}$ is the foreground mask of reference frame, $\mathcal{M}_{\text{cust}}$ is the final customized motion field.

\noindent{\textbf{Attention Motion Refinement.}}  
To further enhance the robustness of the customized motion against noise and discontinuities, we further perform Gaussian smoothing to refine the customed motion field. This process suppresses high-frequency perturbations from the attention map and produces more coherent motion trajectories. 
The final customized motion field $\mathcal{M}_{\text{final}}$ can be represented as $\mathcal{M}_\text{final}=\texttt{GauSmooth}(\mathcal{M}_\text{cust})$. Now we can rewrite Eq.~\ref{eq:motion_opt} by replacing $\mathcal{M}_{\text{ref}}$ with $\mathcal{M}_{\text{final}}$:
\begin{equation}
{z}_t^{*} = 
\arg\min_{{z}_t} 
\| \mathcal{M}_{\text{tgt}} - \mathcal{M}_{\text{final}} \|_2^2.
\label{eq:motion_final}
\end{equation}
As shown in Fig.~\ref{fig:figure_cust}, compared with original reference motion, the final customed motion ensures that the transferred motion aligns with the target object’s semantics and structure, enabling more natural and coherent motion transfer.

\subsection{Unified Motion Transfer Pipelines}%\label{sec:motion_customization}

\emph{MotionAdapter} supports both Image-to-Video (I2V) and Text-to-Video (T2V) transfer through a unified workflow. For the I2V setup, we extract motions from the reference video (Sec.~\ref{para:disentanglement}), compute the correspondence $\mathcal{C}$ between the first reference frame and the target image (Eq.~\ref{eq:feature_match}), and generate the final video using the customized motion $\mathcal{M}_\text{final}$ via Eq.~\ref{eq:motion_final}, and we name this pipeline as \emph{MotionAdapter}-I2V. When no condition image is provided, we extend this to a T2V pipeline by first generating a consultation video $V_c$ using the text prompt. We then perform frame-wise customization between the reference video and $V_c$ to derive the motion field, subsequently following the same optimization-based generation process. Similarly, we name this pipeline as \emph{MotionAdapter}-T2V.
    
\section{Experiments}

\subsection{Experimental Settings}

\label{sec:es}

\textbf{Dataset.}
Following prior work~\cite{pondaven2025video,xiao2024video,yatim2024space}, we evaluate our method on a curated subset of 50 videos from the DAVIS dataset~\cite{pont20172017}. Each reference video is paired with three target prompts corresponding to different difficulty levels: \textbf{Easy}, where the prompt closely matches the reference content. \textbf{Medium}, where the foreground object is changed while the background remains similar. And \textbf{Hard}, where both foreground and background differ significantly. This results in a total of 150 prompt–video pairs. Videos containing fewer than 49 frames are padded to 49 frames for evaluation consistency.

\textbf{Metrics.}
We evaluate our method in terms of both video quality and motion consistency. 
For video quality, following prior work~\cite{wang2025motion,shi2025decouple,pondaven2025video}, we report the average frame-wise CLIP Score (CS)~\cite{hessel2021clipscore} to measure prompt alignment. 
To evaluate video temporal consistency, following prior work~\cite{shi2025decouple}, we report Temporal Consistency (TC)~\cite{huang2024vbench}, which calculates the cosine similarity between the features of consecutive frames to quantify the coherence of the generated video.
For motion consistency, 
we report Motion Fidelity (MF)~\cite{yatim2024space}, a metric compares the similarity of tracking~\cite{karaev2024cotracker} results between reference video and generated video.

\textbf{Competitors.}
We compare \emph{MotionAdapter} against a comprehensive set of open-source motion transfer works, including SMM~\cite{yatim2024space}, MOFT~\cite{xiao2024video}, MotionClone~\cite{ling2025motionclone}, MotionInversion~\cite{wang2025motion}, %MotionDirector~\cite{zhao2024motiondirector}, 
DiTFlow~\cite{pondaven2025video}, DeT~\cite{shi2025decouple}, and FastVMT~\cite{ma2026fastvmt}. Note that DiTFlow, DeT and FastVMT are Full DiT-based methods, while the others are designed for U-Net video diffusion models in their original report.

\begin{figure*}[t]
    \centering
    
    \newcommand{\rotateLabel}[1]{%
        \begin{minipage}[c]{3mm}
            \centering
            \rotatebox{90}{\scalebox{0.75}{\scriptsize \textbf{#1}}} % 我加了 \textbf 加粗，看起来更像标题，不需要可去掉
        \end{minipage}%
    }
    
    \newcommand{\rowWidth}{0.98\linewidth}
    
    \rotateLabel{Reference\quad}\hspace{-3mm}
    \begin{minipage}[c]{\rowWidth}
        \centering
        {\scriptsize \texttt{Flamingo drinking water in a pond} \qquad \texttt{Goat walking over rocks on a mountain} \qquad \texttt{Dog jumping in a grassy field}}\\
        \begin{subfigure}{.105\linewidth}
            \includegraphics[width=\linewidth]{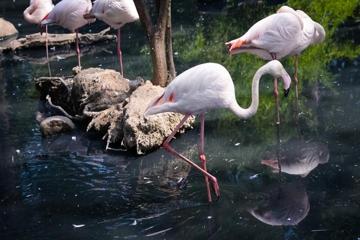}
        \end{subfigure}\hspace{-1mm}
        \begin{subfigure}{.105\linewidth}
            \includegraphics[width=\linewidth]{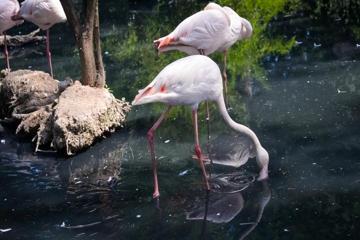}
        \end{subfigure}\hspace{-1mm}
        \begin{subfigure}{.105\linewidth}
            \includegraphics[width=\linewidth]{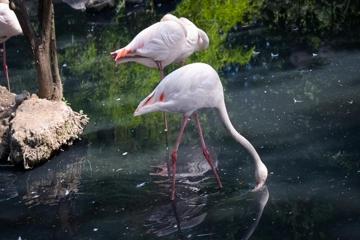}
        \end{subfigure}\hspace{-0.5mm}
        \begin{subfigure}{.105\linewidth}
            \includegraphics[width=\linewidth]{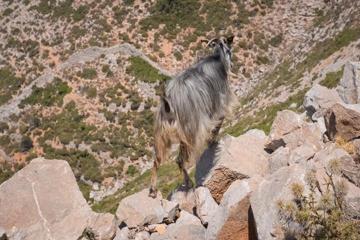}
        \end{subfigure}\hspace{-1mm}
        \begin{subfigure}{.105\linewidth}
            \includegraphics[width=\linewidth]{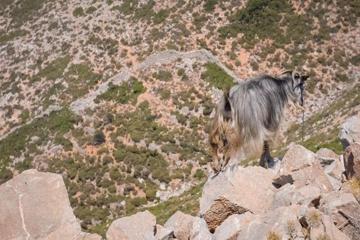}
        \end{subfigure}\hspace{-1mm}
        \begin{subfigure}{.105\linewidth}
            \includegraphics[width=\linewidth]{comparison/goat_hard/ref/00046}
        \end{subfigure}\hspace{-0.5mm}
        \begin{subfigure}{.105\linewidth}
            \includegraphics[width=\linewidth]{comparison/dogs-jump_hard/ref/1}
        \end{subfigure}\hspace{-1mm}
        \begin{subfigure}{.105\linewidth}
            \includegraphics[width=\linewidth]{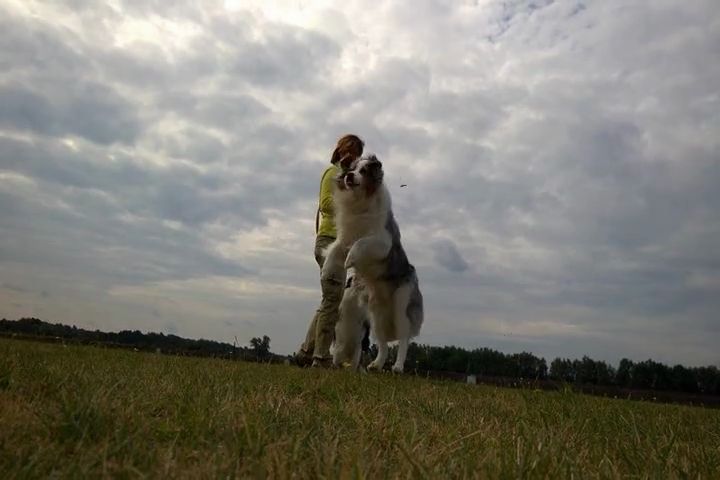}
        \end{subfigure}\hspace{-1mm}
        \begin{subfigure}{.105\linewidth}
            \includegraphics[width=\linewidth]{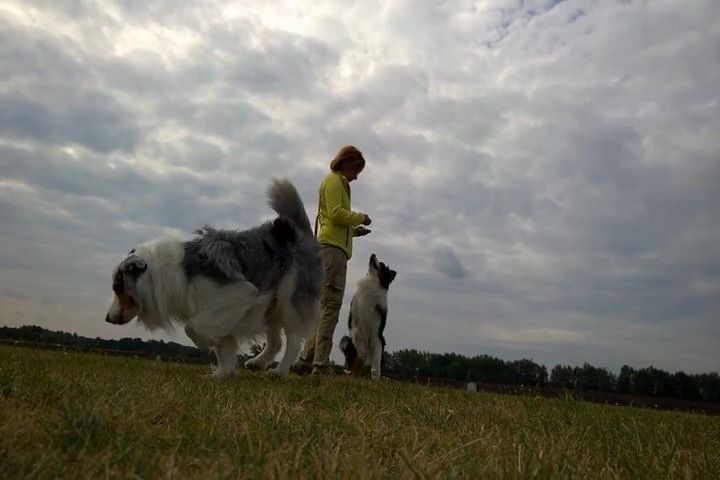}
        \end{subfigure}
    \end{minipage}
    % \vspace{1mm} % 行间距
    % ===== 第2行 (SMM) =====
    \rotateLabel{SMM\quad}\hspace{-3mm}
    %\hfill
    \begin{minipage}[c]{\rowWidth}
        \centering
        {\scriptsize \texttt{~~A swan drinking water from puddle} \qquad \texttt{Jaguar walking in snowy forest} \quad \qquad \texttt{~~Goat jumping between rocks in canyon}}\\
        \begin{subfigure}{.105\linewidth}
            \includegraphics[width=\linewidth]{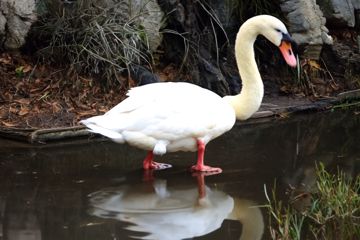}
        \end{subfigure}\hspace{-1mm}
        \begin{subfigure}{.105\linewidth}
            \includegraphics[width=\linewidth]{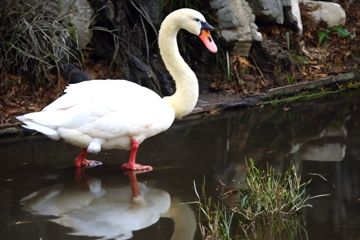}
        \end{subfigure}\hspace{-1mm}
        \begin{subfigure}{.105\linewidth}
            \includegraphics[width=\linewidth]{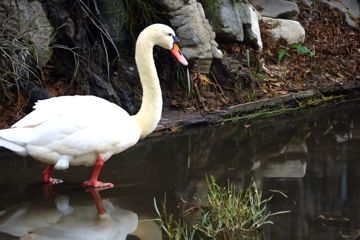}
        \end{subfigure}\hspace{-0.5mm}
        \begin{subfigure}{.105\linewidth}
            \includegraphics[width=\linewidth]{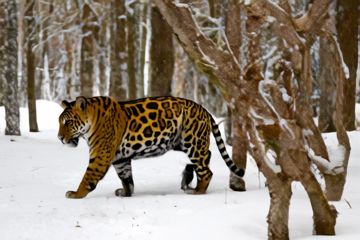}
        \end{subfigure}\hspace{-1mm}
        \begin{subfigure}{.105\linewidth}
            \includegraphics[width=\linewidth]{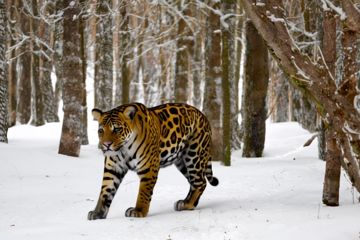}
        \end{subfigure}\hspace{-1mm}
        \begin{subfigure}{.105\linewidth}
            \includegraphics[width=\linewidth]{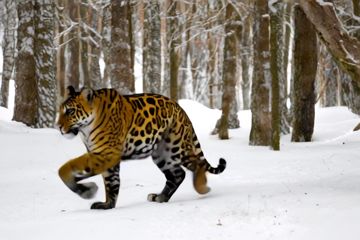}
        \end{subfigure}\hspace{-0.5mm}
        \begin{subfigure}{.105\linewidth}
            \includegraphics[width=\linewidth]{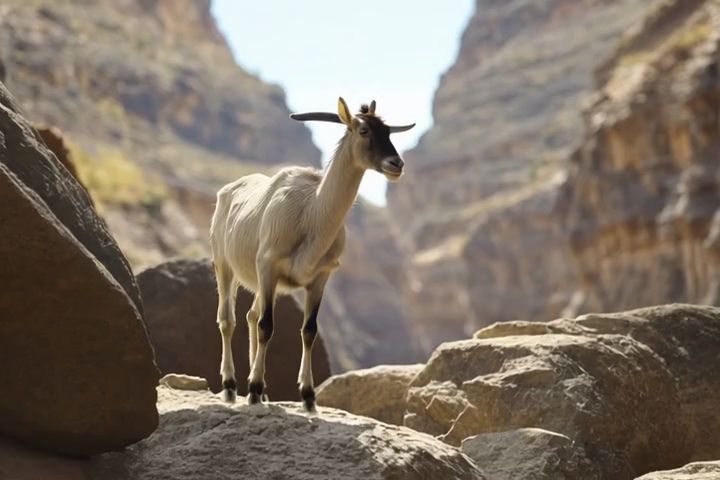}
        \end{subfigure}\hspace{-1mm}
        \begin{subfigure}{.105\linewidth}
            \includegraphics[width=\linewidth]{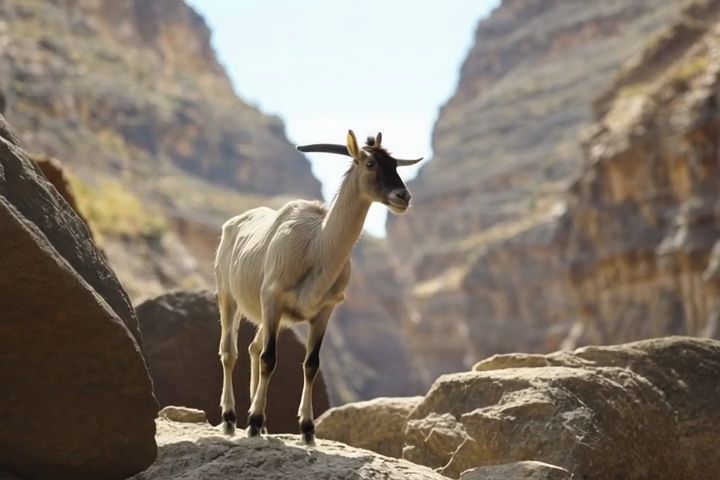}
        \end{subfigure}\hspace{-1mm}
        \begin{subfigure}{.105\linewidth}
            \includegraphics[width=\linewidth]{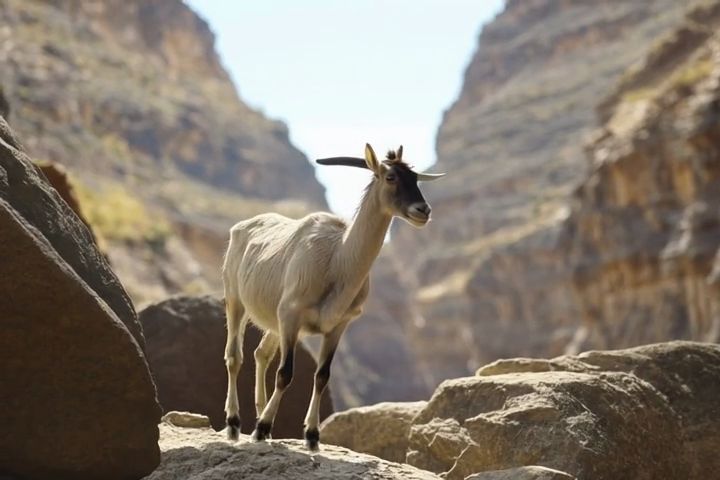}
        \end{subfigure}
    \end{minipage}
    % \vspace{1mm}
    % ===== 第3行 (MClone) =====
    \rotateLabel{MClone}\hspace{-3mm}
    %\hfill
    \begin{minipage}[c]{\rowWidth}
        \centering
        \begin{subfigure}{.105\linewidth}
            \includegraphics[width=\linewidth]{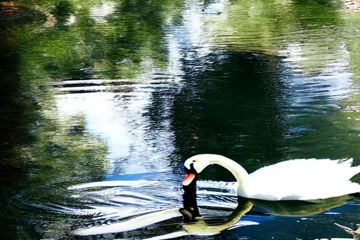}
        \end{subfigure}\hspace{-1mm}
        \begin{subfigure}{.105\linewidth}
            \includegraphics[width=\linewidth]{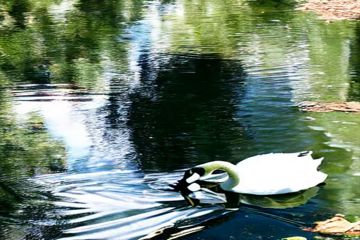}
        \end{subfigure}\hspace{-1mm}
        \begin{subfigure}{.105\linewidth}
            \includegraphics[width=\linewidth]{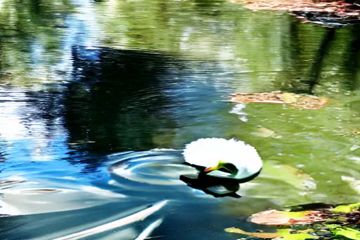}
        \end{subfigure}\hspace{-0.5mm}
        \begin{subfigure}{.105\linewidth}
            \includegraphics[width=\linewidth]{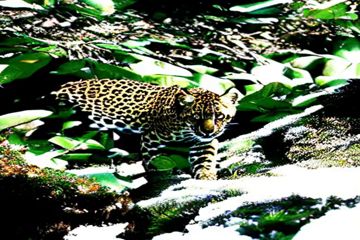}
        \end{subfigure}\hspace{-1mm}
        \begin{subfigure}{.105\linewidth}
            \includegraphics[width=\linewidth]{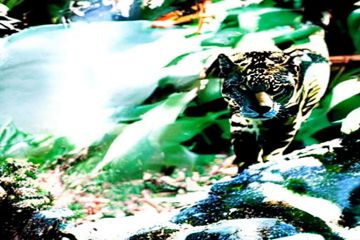}
        \end{subfigure}\hspace{-1mm}
        \begin{subfigure}{.105\linewidth}
            \includegraphics[width=\linewidth]{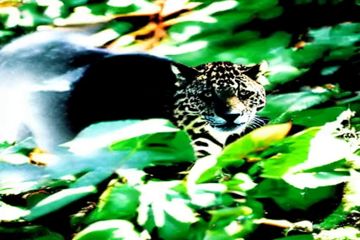}
        \end{subfigure}\hspace{-0.5mm}
        \begin{subfigure}{.105\linewidth}
            \includegraphics[width=\linewidth]{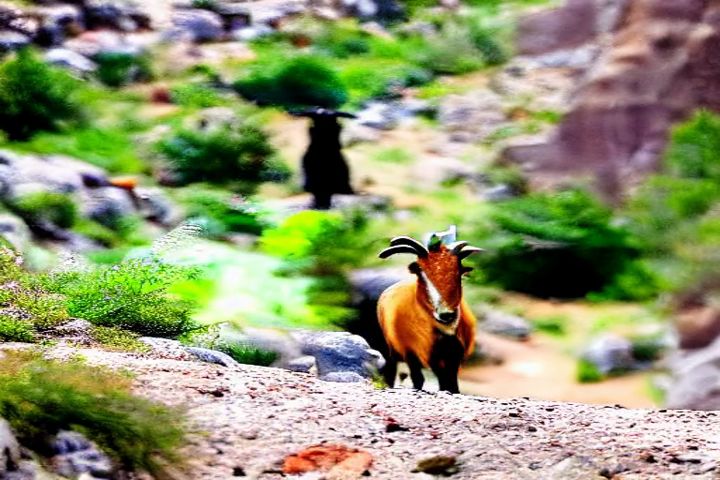}
        \end{subfigure}\hspace{-1mm}
        \begin{subfigure}{.105\linewidth}
            \includegraphics[width=\linewidth]{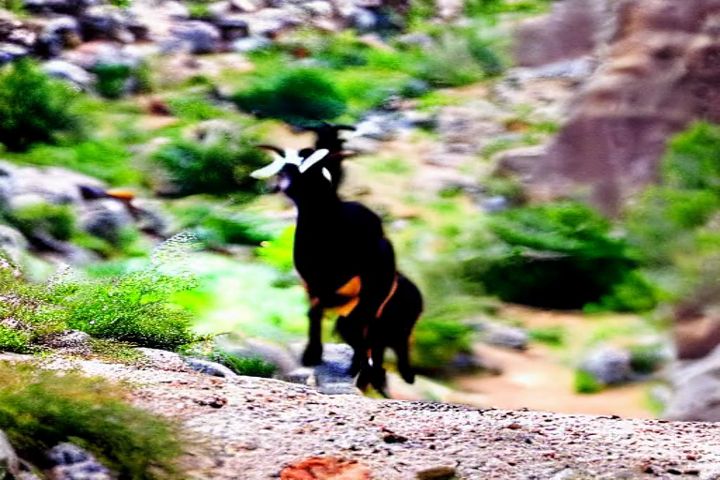}
        \end{subfigure}\hspace{-1mm}
        \begin{subfigure}{.105\linewidth}
            \includegraphics[width=\linewidth]{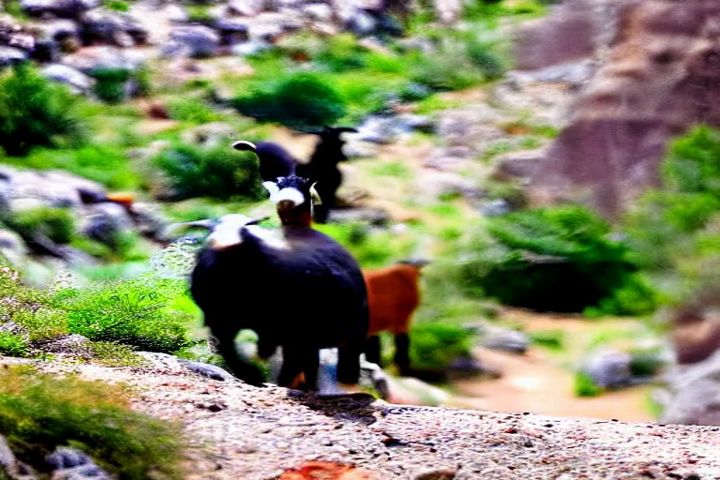}
        \end{subfigure}
    \end{minipage}
    % \vspace{1mm}
    % ===== 第4行 (MInv) =====
    \rotateLabel{MInv}\hspace{-3mm}
    %\hfill
    \begin{minipage}[c]{\rowWidth}
        \centering
        \begin{subfigure}{.105\linewidth}
            \includegraphics[width=\linewidth]{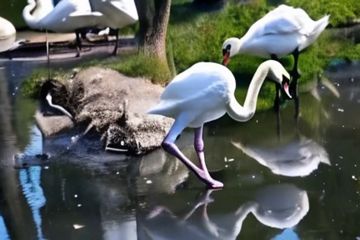}
        \end{subfigure}\hspace{-1mm}
        \begin{subfigure}{.105\linewidth}
            \includegraphics[width=\linewidth]{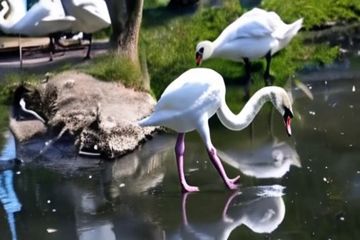}
        \end{subfigure}\hspace{-1mm}
        \begin{subfigure}{.105\linewidth}
            \includegraphics[width=\linewidth]{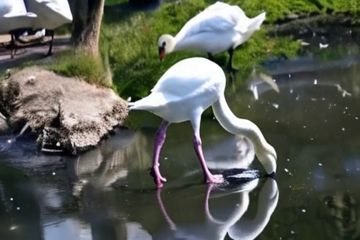}
        \end{subfigure}\hspace{-0.5mm}
        \begin{subfigure}{.105\linewidth}
            \includegraphics[width=\linewidth]{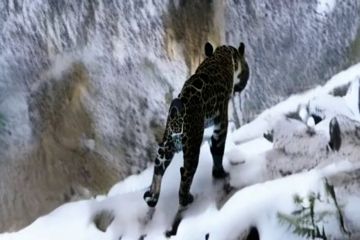}
        \end{subfigure}\hspace{-1mm}
        \begin{subfigure}{.105\linewidth}
            \includegraphics[width=\linewidth]{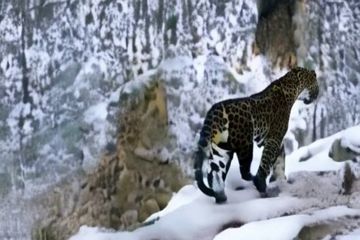}
        \end{subfigure}\hspace{-1mm}
        \begin{subfigure}{.105\linewidth}
            \includegraphics[width=\linewidth]{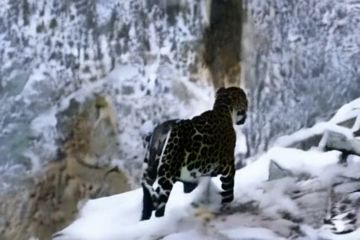}
        \end{subfigure}\hspace{-0.5mm}
        \begin{subfigure}{.105\linewidth}
            \includegraphics[width=\linewidth]{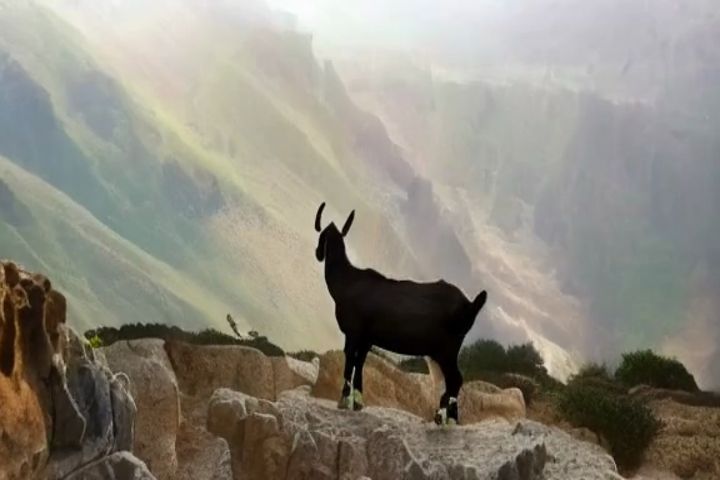}
        \end{subfigure}\hspace{-1mm}
        \begin{subfigure}{.105\linewidth}
            \includegraphics[width=\linewidth]{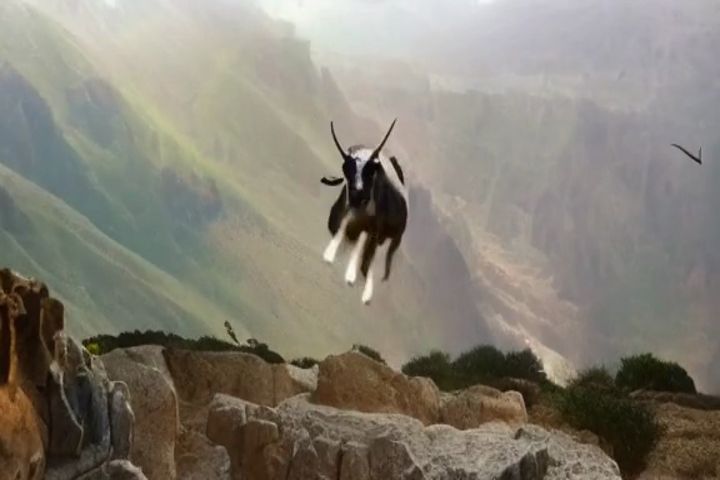}
        \end{subfigure}\hspace{-1mm}
        \begin{subfigure}{.105\linewidth}
            \includegraphics[width=\linewidth]{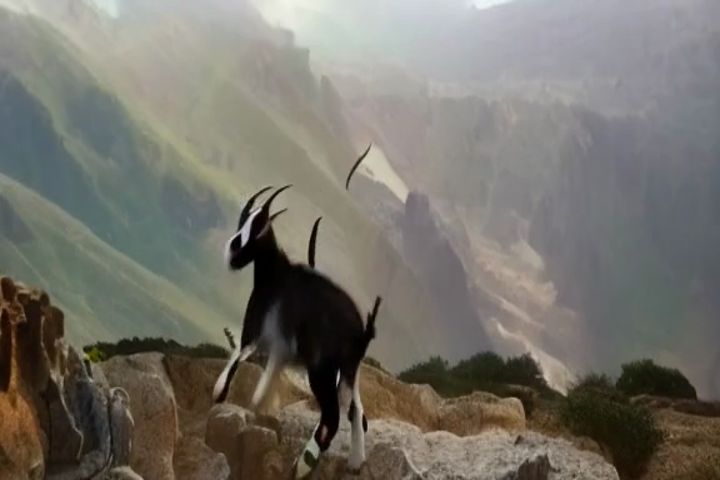}
        \end{subfigure}
    \end{minipage}
    % \vspace{1mm}
    % ===== 第5行 (MOFT) =====
    \rotateLabel{MOFT}\hspace{-3mm}
    %\hfill
    \begin{minipage}[c]{\rowWidth}
        \centering
        \begin{subfigure}{.105\linewidth}
            \includegraphics[width=\linewidth]{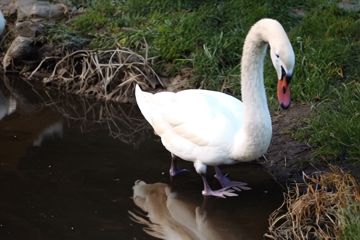}
        \end{subfigure}\hspace{-1mm}
        \begin{subfigure}{.105\linewidth}
            \includegraphics[width=\linewidth]{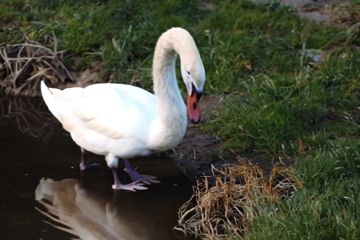}
        \end{subfigure}\hspace{-1mm}
        \begin{subfigure}{.105\linewidth}
            \includegraphics[width=\linewidth]{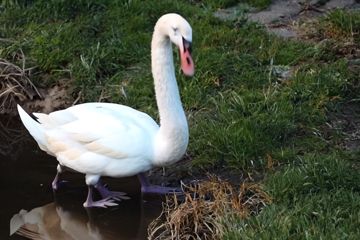}
        \end{subfigure}\hspace{-0.5mm}
        \begin{subfigure}{.105\linewidth}
            \includegraphics[width=\linewidth]{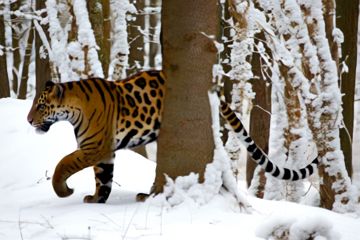}
        \end{subfigure}\hspace{-1mm}
        \begin{subfigure}{.105\linewidth}
            \includegraphics[width=\linewidth]{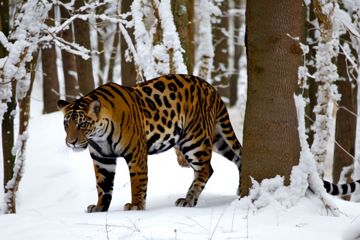}
        \end{subfigure}\hspace{-1mm}
        \begin{subfigure}{.105\linewidth}
            \includegraphics[width=\linewidth]{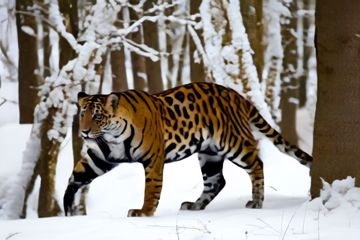}
        \end{subfigure}\hspace{-0.5mm}
        \begin{subfigure}{.105\linewidth}
            \includegraphics[width=\linewidth]{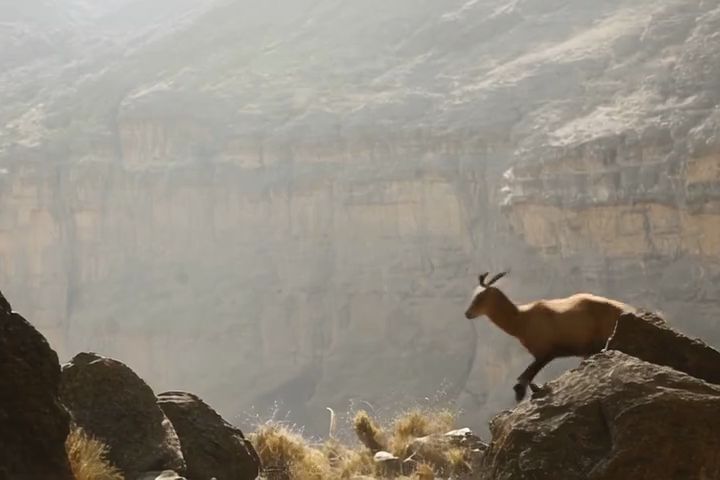}
        \end{subfigure}\hspace{-1mm}
        \begin{subfigure}{.105\linewidth}
            \includegraphics[width=\linewidth]{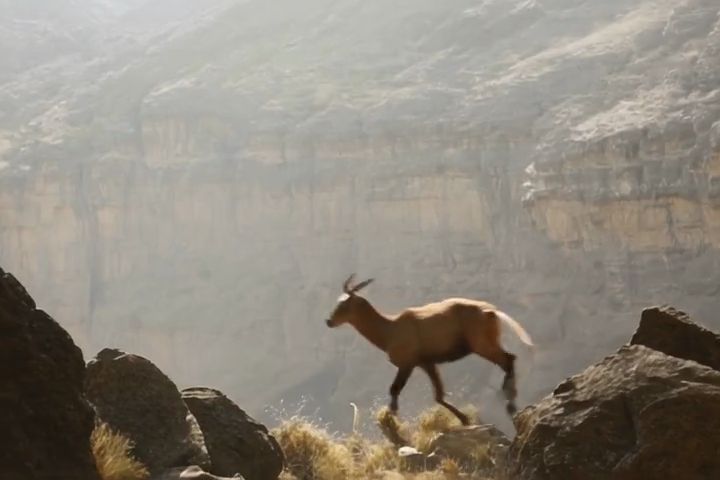}
        \end{subfigure}\hspace{-1mm}
        \begin{subfigure}{.105\linewidth}
            \includegraphics[width=\linewidth]{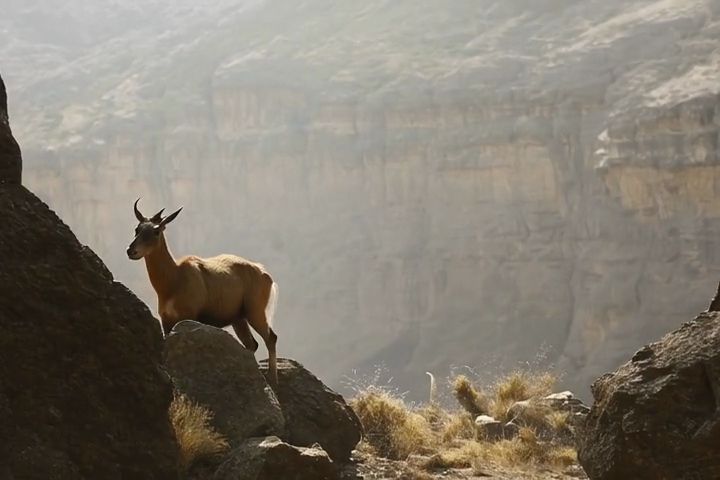}
        \end{subfigure}
    \end{minipage}
    % \vspace{1mm}
    % ===== 第6行 (DiTF) =====
    \rotateLabel{DiTF}\hspace{-3mm}
    %\hfill
    \begin{minipage}[c]{\rowWidth}
        \centering
        \begin{subfigure}{.105\linewidth}
            \includegraphics[width=\linewidth]{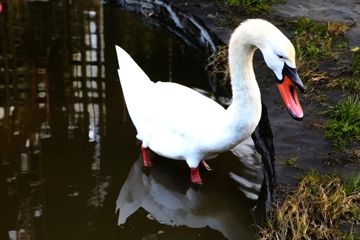}
        \end{subfigure}\hspace{-1mm}
        \begin{subfigure}{.105\linewidth}
            \includegraphics[width=\linewidth]{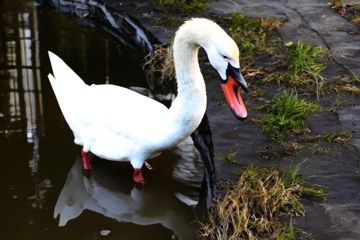}
        \end{subfigure}\hspace{-1mm}
        \begin{subfigure}{.105\linewidth}
            \includegraphics[width=\linewidth]{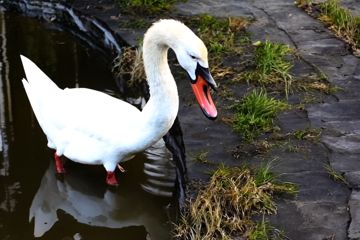}
        \end{subfigure}\hspace{-0.5mm}
        \begin{subfigure}{.105\linewidth}
            \includegraphics[width=\linewidth]{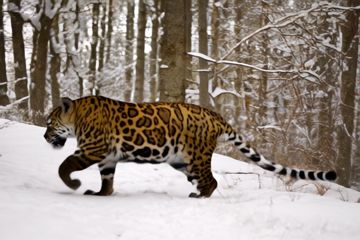}
        \end{subfigure}\hspace{-1mm}
        \begin{subfigure}{.105\linewidth}
            \includegraphics[width=\linewidth]{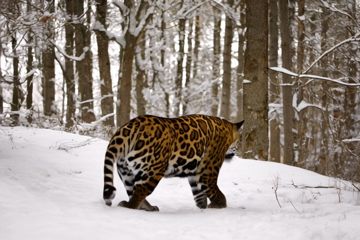}
        \end{subfigure}\hspace{-1mm}
        \begin{subfigure}{.105\linewidth}
            \includegraphics[width=\linewidth]{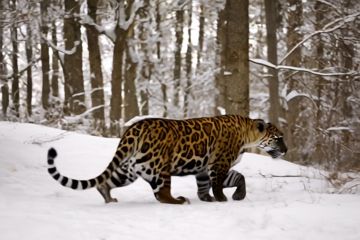}
        \end{subfigure}\hspace{-0.5mm}
        \begin{subfigure}{.105\linewidth}
            \includegraphics[width=\linewidth]{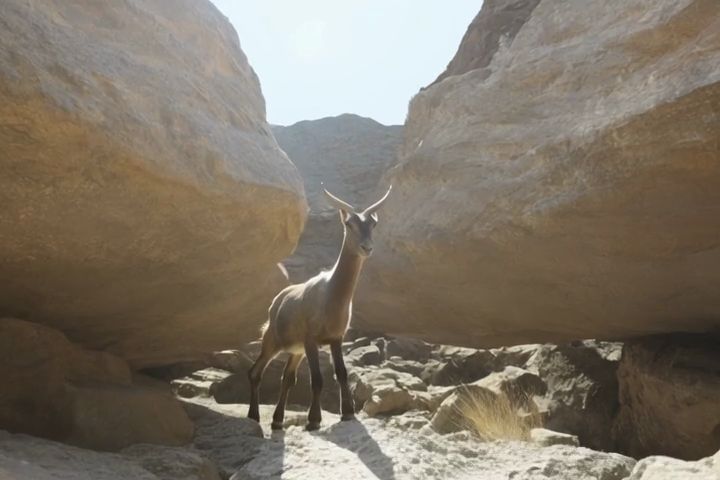}
        \end{subfigure}\hspace{-1mm}
        \begin{subfigure}{.105\linewidth}
            \includegraphics[width=\linewidth]{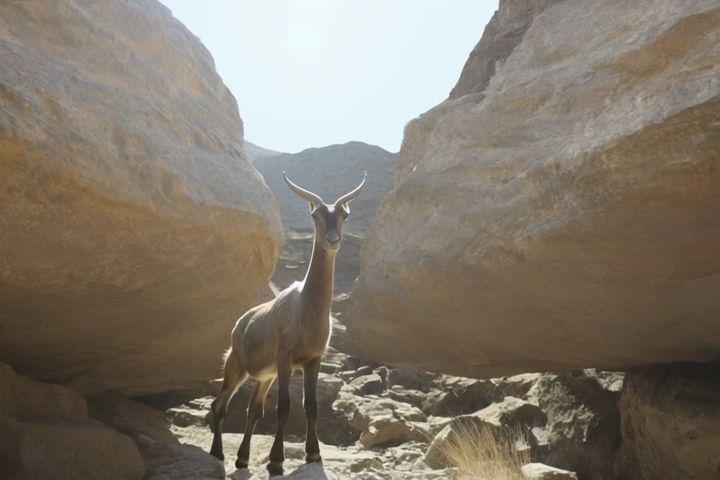}
        \end{subfigure}\hspace{-1mm}
        \begin{subfigure}{.105\linewidth}
            \includegraphics[width=\linewidth]{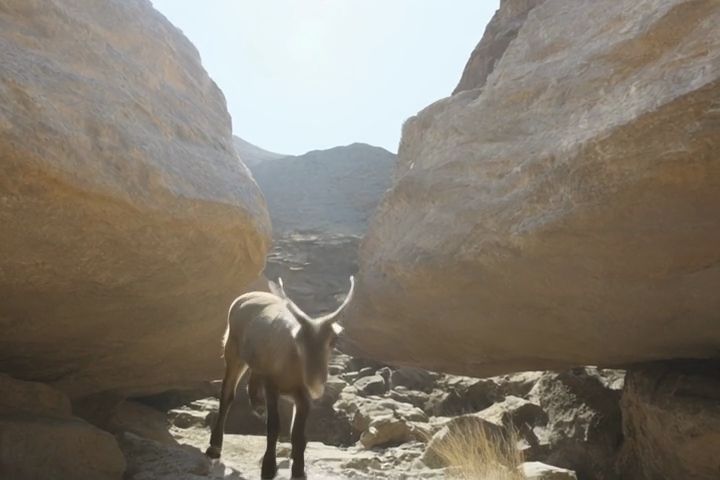}
        \end{subfigure}
    \end{minipage}
    % \vspace{1mm}
    % ===== 第7行 (DeT) =====
    \rotateLabel{DeT}\hspace{-3mm}
    %\hfill
    \begin{minipage}[c]{\rowWidth}
        \centering
        \begin{subfigure}{.105\linewidth}
            \includegraphics[width=\linewidth]{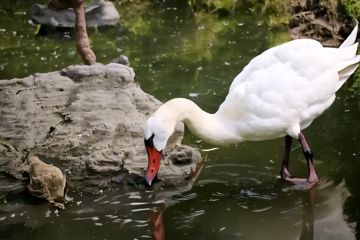}
        \end{subfigure}\hspace{-1mm}
        \begin{subfigure}{.105\linewidth}
            \includegraphics[width=\linewidth]{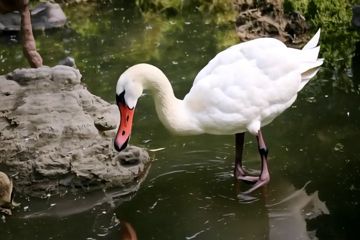}
        \end{subfigure}\hspace{-1mm}
        \begin{subfigure}{.105\linewidth}
            \includegraphics[width=\linewidth]{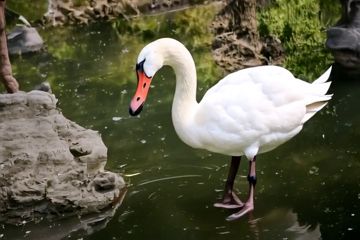}
        \end{subfigure}\hspace{-0.5mm}
        \begin{subfigure}{.105\linewidth}
            \includegraphics[width=\linewidth]{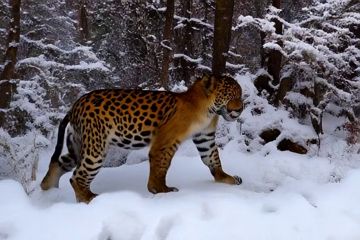}
        \end{subfigure}\hspace{-1mm}
        \begin{subfigure}{.105\linewidth}
            \includegraphics[width=\linewidth]{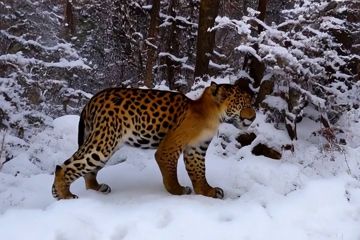}
        \end{subfigure}\hspace{-1mm}
        \begin{subfigure}{.105\linewidth}
            \includegraphics[width=\linewidth]{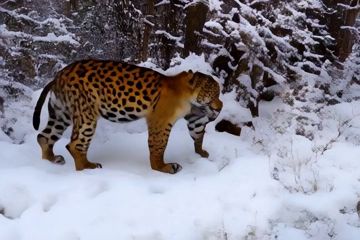}
        \end{subfigure}\hspace{-0.5mm}
        \begin{subfigure}{.105\linewidth}
            \includegraphics[width=\linewidth]{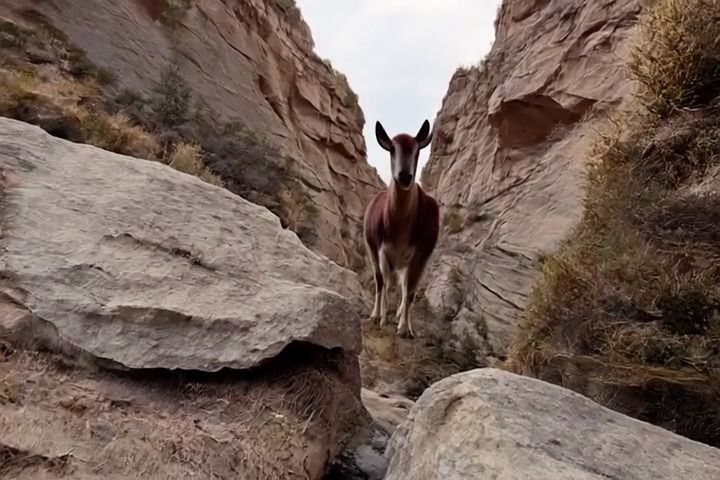}
        \end{subfigure}\hspace{-1mm}
        \begin{subfigure}{.105\linewidth}
            \includegraphics[width=\linewidth]{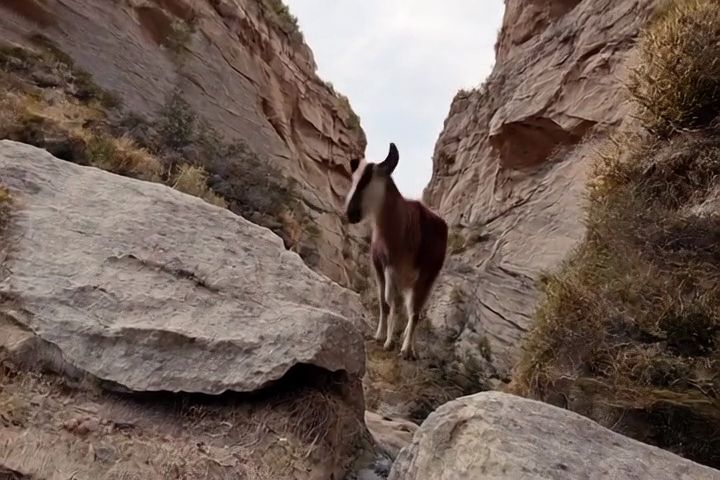}
        \end{subfigure}\hspace{-1mm}
        \begin{subfigure}{.105\linewidth}
            \includegraphics[width=\linewidth]{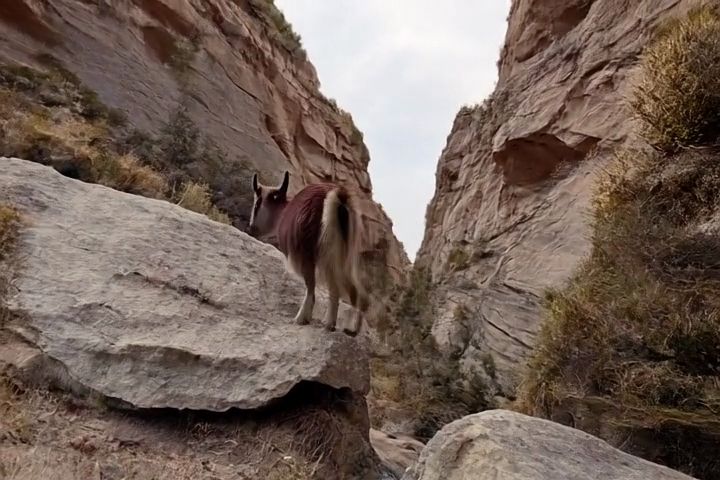}
        \end{subfigure}
    \end{minipage}
    % \vspace{1mm}
    % ===== 第8行 (FastVMT) =====
    \rotateLabel{FastVMT}\hspace{-3mm}
    %\hfill
    \begin{minipage}[c]{\rowWidth}
        \centering
        \begin{subfigure}{.105\linewidth}
            \includegraphics[width=\linewidth]{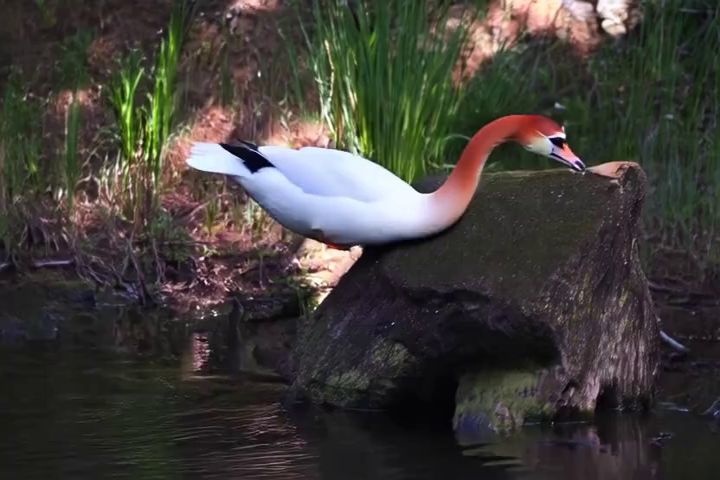}
        \end{subfigure}\hspace{-1mm}
        \begin{subfigure}{.105\linewidth}
            \includegraphics[width=\linewidth]{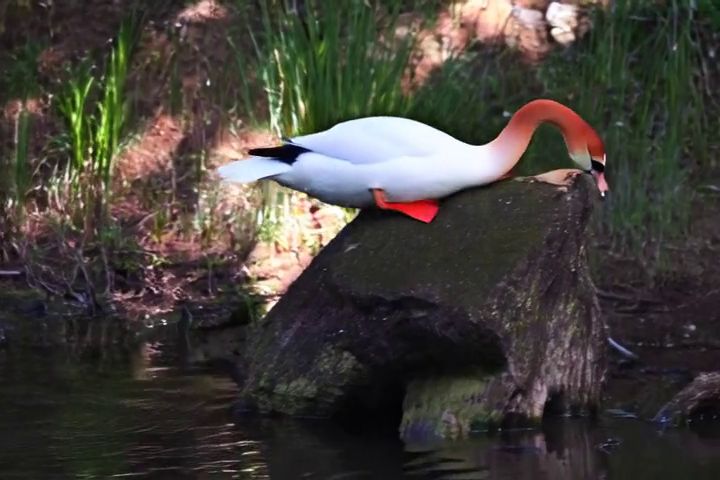}
        \end{subfigure}\hspace{-1mm}
        \begin{subfigure}{.105\linewidth}
            \includegraphics[width=\linewidth]{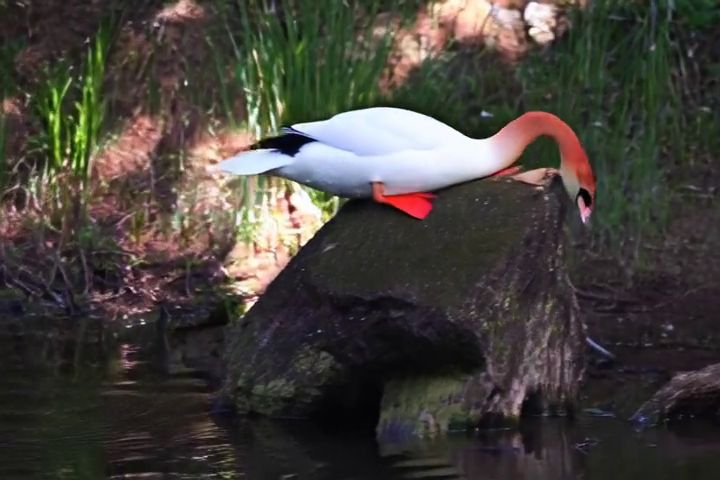}
        \end{subfigure}\hspace{-0.5mm}
        \begin{subfigure}{.105\linewidth}
            \includegraphics[width=\linewidth]{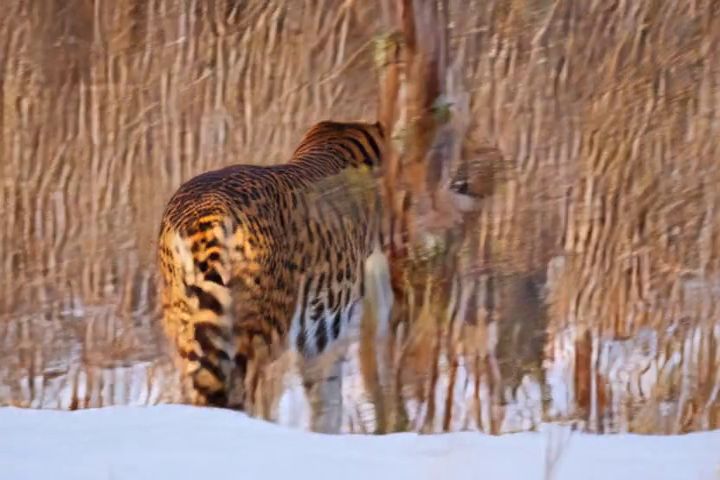}
        \end{subfigure}\hspace{-1mm}
        \begin{subfigure}{.105\linewidth}
            \includegraphics[width=\linewidth]{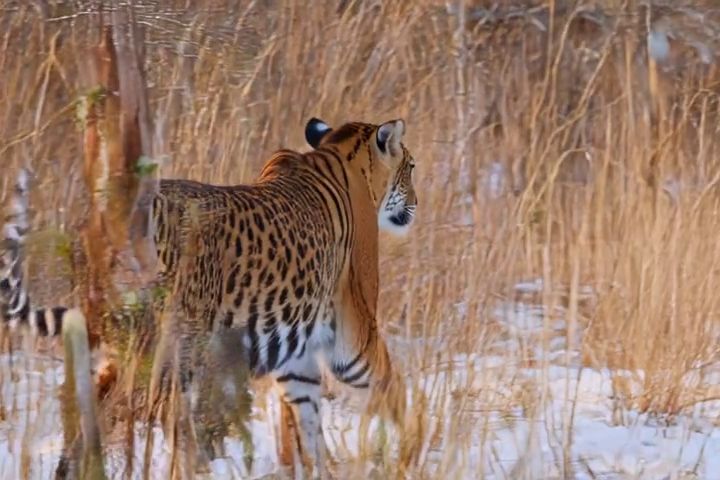}
        \end{subfigure}\hspace{-1mm}
        \begin{subfigure}{.105\linewidth}
            \includegraphics[width=\linewidth]{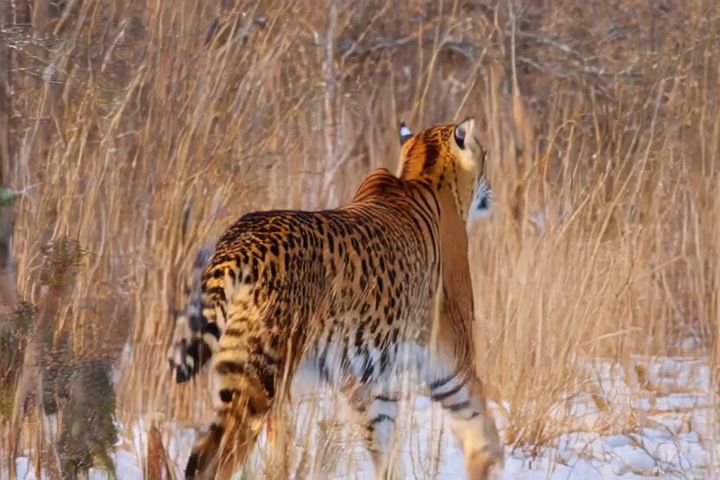}
        \end{subfigure}\hspace{-0.5mm}
        \begin{subfigure}{.105\linewidth}
            \includegraphics[width=\linewidth]{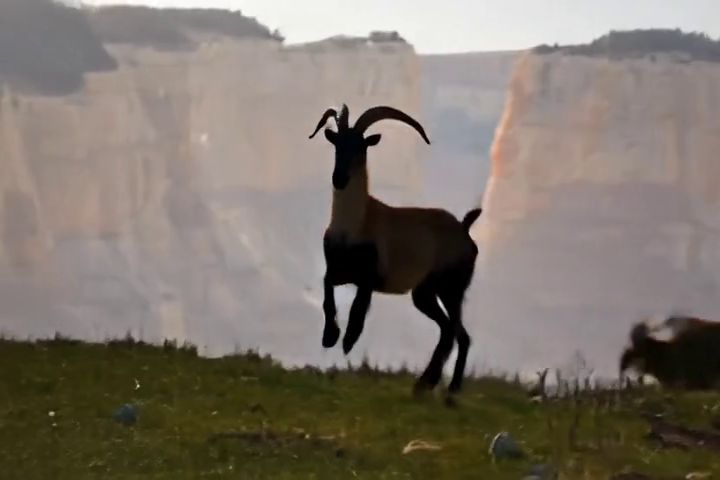}
        \end{subfigure}\hspace{-1mm}
        \begin{subfigure}{.105\linewidth}
            \includegraphics[width=\linewidth]{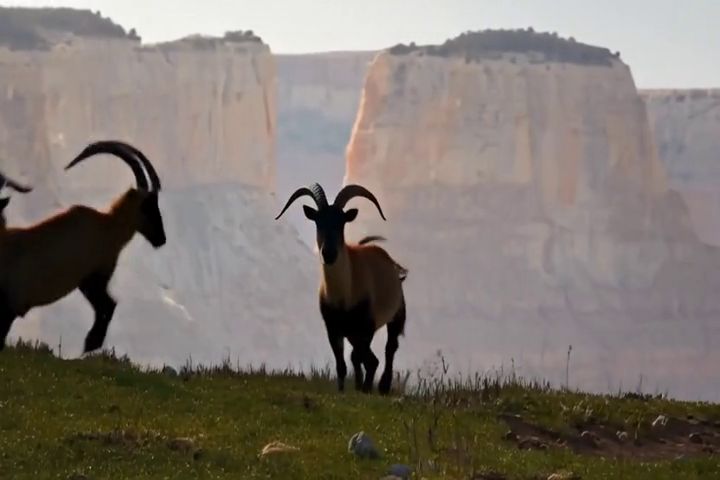}
        \end{subfigure}\hspace{-1mm}
        \begin{subfigure}{.105\linewidth}
            \includegraphics[width=\linewidth]{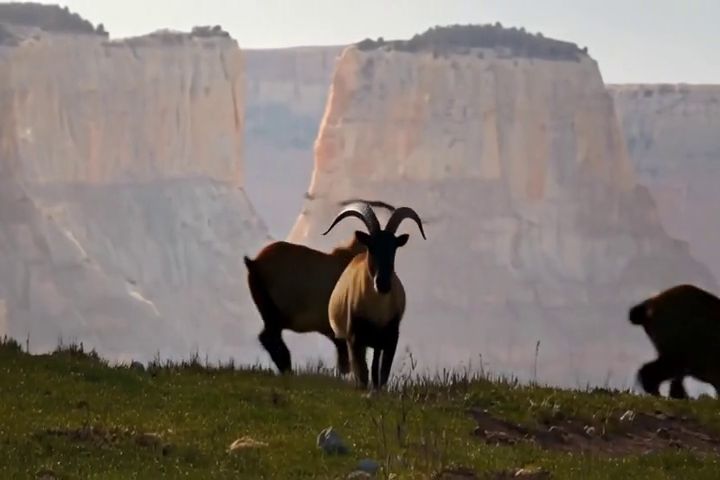}
        \end{subfigure}
    \end{minipage}
    % \vspace{1mm}
    % ===== 第9行 (RoPECraft) =====
    \rotateLabel{RoPEC}\hspace{-3mm}
    %\hfill
    \begin{minipage}[c]{\rowWidth}
        \centering
        \begin{subfigure}{.105\linewidth}
            \includegraphics[width=\linewidth]{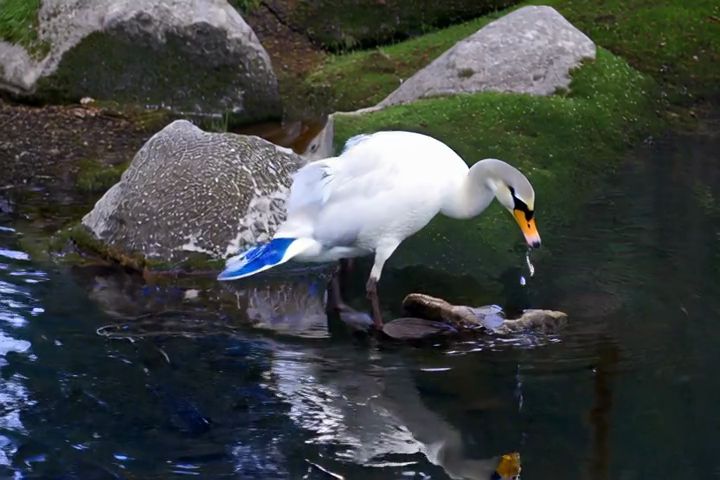}
        \end{subfigure}\hspace{-1mm}
        \begin{subfigure}{.105\linewidth}
            \includegraphics[width=\linewidth]{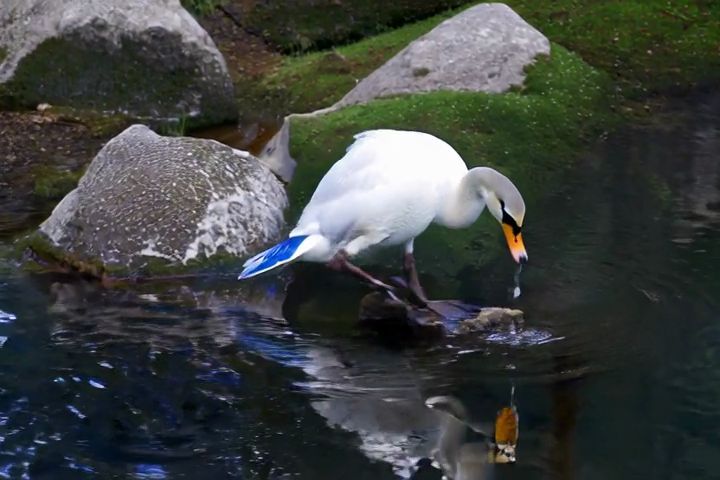}
        \end{subfigure}\hspace{-1mm}
        \begin{subfigure}{.105\linewidth}
            \includegraphics[width=\linewidth]{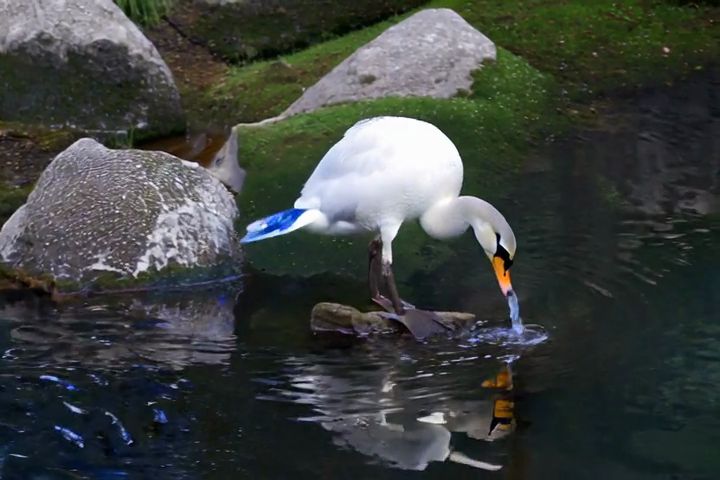}
        \end{subfigure}\hspace{-0.5mm}
        \begin{subfigure}{.105\linewidth}
            \includegraphics[width=\linewidth]{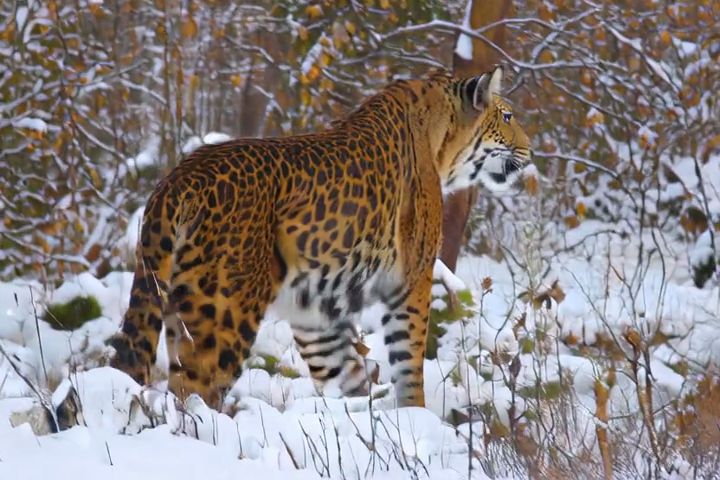}
        \end{subfigure}\hspace{-1mm}
        \begin{subfigure}{.105\linewidth}
            \includegraphics[width=\linewidth]{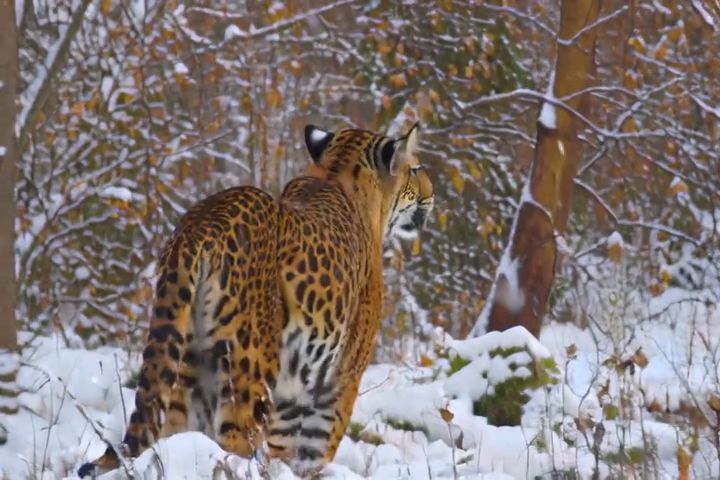}
        \end{subfigure}\hspace{-1mm}
        \begin{subfigure}{.105\linewidth}
            \includegraphics[width=\linewidth]{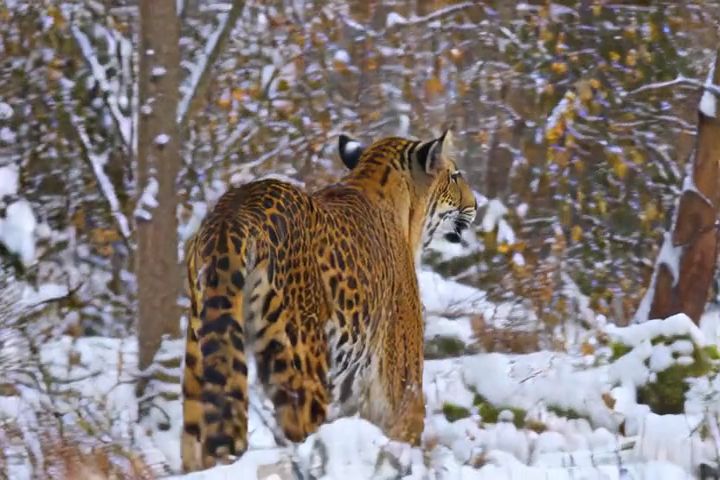}
        \end{subfigure}\hspace{-0.5mm}
        \begin{subfigure}{.105\linewidth}
            \includegraphics[width=\linewidth]{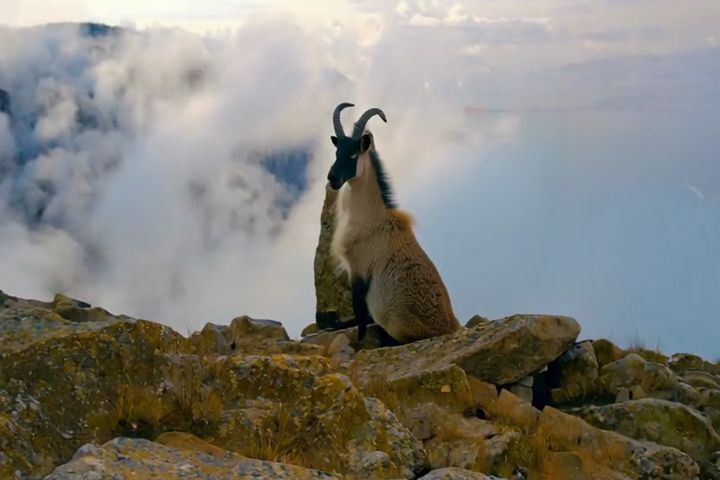}
        \end{subfigure}\hspace{-1mm}
        \begin{subfigure}{.105\linewidth}
            \includegraphics[width=\linewidth]{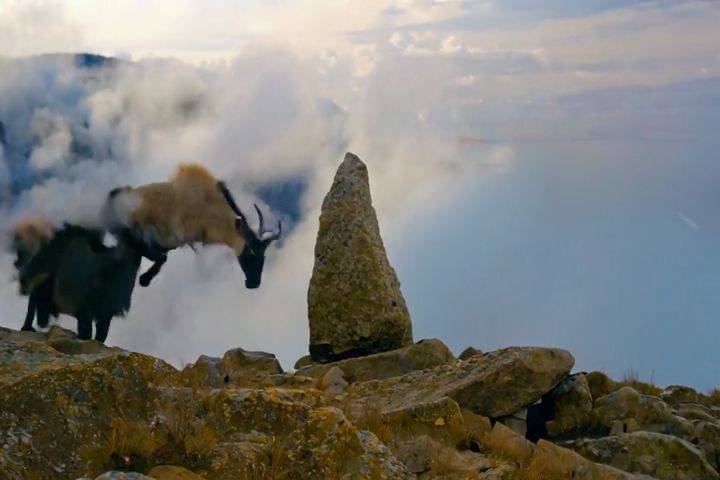}
        \end{subfigure}\hspace{-1mm}
        \begin{subfigure}{.105\linewidth}
            \includegraphics[width=\linewidth]{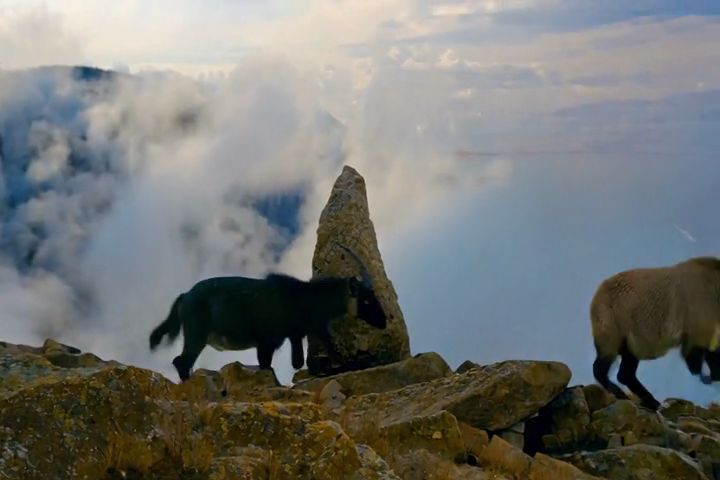}
        \end{subfigure}
    \end{minipage}
    % \vspace{1mm}
    % ===== 第10行 (Ours) =====
    \rotateLabel{Ours}\hspace{-3mm}
    %\hfill
    \begin{minipage}[c]{\rowWidth}
        \centering
        \begin{subfigure}{.105\linewidth}
            \includegraphics[width=\linewidth]{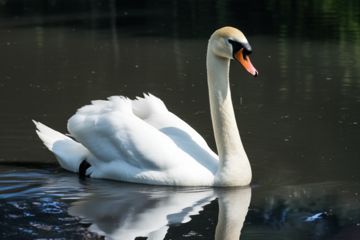}
        \end{subfigure}\hspace{-1mm}
        \begin{subfigure}{.105\linewidth}
            \includegraphics[width=\linewidth]{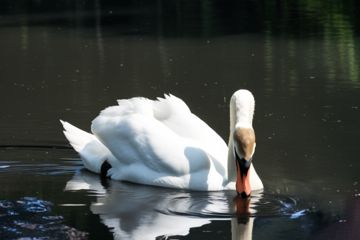}
        \end{subfigure}\hspace{-1mm}
        \begin{subfigure}{.105\linewidth}
            \includegraphics[width=\linewidth]{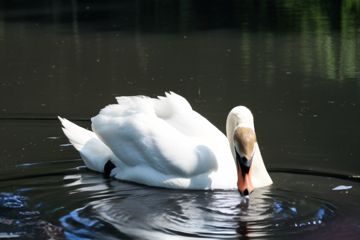}
        \end{subfigure}\hspace{-0.5mm}
        \begin{subfigure}{.105\linewidth}
            \includegraphics[width=\linewidth]{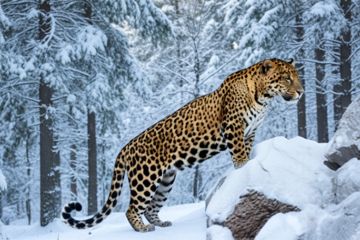}
        \end{subfigure}\hspace{-1mm}
        \begin{subfigure}{.105\linewidth}
            \includegraphics[width=\linewidth]{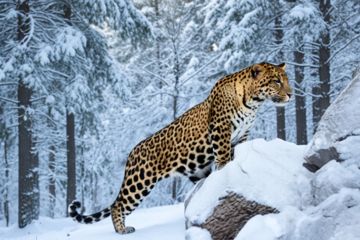}
        \end{subfigure}\hspace{-1mm}
        \begin{subfigure}{.105\linewidth}
            \includegraphics[width=\linewidth]{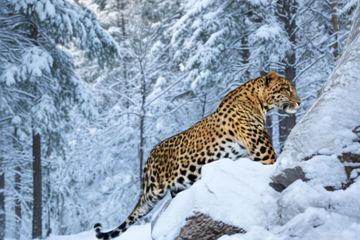}
        \end{subfigure}\hspace{-0.5mm}
        \begin{subfigure}{.105\linewidth}
            \includegraphics[width=\linewidth]{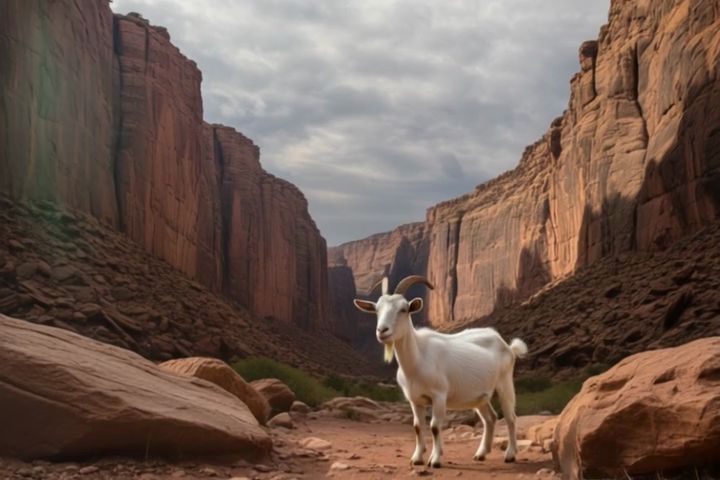}
        \end{subfigure}\hspace{-1mm}
        \begin{subfigure}{.105\linewidth}
            \includegraphics[width=\linewidth]{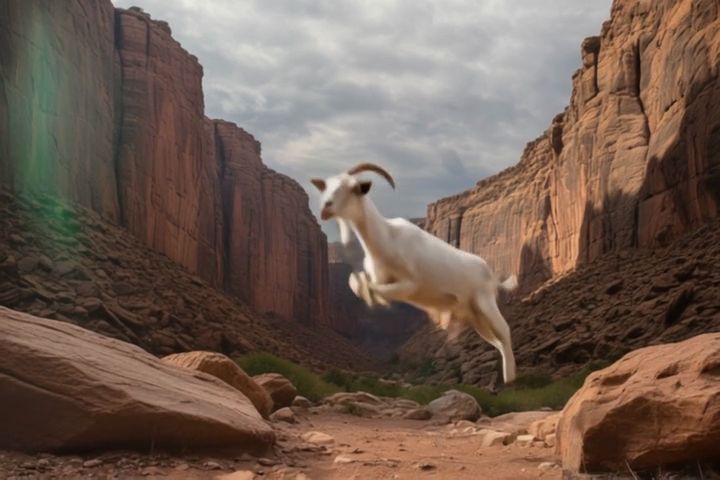}
        \end{subfigure}\hspace{-1mm}
        \begin{subfigure}{.105\linewidth}
            \includegraphics[width=\linewidth]{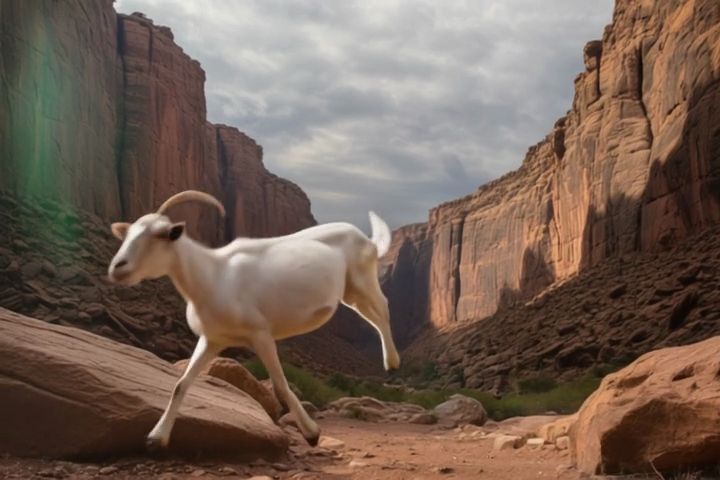}
        \end{subfigure}
    \end{minipage}
    
    % \vspace{-2mm}
    \caption{Qualitative comparison of motion transfer methods. Our \emph{MotionAdapter} enables robust, content-aware motion transfer, producing temporally coherent and semantically aligned videos that preserve both reference motion and target appearance, even under large semantic gaps and complex scenarios.}
    \label{fig:main}
    % \vspace{-5mm}
\end{figure*}

\textbf{Implementation Details.}
We conduct all experiments using PyTorch and adopt the open-source CogVideoX-5B~\cite{yang2025cogvideox} model as our backbone. 
Since CogVideoX-5B supports both T2V and I2V generation, our \emph{MotionAdapter} naturally supports both pipelines. 
In our image-to-video setting, we generate the initial target frame using QwenImage~\cite{wu2025qwen}.
More details of \emph{MotionAdapter} Text-to-Video pipeline are provided in our supplementary.
Following the setup in~\cite{pondaven2025video}, we use 50 denoising steps and apply motion guidance only during the first 20\% of the denoising process, we also chose Top-3 nearest pixels empirically in cross-frame attention motion extraction. 
For DiTFlow~\cite{pondaven2025video}, SMM~\cite{yatim2024space}, and MOFT~\cite{xiao2024video}, we follow DiTFlow that re-implement the methods that configured with the CogVideoX-5B backbone for fair comparison. 
For other baselines, we use backbones same as their official setting.
All remaining baselines are evaluated using their official implementations. For all DiT-based methods, we generate videos at a resolution of 720$\times$480 for 49 frames. For non-DiT methods, we follow the default video resolution and sequence length specified in their original implementations. Note that our \emph{MotionAdapter} is also compitable to other VDMs, one can first analyze the backbone feature as Sec.~\ref{para:disentanglement} to find the best diffusion block for motion extraction and guidance, then apply the proposed cross-frame attention motion extraction, motion customization, and motion refinement modules to perform motion transfer, we provide the on other VDMs in supplementary. 

\begin{figure*}[t]
    \centering
    
    % ================== Case 1 ==================
    \begin{minipage}{0.98\columnwidth}
    \centering
    
    {\scriptsize \textbf{Reference:} \texttt{Rollerblader jump trick on road}}\\
    
    \begin{subfigure}{.32\linewidth}
    \includegraphics[width=\linewidth]{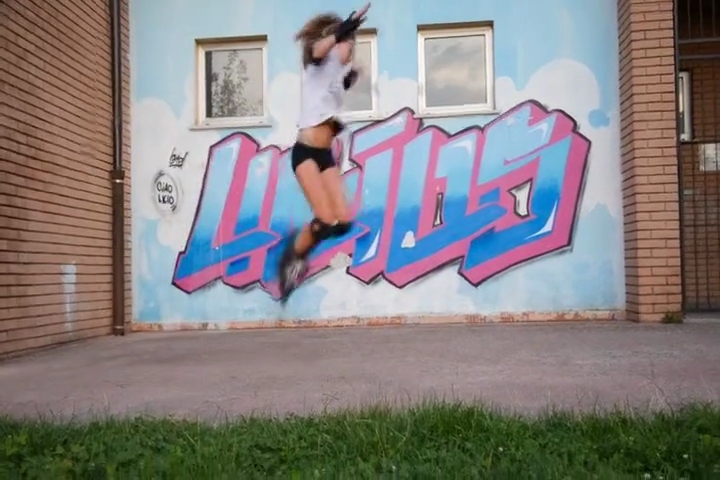}
    \end{subfigure}\hspace{-0.5mm}
    \begin{subfigure}{.32\linewidth}
    \includegraphics[width=\linewidth]{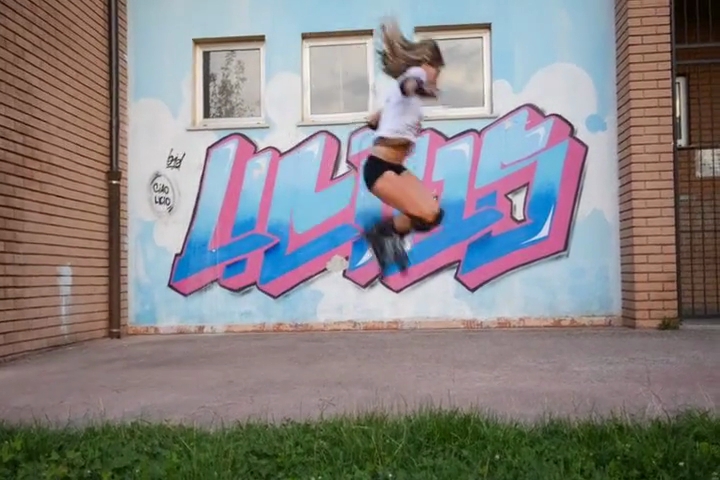}
    \end{subfigure}\hspace{-0.5mm}
    \begin{subfigure}{.32\linewidth}
    \includegraphics[width=\linewidth]{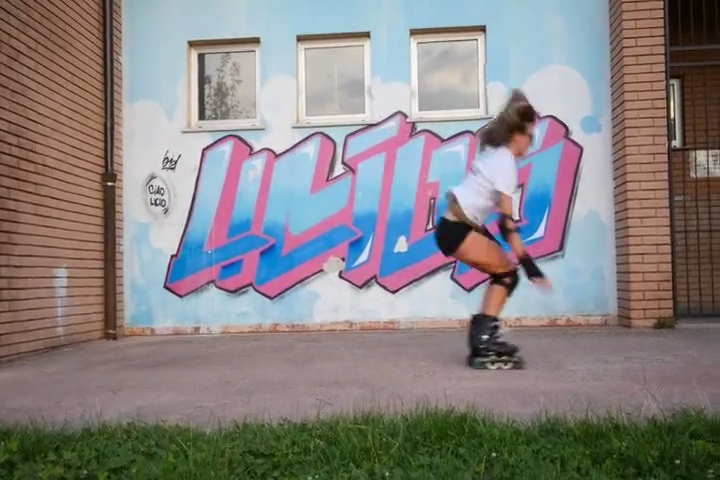}
    \end{subfigure}
    {\scriptsize \texttt{Biker riding past and jump trick in the air}}\\
    \begin{subfigure}{.32\linewidth}
    \includegraphics[width=\linewidth]{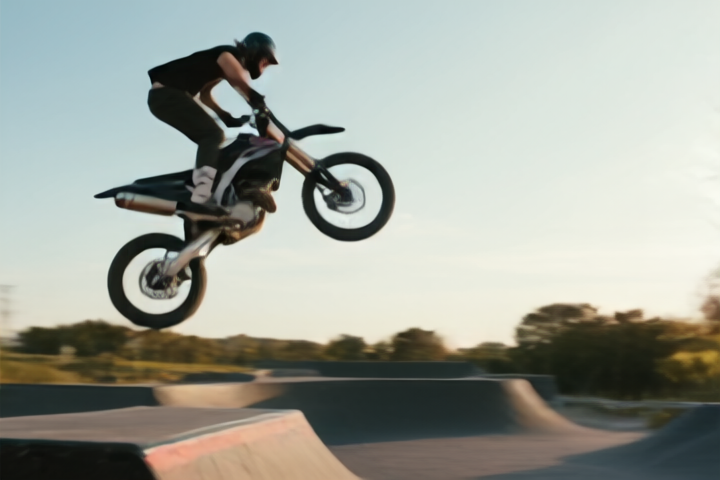}
    \end{subfigure}\hspace{-0.5mm}
    \begin{subfigure}{.32\linewidth}
    \includegraphics[width=\linewidth]{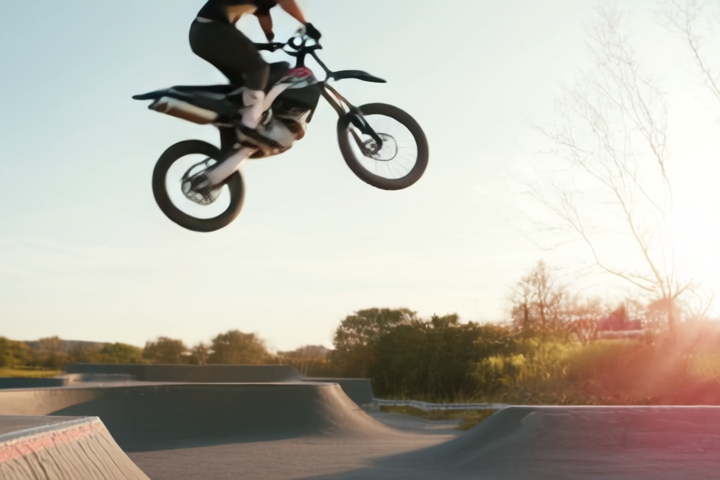}
    \end{subfigure}\hspace{-0.5mm}
    \begin{subfigure}{.32\linewidth}
    \includegraphics[width=\linewidth]{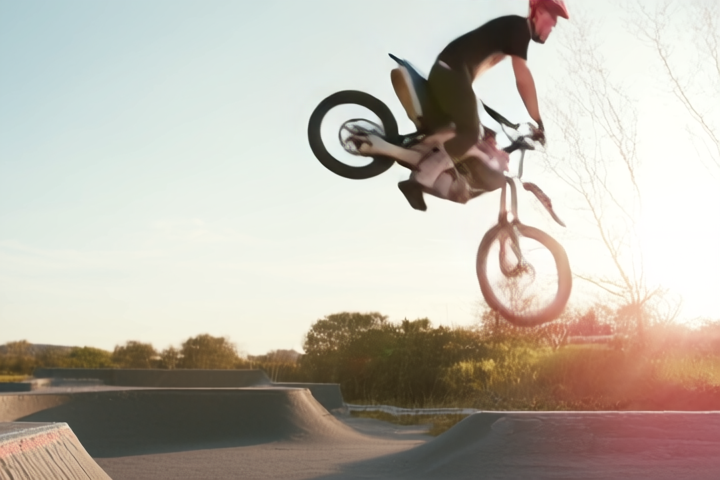}
    \end{subfigure}
    \end{minipage}
    % ================== Case 2 ==================
    \begin{minipage}{0.98\columnwidth}
    \centering
    {\scriptsize \textbf{Reference:} \texttt{Kid riding a bike up a hill}}\\
    \begin{subfigure}{.32\linewidth}
    \includegraphics[width=\linewidth]{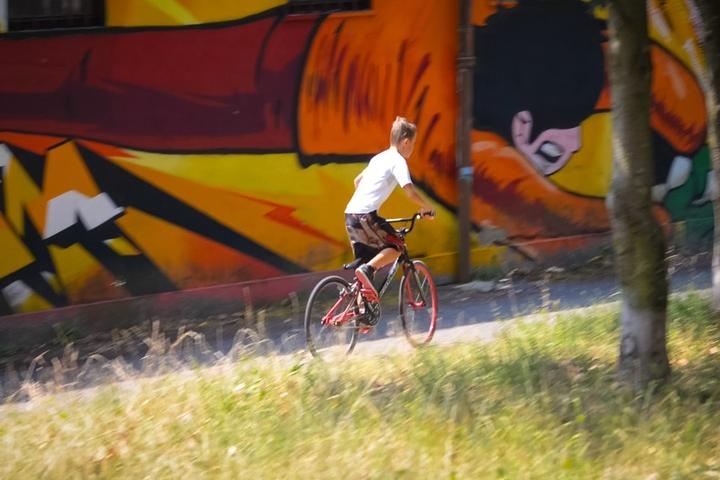}
    \end{subfigure}\hspace{-0.5mm}
    \begin{subfigure}{.32\linewidth}
    \includegraphics[width=\linewidth]{visualize/case2/2.jpg}
    \end{subfigure}\hspace{-0.5mm}
    \begin{subfigure}{.32\linewidth}
    \includegraphics[width=\linewidth]{visualize/case2/3.jpg}
    \end{subfigure}
    
    {\scriptsize \texttt{Leopard running up a snowy hill in a forest}}\\
    \begin{subfigure}{.32\linewidth}
    \includegraphics[width=\linewidth]{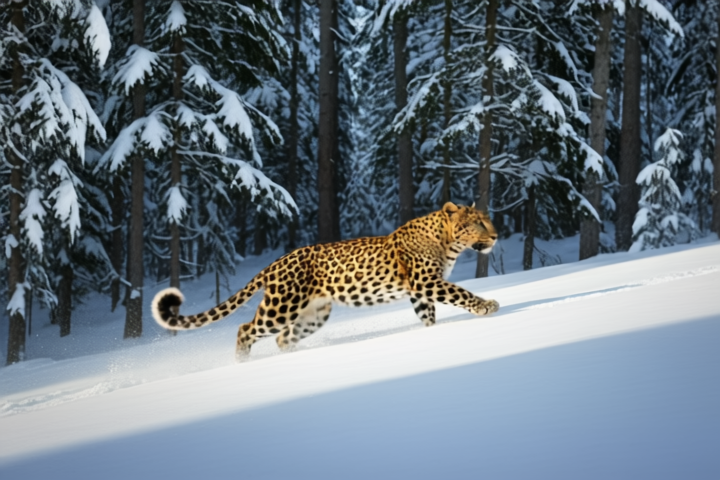}
    \end{subfigure}\hspace{-0.5mm}
    \begin{subfigure}{.32\linewidth}
    \includegraphics[width=\linewidth]{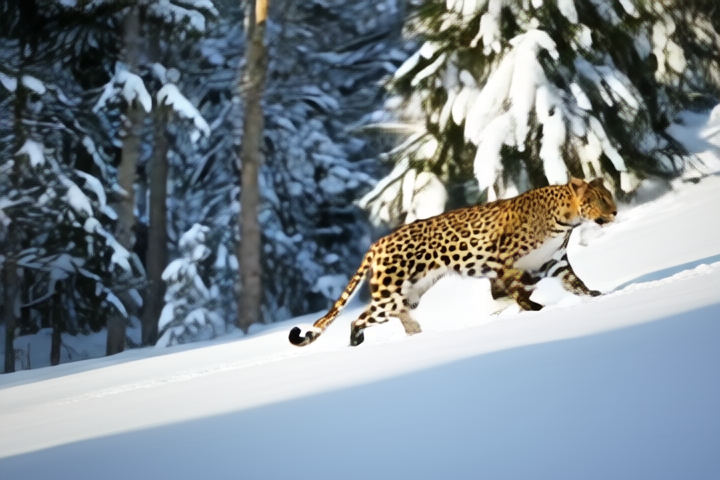}
    \end{subfigure}\hspace{-0.5mm}
    \begin{subfigure}{.32\linewidth}
    \includegraphics[width=\linewidth]{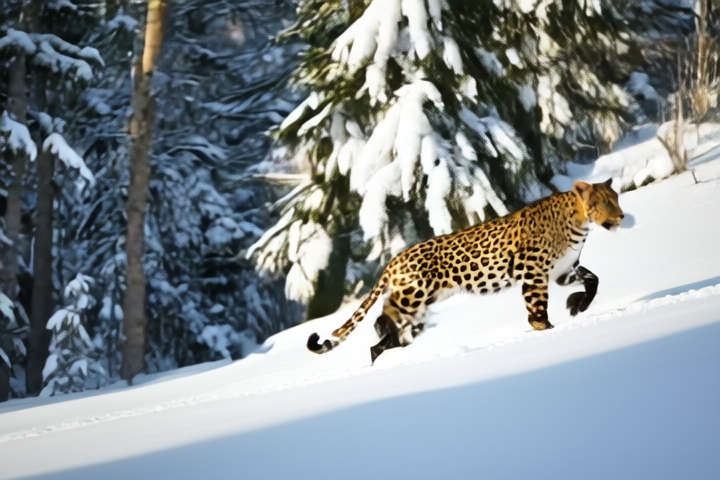}
    \end{subfigure}
    
    \end{minipage}
    
    \vspace{-1mm}
    \caption{
    \emph{MotionAdapter} can perform effective motion transfer even in the presence of large content gaps.
    }
    \label{vis_fig}
    \vspace{-3mm}
\end{figure*}

\subsection{Qualitative comparison}
We present qualitative comparisons between \emph{MotionAdapter} and competing methods in Fig.~\ref{fig:main}. In the first example, the significant shape disparity between the ``\texttt{Flamingo}'' and the ``\texttt{Swan}'' causes existing methods to fail in transferring the flamingo’s motion. As a result, the generated swan does not follow the reference motion or spatial trajectory. With our Motion Customization module, \emph{MotionAdapter} successfully adapts the flamingo’s motion to the swan despite their structural differences.

In the second example, the ``\texttt{Goat}'' in the reference video occupies only a small region, making its motion difficult to capture and transfer to the ``\texttt{Jaguar}''. MotionClone and MotionInversion produce low-quality generations, while other baselines fail to reproduce the intended motion. In contrast, \emph{MotionAdapter} effectively accounts for the structural gap and accurately transfers the motion. 
{
The third example illustrates a challenging case involving complex foreground and background dynamics. The jumping ``\texttt{Dog}'' introduces strong motion signals that degrade the performance of MotionClone, MotionInversion, and DeT. By decoupling and separately handling foreground and background motion, \emph{MotionAdapter} avoid the interference of the person and robustly transfers complex motions while maintaining prompt consistency.
}

\begin{figure*}[t]
\centering
% \vspace{-9mm}

% ================== Left Column (3 items) ==================
\begin{minipage}{0.98\columnwidth}
\centering

{\scriptsize \textbf{Reference 1:} \texttt{Woman walking in a park}}\\
\begin{subfigure}{.32\linewidth}
\includegraphics[width=\linewidth]{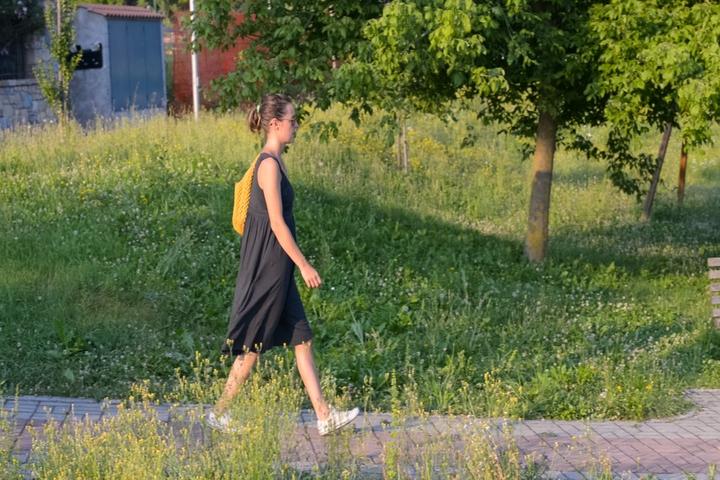}
\end{subfigure}\hspace{-0.5mm}
\begin{subfigure}{.32\linewidth}
\includegraphics[width=\linewidth]{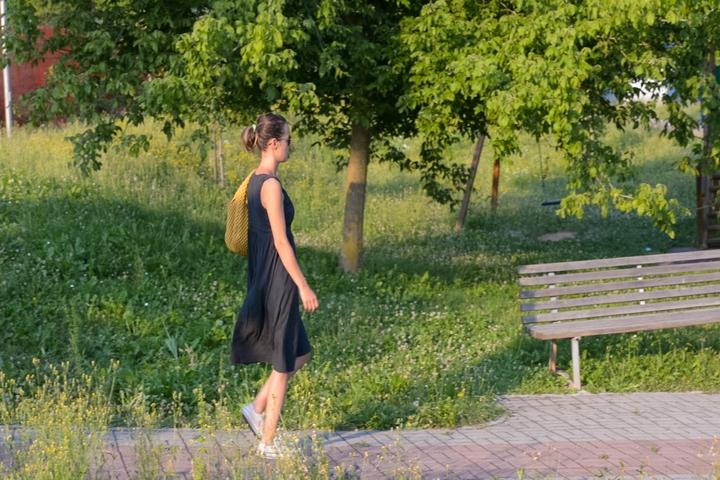}
\end{subfigure}\hspace{-0.5mm}
\begin{subfigure}{.32\linewidth}
\includegraphics[width=\linewidth]{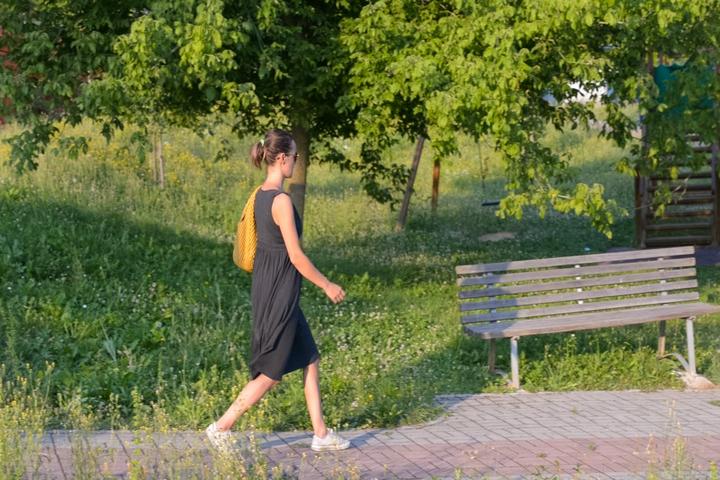}
\end{subfigure}
{\scriptsize \textbf{Zoom In:} \texttt{Robot walking in a park}}\\
\begin{subfigure}{.32\linewidth}
\includegraphics[width=\linewidth]{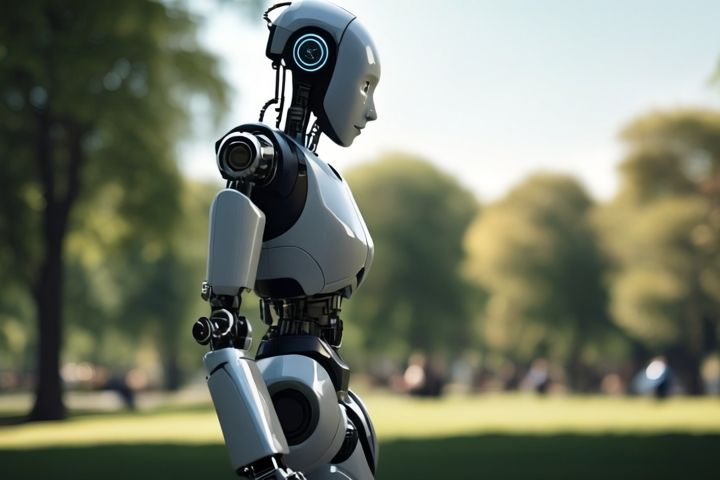}
\end{subfigure}\hspace{-0.5mm}
\begin{subfigure}{.32\linewidth}
\includegraphics[width=\linewidth]{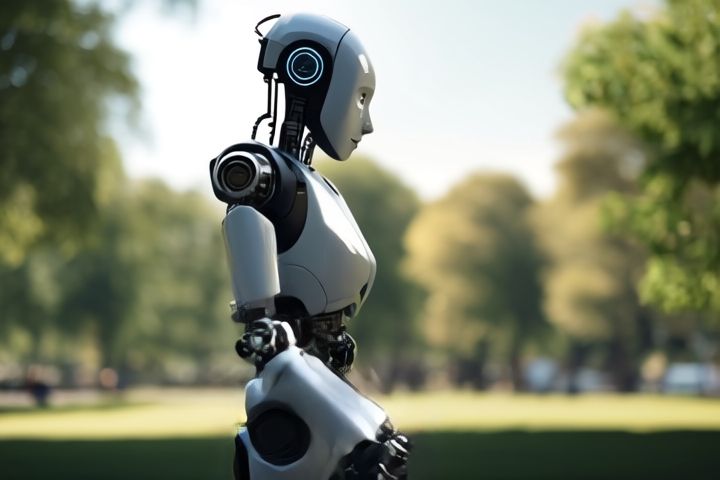}
\end{subfigure}\hspace{-0.5mm}
\begin{subfigure}{.32\linewidth}
\includegraphics[width=\linewidth]{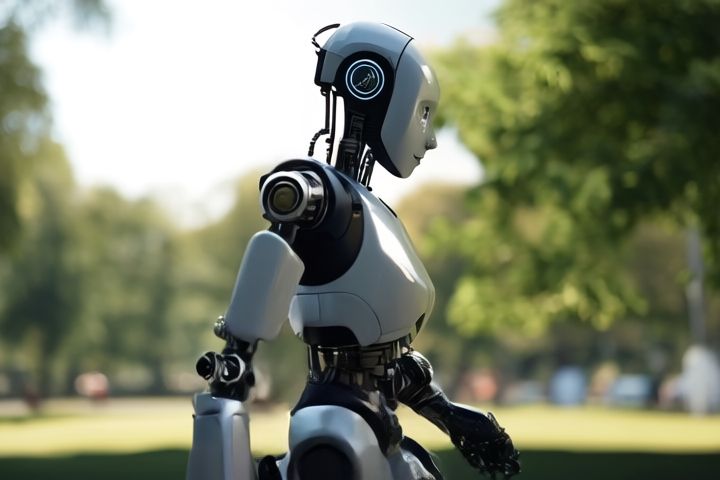}
\end{subfigure}
{\scriptsize \textbf{Zoom Out:} \texttt{Robot walking in a park}}\\
\begin{subfigure}{.32\linewidth}
\includegraphics[width=\linewidth]{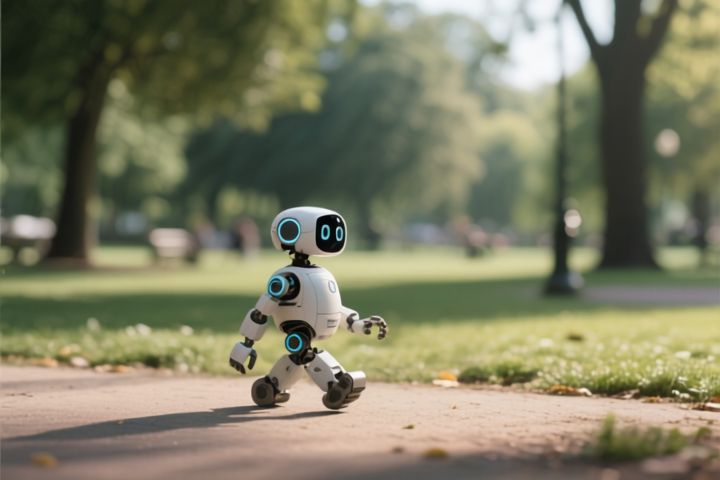}
\end{subfigure}\hspace{-0.5mm}
\begin{subfigure}{.32\linewidth}
\includegraphics[width=\linewidth]{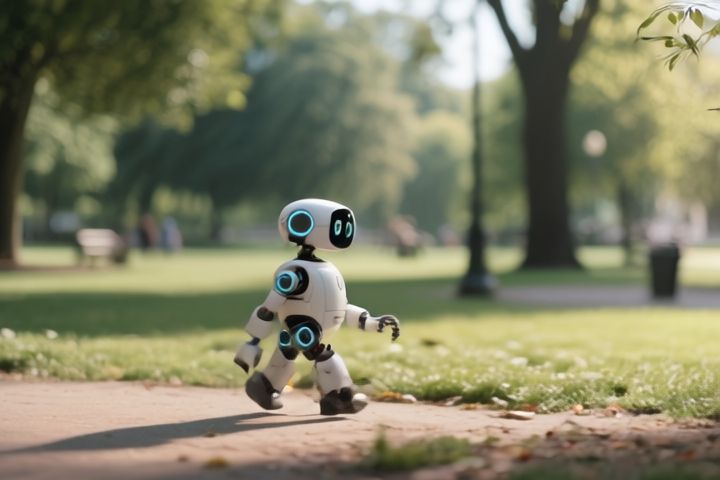}
\end{subfigure}\hspace{-0.5mm}
\begin{subfigure}{.32\linewidth}
\includegraphics[width=\linewidth]{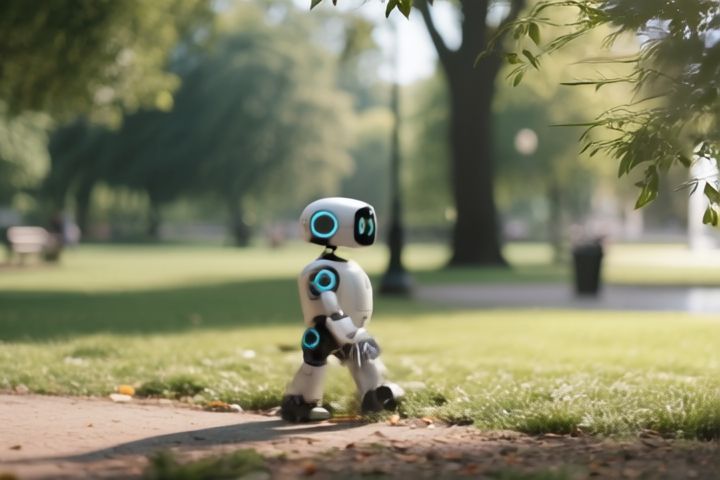}
\end{subfigure}
\end{minipage}
% \hfill % 使用 \hfill 将左右两栏撑开
% ================== Right Column (3 items) ==================
\begin{minipage}{0.98\columnwidth}
\centering
{\scriptsize \textbf{Reference 2:} \texttt{Car driving around a roundabout}}\\

\begin{subfigure}{.32\linewidth}
\includegraphics[width=\linewidth]{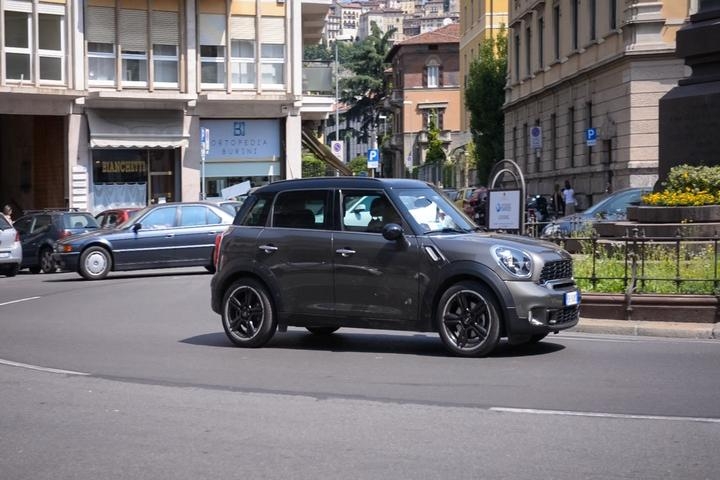}
\end{subfigure}\hspace{-0.5mm}
\begin{subfigure}{.32\linewidth}
\includegraphics[width=\linewidth]{visualize/car/2}
\end{subfigure}\hspace{-0.5mm}
\begin{subfigure}{.32\linewidth}
\includegraphics[width=\linewidth]{visualize/car/3}
\end{subfigure}
% \vspace{1mm}
{\scriptsize \textbf{Case 1:} \texttt{Woman} (\textbf{R.1}) \texttt{walking with dog} (\textbf{R.2})}\\
\begin{subfigure}{.32\linewidth}
\includegraphics[width=\linewidth]{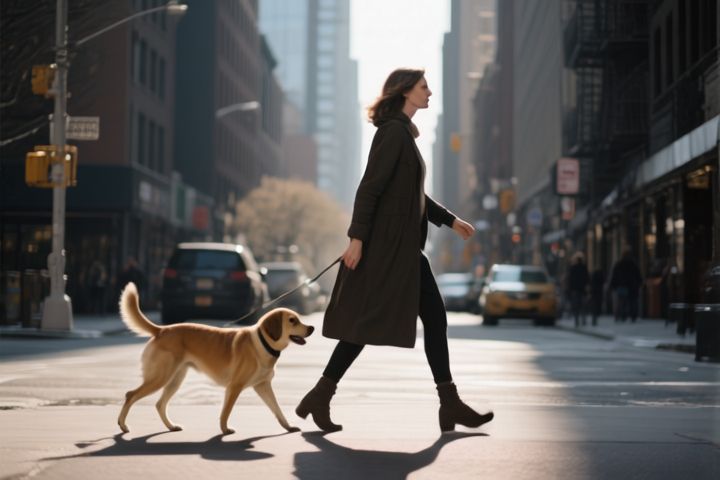}
\end{subfigure}\hspace{-0.5mm}
\begin{subfigure}{.32\linewidth}
\includegraphics[width=\linewidth]{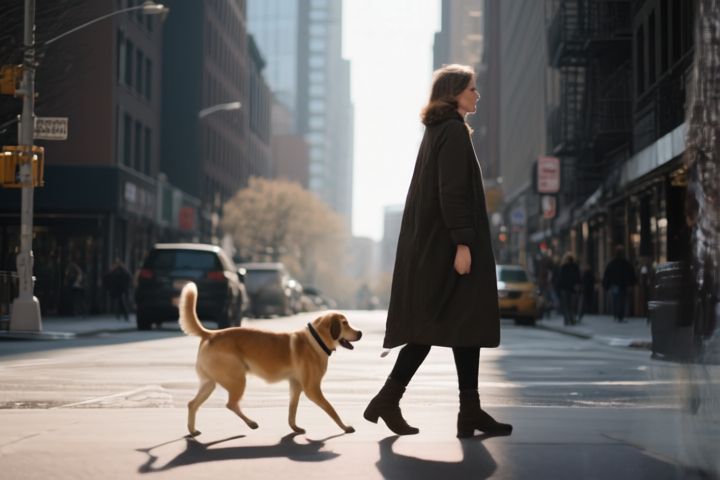}
\end{subfigure}\hspace{-0.5mm}
\begin{subfigure}{.32\linewidth}
\includegraphics[width=\linewidth]{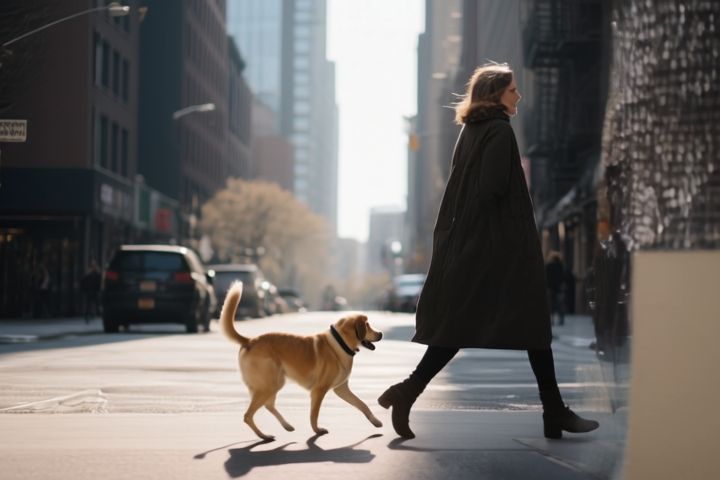}
\end{subfigure}
% \vspace{1mm}
% 注意：这组图片在你给的原代码中重复了一次，我保留了它。如果是不小心复制多了，可以直接删掉这块。
{\scriptsize \textbf{Case 2:} \texttt{Woman} (\textbf{R.2}) \texttt{walking with dog} (\textbf{R.1})}\\
\begin{subfigure}{.32\linewidth}
\includegraphics[width=\linewidth]{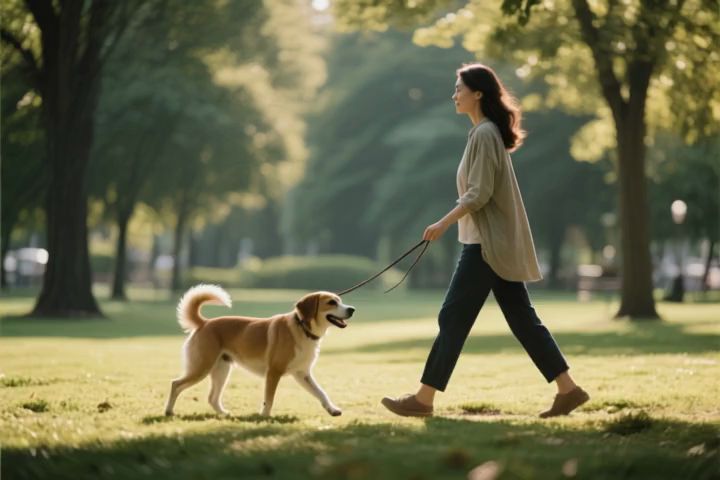}
\end{subfigure}\hspace{-0.5mm}
\begin{subfigure}{.32\linewidth}
\includegraphics[width=\linewidth]{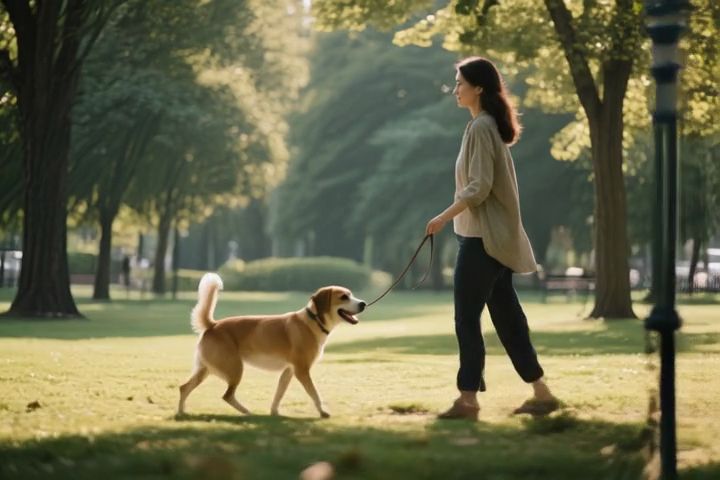}
\end{subfigure}\hspace{-0.5mm}
\begin{subfigure}{.32\linewidth}
\includegraphics[width=\linewidth]{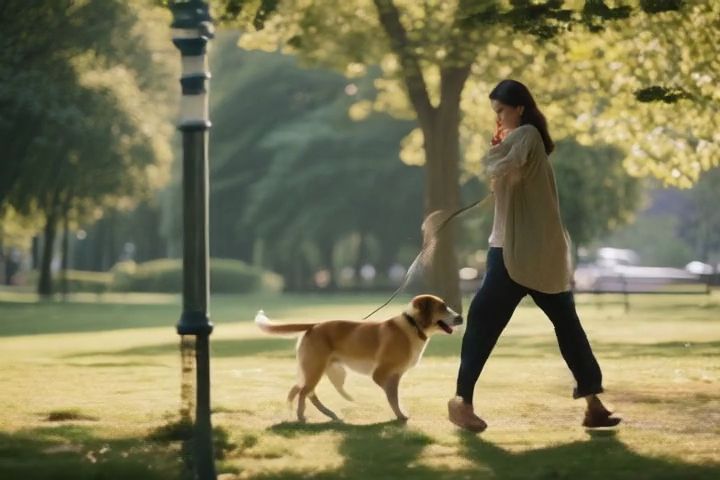}
\end{subfigure}
\end{minipage}
\vspace{-1mm}
\caption{\emph{MotionAdapter} supports various motion editings, such as motion zoom in/out and motion composition.}
\label{fig:zoom_in_out}
\vspace{-3mm}
\end{figure*}

We further providing more challenge results in Fig.~\ref{vis_fig}. In the first example, the ``\texttt{rollerblader}'' in the reference video performs a jump–landing sequence with complex, motion. Despite the highly nonstationary motion, \emph{MotionAdapter} customizes the reference motion and robustly customize it to the ``\texttt{Biker}''. In the second example, the reference video shows a  ``\texttt{Boy}'' has been partially occluded. By refining motion cues under occlusion, \emph{MotionAdapter} successfully refined the reference motion and transfers them to the target ``\texttt{Leopard}''. 

\noindent{\textbf{Motion edit.}} \emph{MotionAdapter} supports various motion edit such as zoom-in/out and mixed motion. As shown left columns in Fig.~\ref{fig:zoom_in_out}, by applying zoom-in/zoom-out to the reference motion of ``\texttt{Woman}'', \emph{MotionAdapter} achieves controllable target object ``\texttt{Dog}'' scaling, while preserving the intended motion. 

Additionally, it also support motion transfer from multiple reference videos with multiple subjects. As shown in right columns of Fig.~\ref{fig:zoom_in_out}, the motion of ``\texttt{Dog}'' and ``\texttt{Woman}'' can be delivered from videos \textbf{Reference 1} or \textbf{Reference 2}, and our \emph{MotionAdapter} effectively transfers the motion of each subject from the reference video to the target video, while maintaining semantic alignment and temporal coherence.

\subsection{Quantitative comparison}

Table~\ref{table:main} reports the quantitative comparison between our approach and several baseline methods. Both versions of our framework (Ours-I2V and Ours-T2V) outperform the majority of competitors across three key metrics at various difficulty levels. This demonstrates that our MotionAdapter is robust and generalizes well to diverse motion complexities.

Our method achieves the highest performance in CLIP Score (CS) and Temporal Consistency (TC), indicating that videos generated by MotionAdapter effectively adhere to target prompts while maintaining high temporal coherence. Interestingly, the I2V version of our method exhibits lower Motion Fidelity (MF) compared to the T2V version. We attribute this to the stochastic nature of the initial target frame, since target frames generated by QwenImage often exhibit significant size discrepancies between reference and target objects, the model requires extensive customization, which inherently reduces performance on Motion Fidelity metric.

Regarding inference efficiency, MotionClone utilizes a lightweight U-Net architecture, requires the least inference time. However, it necessitates significant computational overhead during training. While FastVMT prioritizes accelerating motion transfer, it suffers from diminished video quality and lower motion consistency. Although our pipeline incorporates SAM and DINO computations, it does not significantly increase total processing time compared to related works. Among DiT-based motion transfer frameworks, our method maintains a comparable inference cost to SMM, MOFT, and DiTFlow.

\begin{table*}[t]
    \centering
    % \vspace{-10mm}
    \caption{Quantitative comparison with existing methods on the DAVIS-based dataset. 
    The values of \textbf{CS} and \textbf{TC} are multiplied by $10^4$ for better readability.
    The \textbf{bold} values indicate the best performance. \textbf{TF} means whether the method is training-free.
    MInv is MotionInversion, MClone is MotionClone, and RoPEC is RoPECraft.
    Inference time is measured on a single NVIDIA RTX6000Ada GPU to generate a single video.
    } %and the \textcolor{black!40}{grey} value means the time cost of a U-Net based method.
    \label{table:main}
    \vspace{-0.5em}
    % \small
    % \foot\ding{55}tesize
    \fontsize{7.5}{9}\selectfont
    \scriptsize
    % \tiny
    % \setlength{\tabcolsep}{1.1pt}
    \resizebox{\linewidth}{!}{
    \begin{tabular}{c|c|c|ccc|ccc|ccc|ccc|c}
        \toprule
        \toprule
        \multirow{2}{*}{\textbf{Method}}
        & \multirow{2}{*}{\makecell{\textbf{TF} \\ \textbf{?}}}
        & \multirow{2}{*}{\makecell{\textbf{Struc-}\\ \textbf{ture}}}
        & \multicolumn{3}{c|}{\textbf{Easy}}
        & \multicolumn{3}{c|}{\textbf{Medium}}
        & \multicolumn{3}{c|}{\textbf{Hard}}
        & \multicolumn{3}{c|}{\textbf{ALL}}
        & \multirow{2}{*}{\makecell{\fontsize{6}{7.5}\selectfont \textbf{Infer.}
         \\ \fontsize{6}{7.5}\selectfont \textbf{Time$\downarrow$} \\ \fontsize{6}{7.5}\selectfont \textbf{(mins)}}} \\
        \cmidrule(lr){4-6} 
        \cmidrule(lr){7-9} 
        \cmidrule(lr){10-12} 
        \cmidrule(lr){13-15}
        &  &
        & \textbf{CS$\uparrow$} & \textbf{TC$\uparrow$} & \textbf{MF$\uparrow$}
        & \textbf{CS$\uparrow$} & \textbf{TC$\uparrow$} & \textbf{MF$\uparrow$}
        & \textbf{CS$\uparrow$} & \textbf{TC$\uparrow$} & \textbf{MF$\uparrow$}
        & \textbf{CS$\uparrow$} & \textbf{TC$\uparrow$} & \textbf{MF$\uparrow$}
        & \\
        \midrule
        
        % SMM
        % &  \textcolor{green}{\ding{51}}
        % & 3202 & 9057 &  
        % & 3244 & 8921 &  
        % & 3186 & 9048 &  
        % & 3210 & 9009 &
        % &  \\
        SMM % ditflow official CS&MF
        &  \textcolor{green}{\ding{51}}
        & DiT
        & 3130 & 9057 & 0.782
        & 3170 & 8921 & 0.741 
        & 3160 & 9048 & 0.776
        & 3150 & 9009 & 0.766
        & 10.3 \\
        
        % MOFT     
        % &  \textcolor{green}{\ding{51}}
        % & 3188 & 8823 &  
        % & 3228 & 8952 &  
        % & 3218 & 8864 &  
        % & 3211 & 8880 &
        % &  \\
        MOFT % ditflow official CS&MF
        & \textcolor{green}{\ding{51}}
        & DiT
        & 3130 & 8823 & 0.728
        & 3210 & 8952 & 0.728
        & 3190 & 8864 & 0.722
        & 3180 & 8880 & 0.726
        & 10.3 \\
        
        % \makecell{ \fontsize{6}{7.5}\selectfont MotionInversion}     
        \makecell{ MInv}                
        & \textcolor{red}{\ding{55}}
        & UNet
        & \textbf{3214} & 8634 & 0.724 
        & 3244 & 8493 &  0.722
        & 3179 & 8415 &  0.728
        & 3212 & 8515 &  0.725
        & 1.6 \\ % \textcolor{black!40}{1.6}
        
        % MotionClone
        MClone
        & \textcolor{green}{\ding{51}}
        & UNet
        & 3040 & 7500 &  0.650
        & 3052 & 7445 &  0.650
        & 3012 & 7361 &  0.637
        & 3035 & 7436 &  0.646
        & \textbf{1.0} \\ % \textcolor{black!40}{1.0}
        
        % DiTFlow
        % &  \textcolor{green}{\ding{51}}
        % & 3183 & 9151 &  
        % & 3243 & 9064 &  
        % & 3224 & 9094 &  
        % & 3217 & 9103 &
        % & 10.2 \\
        % DiTFlow % ditflow official CS&MF
        \makecell{ \fontsize{6}{7.5}\selectfont DiTFlow} % ditflow official CS&MF
        & \textcolor{green}{\ding{51}}
        & DiT
        & 3160 & 9151 & 0.790
        & 3210 & 9064 & 0.775
        & 3190 & 9094 & 0.789
        & 3190 & 9103 & 0.785
        & 10.4 \\
        
        DeT
        & \textcolor{red}{\ding{55}}
        & DiT
        & 3149 & 9062 & 0.781
        & 3225 & 9008 & 0.766
        & 3257 & 8978 & 0.754
        & 3210 & 9017 & 0.767
        & 6.1 \\
        
        \makecell{ \fontsize{6}{7.5}\selectfont FastVMT }
        & \textcolor{green}{\ding{51}}
        & DiT
        & 3072 & 8778 &  0.690
        & 2993 & 8746 &  0.664
        & 2971 & 8667 &  0.672
        & 3012 & 8731 &  0.675
        & 2.1 \\
        
        % RoPECraft
        RoPEC
        & \textcolor{green}{\ding{51}}
        & DiT
        & 3112 & 8789 &  0.760
        & 3023 & 8623 &  0.761
        & 3010 & 8530 &  0.723
        & 3049 & 8654 &  0.748
        & 27.0 \\
        \midrule

        % \makecell{\emph{MotionAdapter} \\ (T2V) }
        % \makecell{\emph{MotionAdapter} \\ (I2V) }
        \makecell{\fontsize{6}{7.5}\selectfont {Ours}-I2V }
        &  \textcolor{green}{\ding{51}}
        & DiT
        & 3153 & 9297 &  0.665
        & 3247 & \textbf{9346} &  0.626
        & \textbf{3272} & 9333 &  0.632
        & 3225 & \textbf{9325} &  0.641
        & 10.4 \\

        \makecell{\fontsize{5.5}{7.5}\selectfont {Ours}-T2V }
        &  \textcolor{green}{\ding{51}}
        & DiT
        & 3191 & \textbf{9336} & \textbf{0.803}
        & \textbf{3259} & 9254 & \textbf{0.789}
        & 3270 & \textbf{9380} & \textbf{0.790}
        & \textbf{3240} & 9323 & \textbf{0.794}
        & 14.5 \\
        \bottomrule
        \bottomrule
        \end{tabular}
    }
\vspace{-3mm}
\end{table*}

\textbf{User Study.}
To further validate the motion transfer performance, we conduct a user study to assess the quality of motion transfer results. 
\begin{figure}
\centering
% \vspace{-8mm}
\begin{minipage}{\columnwidth} % 图片区域
    % ===== 第1行 =====
    \parbox{\linewidth}{
        \centering
        % {\scriptsize \texttt{\textbf{Reference Video:} Rollerblader doing jump trick on road}}\\
        \begin{subfigure}{.48\columnwidth}
            \centering
            \includegraphics[width=\columnwidth]{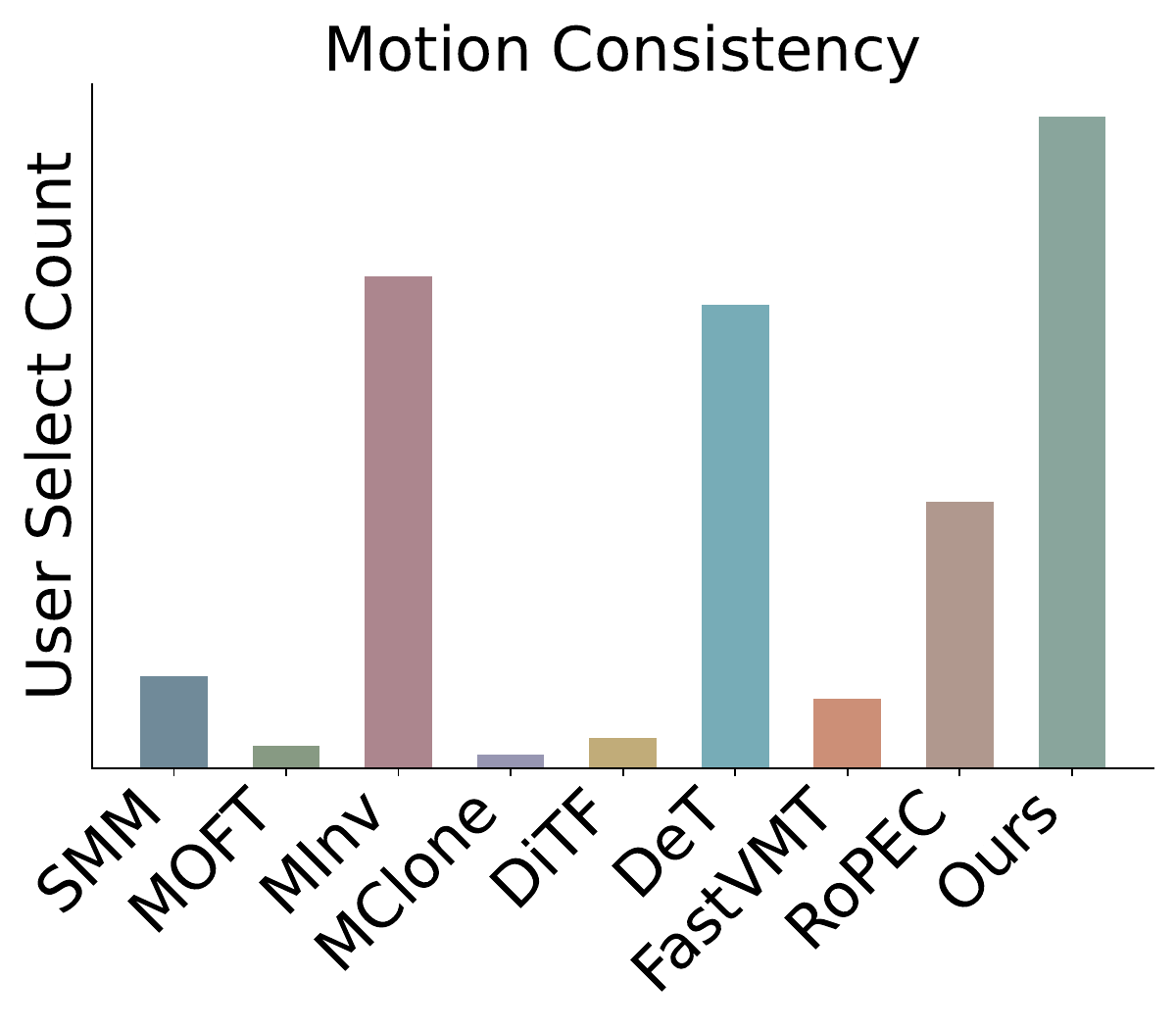}
        \end{subfigure}\hspace{-0.5mm}
        \begin{subfigure}{.48\columnwidth}
            \centering
            \includegraphics[width=\columnwidth]{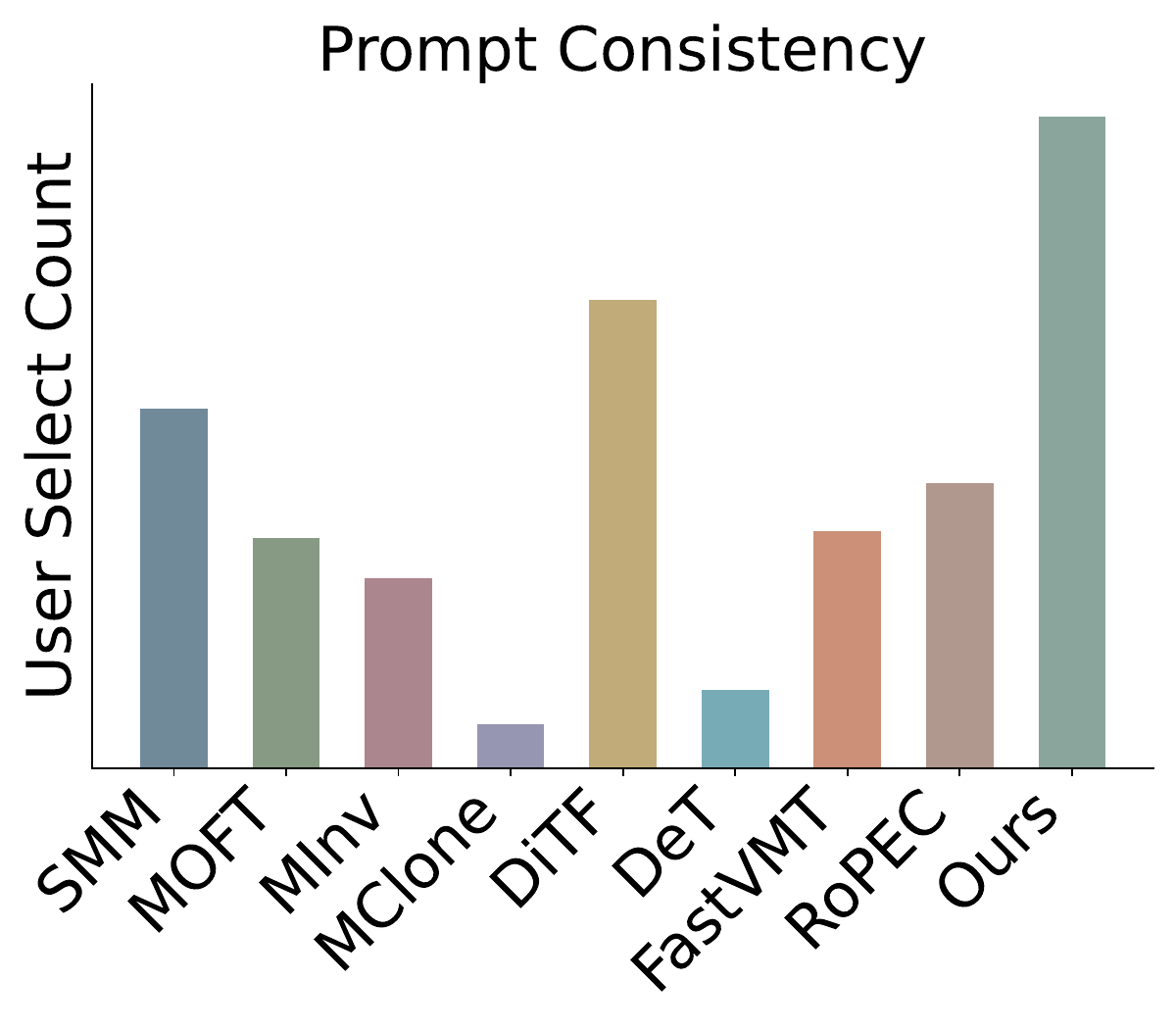}
        \end{subfigure}
    }
\end{minipage}%
\vspace{-1mm}
\caption{Results of user study on motion and text alignment. \emph{MotionAdapter} outperforms the all baseline methods on both aspects.}
\label{user_fig}
\vspace{-3mm}
\end{figure}
We ask 20 participants to compare the motion consistency of videos generated by \emph{MotionAdapter} and several baseline methods with reference videos from DAVIS subset metioned in~\ref{sec:es}. Each participant is shown pairs of videos generated by \emph{MotionAdapter} and each baseline method along with the reference video. They are instructed to select the video that best matches the reference video motion. Additionally, participants are asked to choose the video that is more consistent with the target prompt. We present the user study results in Fig.~\ref{user_fig}. we can see that the majority of participants favored \emph{MotionAdapter} over other baseline methods in terms of motion consistency with the reference video. Additionally, \emph{MotionAdapter} was consistently selected as the method that better aligns with the target prompt. These results further demonstrate the effectiveness of \emph{MotionAdapter} in transferring motion while maintaining semantic consistency, validating its superior performance in comparison to other methods.

\subsection{Ablation Study}
In this section, we conduct ablation studies to validate the effectiveness of each component in \emph{MotionAdapter}-T2V. We develop several variants to analyze the contributions of different modules. 
To evaluate the effectiveness of our cross-frame attention motion extraction and motion transfer guidance, we propose the variant \emph{w/o} Motion Transfer reports the CogVideoX backbone performance without any motion guidance. 
The variant \emph{w/o} Motion Customization applies motion transfer using the extracted attention motion without context-aware motion customization, \ie, optimizing the latent using Eq.~\ref{eq:motion_opt}. 
To evaluate the contribution of the motion refinement module, we propose the variant \emph{w/o} Motion Refinement, which removes the motion refinement module entirely.

\begin{table}[t]
    % \vspace{-12mm}
    \centering
    % \scriptsize
    % \footnotesize
    \caption{Ablation study on the effectiveness of each component in \emph{MotionAdapter}-T2V. The values of \textbf{CS} and \textbf{TC} are multiplied by $10^4$ for better readability.}
    \label{tab:ablation}
    \begin{tabular}{c|c c c}
    \toprule
    \toprule
    \textbf{Variants} & \textbf{CS}$\uparrow$ & \textbf{TC}$\uparrow$ & \textbf{MF}$\uparrow$ \\
    \midrule
   \emph{w/o} Motion Transfer         & 3141 & 9125 & / \\
   \emph{w/o} Motion Refinement       & 3190 & 9312 & 0.776 \\
   \emph{w/o} Motion Customization    & 3192 & 9321 & \textbf{0.802} \\
    \midrule
    \emph{MotionAdapter}-T2V              & \textbf{3240} & \textbf{9323} & 0.794 \\
    \bottomrule
    \bottomrule
	\end{tabular}
    \vspace{-5mm}
\end{table}

The ablation results in Tab.~\ref{tab:ablation} validate the individual contributions of each module within \emph{MotionAdapter}. Compared to the \emph{w/o} Motion Transfer variant, which lacks motion extraction and guidance, the base CogVideoX fails to transfer motion from the reference video, leading to a significant degradation in motion fidelity. Both the motion refinement and context-aware motion customization modules consistently enhance the CLIP Score and Temporal Consistency. Interestingly, removing the Motion Customization module results in a slight decrease in Motion Fidelity. We attribute this to the fact that customization module adapts the reference motion to the target object. While this improves the visual realism of the transfer, it naturally results in a lower Motion Fidelity score when measured with the original motion. Ultimately, the full configuration achieves the best overall performance, confirming that these components play complementary roles in enabling accurate and high-quality motion transfer.
  
% \vspace{-1mm}
\section{Conclusion and Discussion}
% \vspace{-1mm}
\label{sec:conclusion}

In this paper, we introduce \emph{MotionAdapter}, a content-aware video motion transfer framework that refines and customizes attention-based motion representations. By analyzing 3D full-attention maps in DiT, our method disentangles motion from appearance and extracts robust temporal dynamics. To adapt source motions to target content, we propose a content-aware motion customization module that leverages semantic correspondences for foreground objects and merges them with background motions. Extensive experiments demonstrate that our framework achieves temproal coherent and semantically aligned motion transfer, and also supports flexible motion editing tasks. 
\emph{MotionAdapter} depends on the accuracy of DINO-based semantic correspondences, hence inherits the limitations of DINO. Failures in semantic matching can cause imperfect motion transfer. This limitation can be addressed by proposing more advanced object matching models in future.

{
    \small
    \bibliographystyle{ieeenat_fullname}
    \bibliography{egbib}
}

% ########################################### supplementary material ###########################################


\clearpage
\setcounter{page}{1}
\maketitlesupplementary

In this supplementary, we provides more experimental details and results of our \emph{MotionAdapter}.

\section{Algorithm of \emph{MotionAdapter}}
To clearly articulate our approach, we present the algorithm of \emph{MotionAdapter}-I2V in Alg.~\ref{alg:method}.
Lines 1 - 3 extract the cross-frame motion from the reference video. 
In Lines 4 - 12, a dynamic programming algorithm is employed to align the extracted motion relative to the first frame. 
Lines 12 - 21 present the Attention Motion Customization module of the \emph{MotionAdapter}.
Facilitated by DINO ~\cite{oquab2024dinov} and LangSAM~\cite{medeiros2023langsegmentanything}, we obtain the correspondence map $C$ between the first frame of the reference video and the target video. 
Subsequently, based on $C$, the reference motion $\mathcal{M}_\text{ref}$ is customized for the target video to yield $\mathcal{M}_\text{cust}$. 
Line 22 denotes the Attention Motion Refinement module. During the guidance step (Lines 25 - 35), $\mathcal{M}_\text{tgt}$ is extracted following the same protocol as $\mathcal{M}_\text{ref}$, which is then utilized to guide the optimization of $z_\text{t}$ in Line 36. 
Finally, Lines 38 and 41 correspond to the denoising process of the DiT. For more details, please refer to Sec.~\ref{sec:method} of the main manuscript.

% \begin{algorithm}[htbp]
%     \textbf{Input}: reference video $V_ref$, image $I$ and prompt $P$\\
%     \textbf{Output}: target video  \\
%     \Comment{$n^{th}$ frame motion compared to the first frame}
%     \begin{algorithmic}[1]
%         \State $M'_0 \gets \mathbf{0}_{h\times w}$ \Comment{initialize first frame motion $M'_0$}
%         \For{$i \gets 1$ \textbf{to} $f-1$}
%             \State $M'_i \gets 0$
%             \For{$j \gets 0$ \textbf{to} $i-1$}
%                 % \State $X \gets \textsc{Transform}(j,i)$ \Comment{$X = M^{j\rightarrow i}$}
%                 \State $S \gets S + f_{\text{splice}}(M'_j,\,  \mathcal{M}_{j\rightarrow i})$
%             \EndFor
%             \State $M'_i \gets \dfrac{1}{f}\, M'_i$
%         \EndFor
%     \end{algorithmic}
%     \caption{Overview of MotionAdapter}
%     \label{alg:method}
% \end{algorithm}

\begin{algorithm}[htbp]
    % \vspace{-0.05cm}
    \caption{MotionAdapter Algorithm}
    \label{alg:method}
    \small
    \begin{algorithmic}[1]
        \Require Reference Video $\mathcal{V}_\texttt{ref} = {I_0 \dots I_{f-1}}$,  Target First Frame $I$, Target Prompt $\mathcal{P}$
        \State $z_{\text{ref}} \gets \text{AddNoise}(\mathcal{E}(\mathcal{V}_\text{ref}), t_{\text{ref}})$
        \State \textbf{Extract} $\mathcal{A}_{\text{ref}}$ from $\epsilon_\theta (z_{\text{ref}}, \tau_\theta(\text{``''}), t_{\text{ref}})$
        \State \textbf{Extract} $\mathcal{M}_{\text{ref}}^{i \to j}$ ($i,j \in [0,f)$) from $\mathcal{A}_{\text{ref}}$
        
        % --- Reference Motion Calculation ---
        \State $M'^{0}_{\text{ref}} \gets \mathbf{0}^{h \times w}$
        \For{$i \gets 1$ \textbf{to} $f - 1$}
            \State $M'^{i}_{\text{ref}} \gets \mathbf{0}^{h \times w}$
            \For{$j \gets 0$ \textbf{to} $i - 1$}
                \State $M'^{i}_{\text{ref}} \gets M'^{i}_{\text{ref}} + \text{$f_\text{splice}$}(M'^{j}_{\text{ref}}, \mathcal{M}_{\text{ref}}^{j \to i})$
            \EndFor
            \State $M'^{i}_{\text{ref}} \gets \frac{1}{f} M'^{i}_{\text{ref}}$
        \EndFor
        
        % [新增部分 1] Ref Set Definition
        \State $\mathcal{M}_{\text{ref}} \gets \{ M'^{0}_{\text{ref}}, \dots, M'^{f-1}_{\text{ref}} \}$
        
        % --- Correspondence and Warping ---
        \State $I_{\text{ref}}^f, \text{Mask}_b \gets \text{LangSAM}(I_0)$
        \State $I_{\text{tgt}}^f \gets \text{LangSAM}(I)$
        \State $O_{\text{ref}}, O_{\text{tgt}} \gets \mathcal{E}_{\text{DINO}}(I_{\text{ref}}^f), \mathcal{E}_{\text{DINO}}(I_{\text{tgt}}^f)$
        % \State $O_{\text{tgt}} \gets \mathcal{E}_{\text{DINO}}(I_{\text{tgt}}^f)$
        \State $\mathcal{C} \gets \{(j,k) \leftrightarrow (\hat{j},\hat{k}) \mid \texttt{Hungarian}(O_{\text{tgt}}(j,k), O_{\text{ref}}(\hat{j},\hat{k}))\}$
        
        \State $\mathcal{M}_{\text{ref}}^f \gets \mathcal{M}_{\text{ref}} \odot (1 - \text{Mask}_b)$
        \State $\mathcal{M}_{\text{ref}}^b \gets \mathcal{M}_{\text{ref}} \odot \text{Mask}_b$
        \State $\mathcal{M}_{\text{cust}}^f \gets \text{Warp}(\mathcal{M}_{\text{ref}}^f, \mathcal{C})$
        \State $\mathcal{M}_{\text{cust}}^b \gets \text{NN}(\mathcal{M}_{\text{ref}}^b, \text{Mask})$
        \State $\mathcal{M}_{\text{cust}} \gets (1 - \text{Mask}_b) \odot \mathcal{M}_{\text{cust}}^f + \text{Mask}_b \odot \mathcal{M}_{\text{cust}}^b$
        \State $\mathcal{M}_{\text{final}} \gets \text{GauSmooth}(\mathcal{M}_{\text{cust}})$
        
        % --- Guidance Loop ---
        % 使用 down to 表示倒序
        \For{$t \gets T$ \textbf{down to} $T-\text{num}_{\text{guidance step}}$}
            \For{$k \gets 0$ \textbf{to} $\text{num}_{\text{optimize step}}$}
                \State \textbf{Extract} $\mathcal{A}_t$ from $\epsilon_\theta (z_t, \tau_\theta(\mathcal{P}), t)$
                \State \textbf{Extract} $\mathcal{M}_{\text{tgt}}^{i \to j}$ ($i,j \in [0,f)$) from $\mathcal{A}_t$
                
                % [新增部分 2] Target Motion Calculation & Optimization
                \State $\mathcal{M}'^{0}_{\text{tgt}} \gets \mathbf{0}^{h \times w}$
                \For{$i \gets 1$ \textbf{to} $f - 1$}
                    \State $\mathcal{M}'^{i}_{\text{tgt}} \gets \mathbf{0}^{h \times w}$
                    \For{$j \gets 0$ \textbf{to} $i - 1$}
                        \State $\mathcal{M}'^{i}_{\text{tgt}} \gets \mathcal{M}'^{i}_{\text{tgt}} + \text{$f_\text{splice}$}(\mathcal{M}'^{j}_{\text{tgt}}, \mathcal{M}_{\text{tgt}}^{j \to i})$
                    \EndFor
                    \State $\mathcal{M}'^{i}_{\text{tgt}} \gets \frac{1}{f} \mathcal{M}'^{i}_{\text{tgt}}$
                \EndFor
                
                \State $\mathcal{M}_{\text{tgt}} \gets \{ \mathcal{M}'^{0}_{\text{tgt}}, \dots, \mathcal{M}'^{f-1}_{\text{tgt}} \}$
                \State $z_t \gets \arg\min{z_t} \| \mathcal{M}_{\text{tgt}} - \mathcal{M}_{\text{final}} \|_2^2$
                % \State $z_t \gets z^*_t$
            \EndFor
            
            % [新增部分 3] Denoising step
            \State $z_{t-1} \gets \epsilon_\theta (z_t, \tau_\theta(\mathcal{P}), t)$
        \EndFor
        
        % [新增部分 4] Remaining Denoising Steps
        \For{$t \gets (T - \text{num}_{\text{guidance step}})$ \textbf{down to} $0$}
            \State $z_{t-1} \gets \epsilon_\theta (z_t, \tau_\theta(\mathcal{P}), t)$
        \EndFor
    \end{algorithmic}
    % \vspace{-0.1cm}
\end{algorithm}

\begin{table*}[t]
    \centering
    \caption{Ablation study on the selection of guidance blocks and the Top-K parameter in MotionAdapter.}
    \label{tab:sup_ablation}

    \setlength{\tabcolsep}{7pt} 
    \renewcommand{\arraystretch}{1.2}
    
    \begin{tabular}{c|c|ccc}
        \toprule
        \toprule
        \multicolumn{2}{c|}{\makecell{\textbf{Variants}}} &  \textbf{\makecell{CLIP \\ Score $\uparrow$}} & \textbf{\makecell{Temporal \\ Consistency  $\uparrow$}} & \textbf{\makecell{ Motion \\ Fidelity $\uparrow$ }} \\
        \midrule
        \multirow{2}{*}{\makecell{Different Motion \\ Extraction Block}} & $7_{th}$ Block              & 0.3167  & 0.9149 & 0.584  \\
                                                       & $36_{th}$ Block             & 0.3111  & 0.8918 & 0.366  \\
        \midrule
        \multirow{2}{*}{\makecell{Motion Extraction \\ Top-$K$ Selection}} & $K$ = 1                       & 0.3171  & 0.9152 & 0.627  \\
                                                               & $K$ = 10                      & 0.3170  & 0.9146 & 0.606  \\
        \midrule
        \multicolumn{2}{c|}{\makecell{ \emph{MotionAdapter}-I2V \\ ($18_{th}$ block, $K = 3$) }} 
        & \textbf{0.3225}  & \textbf{0.9325} & \textbf{0.641} \\
        \bottomrule
        \bottomrule
    \end{tabular}
\end{table*}
\section{More Ablation Studies}
In this Section, we conduct ablation studies on parameter selection of cross-frame attention motion extraction module.
To evaluate the impact of the extraction block selection within our cross-frame attention motion extraction module on the performance of the \emph{MotionAdapter}, we introduce variants that perform extraction at earlier ($7_{th}$) and later ($36_{th}$) blocks.
Furthermore, to assess how the Top-$K$ selection in Eq.4 for motion field computation influences the \emph{MotionAdapter}, we propose variants utilizing a more restricted ($K$=1) or a more expansive ($K$=10) set of values.
To better evaluate the performance of  \emph{MotionAdapter} when a significant gap exists between the reference video and the target, and to demonstrate its advantages in motion customization, we conduct experiments using the \emph{MotionAdapter}-I2V setting.

As shown in Tab.~\ref{tab:sup_ablation}, the $18_{th}$ block achieves significantly better performance compared to early (e.g., the $7_{th}$ block) and late (e.g., the $36_{th}$ block) blocks, which is consistent with our analysis in Sec.~\ref{para:disentanglement}.
Both $K$=1 and $K$=10 variants underperform relative to the $K$=3 setting, suggesting that $K$=3 setting provides an optimal balance for motion field computation. %~\ref{para:disentanglement}
See our project page for visualized results.

\textbf{Additional Time Cost Details.}
In our implementation, the execution times for DINO matching and LangSAM are 1.45 and 1.56 seconds, respectively. Since \emph{MotionAdapter}-I2V requires only about 4.5 seconds to process these modules, the impact on the overall inference time is negligible.

\balance

\section{Different Backbones}
MotionAdapter can be generalized to any Video diffusion backbones with 3D Diffusion Transformer following the implementation details outlined in Section 4.1 of the main manuscript, such as CogVideoX-2B~\cite{yang2025cogvideox}, Wan~\cite{wan2025wan}, and LTX-Video~\cite{hacohen2024ltx}. 
Furthermore, we provide comparative results between CogVideoX-2B and CogVideoX-5B in our project page.

\section{Failure Cases}
\begin{figure}[t]
  \centering

  % ===== Row 1: 三张图（上方有该行caption）=====
  \begin{subfigure}{\columnwidth}
    \caption{\texttt{\textbf{Reference:} A camel walking in a zoo}}
    \centering
    \includegraphics[width=.32\linewidth]{./For_Supplementary/fail_case/camel/00000.jpg}\hfill
    \includegraphics[width=.32\linewidth]{./For_Supplementary/fail_case/camel/00025.jpg}\hfill
    \includegraphics[width=.32\linewidth]{./For_Supplementary/fail_case/camel/00040.jpg}
  \end{subfigure}

  \medskip
  \vspace{-0.2cm}

  % ===== Row 2: 一张通栏图（上方有该行caption）=====
  \begin{subfigure}{\columnwidth}
    \caption{DINO Correspondence Result}
    \centering
    \includegraphics[width=0.5\linewidth]{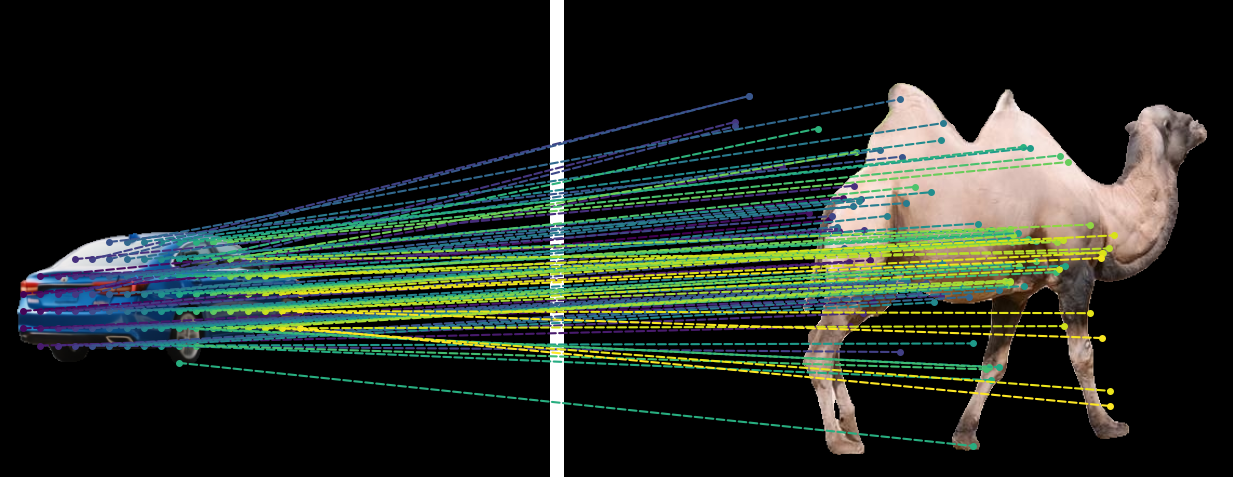}
  \end{subfigure}

  \medskip

  \vspace{-0.2cm}
  % ===== Row 3: 三张图（上方有该行caption）=====
  \begin{subfigure}{\columnwidth}
    \caption{\texttt{A blue Sedan car turning into a driveway}}
    \centering
    \includegraphics[width=.32\linewidth]{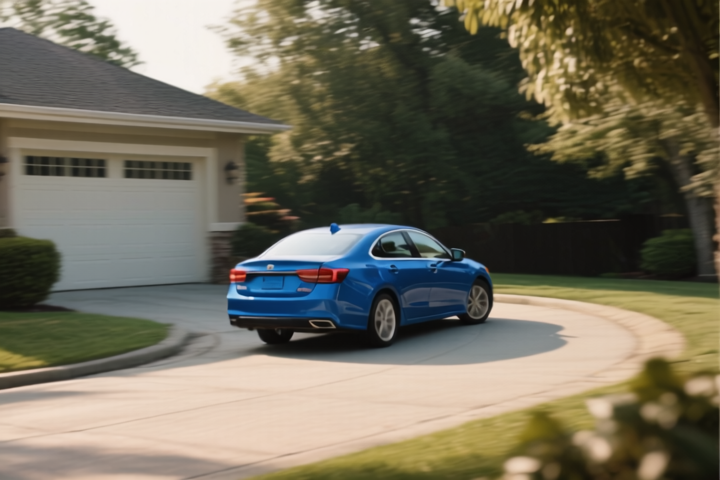}\hfill
    \includegraphics[width=.32\linewidth]{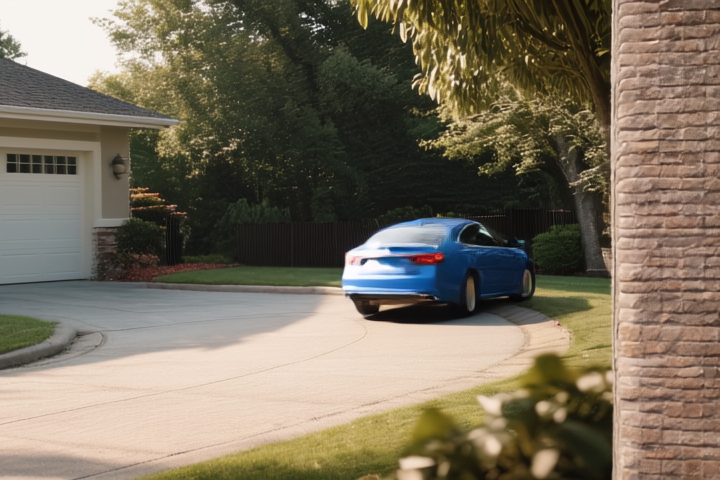}\hfill
    \includegraphics[width=.32\linewidth]{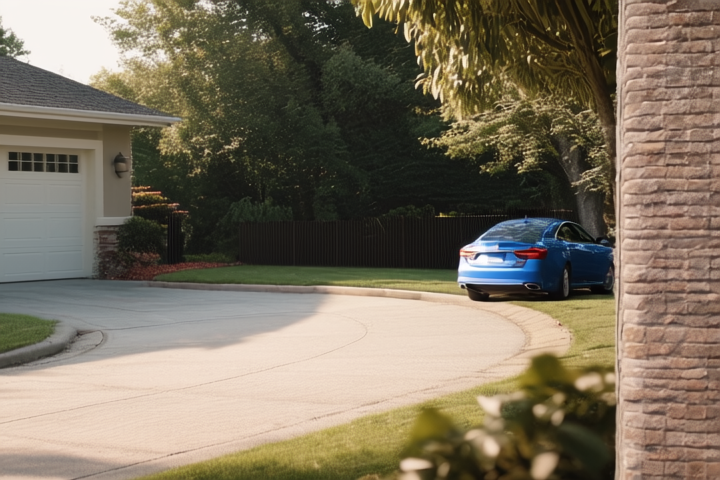}
  \end{subfigure}

  \vspace{-0.2cm}
  \caption{Visualization of a failure case.}
  \label{fig:fail}
  \vspace{-0.5cm}
\end{figure}

We present a failure case of our \emph{MotionAdapter} in Fig.~\ref{fig:fail}. In this instance, DINO~\cite{oquab2024dinov} fails to establish a correct correspondence between the front of the car and the head of the camel, causing the car to miss the turning motion in subsequent frames. Given that current DINO matching can occasionally be inaccurate, integrating more robust matching methods in future work could further enhance the overall performance of \emph{MotionAdapter}.

% \newpage

\end{document}